\documentclass[10pt]{article} 
\usepackage[accepted]{tmlr}

\usepackage{amsmath,amsfonts,bm}

\def\eqref#1{equation~\ref{#1}}

\def\1{\bm{1}}

\DeclareMathAlphabet{\mathsfit}{\encodingdefault}{\sfdefault}{m}{sl}
\SetMathAlphabet{\mathsfit}{bold}{\encodingdefault}{\sfdefault}{bx}{n}

\usepackage{url}

\usepackage{url}            
\usepackage{booktabs}       
\usepackage{amsfonts}       
\usepackage{nicefrac}       
\usepackage{microtype}      
\usepackage{xcolor}         

\usepackage{graphicx}
\usepackage{wrapfig}
\usepackage{rotating}
\usepackage{subcaption} 
\usepackage{listings}
\usepackage{xcolor}
\usepackage{colortbl}  
\usepackage{algorithm}
\usepackage{algpseudocode}
\usepackage{enumitem}

\usepackage{amsmath}
\usepackage{amssymb}
\usepackage{mathtools}
\usepackage{amsthm}

\usepackage[unicode,linktocpage,backref=page]{hyperref}
\usepackage[capitalize,noabbrev]{cleveref}

\usepackage{afterpage}
\usepackage{float}
\usepackage{wrapfig}
\usepackage{lipsum}

\usepackage{tikzsymbols}
\usepackage{textcomp}

\usepackage{multirow}
\usepackage{overpic} 
\usepackage{soul}
\usepackage[most]{tcolorbox}
\usepackage{adjustbox}
\usepackage[acronym]{glossaries}
\glsdisablehyper
\newacronym{dl}{DL}{Deep Learning}
\newacronym{cv}{CV}{Computer Vision}
\newacronym{nlp}{NLP}{Natural Language Processing}
\newacronym{rr}{RR}{Relative Representation}
\newacronym{qkv}{QKV}{Query, Key, Value}
\newacronym{cls}{[CLS]}{Classify token}
\newacronym{fm}{FM}{Foundation Model}
\newacronym{sota}{SOTA}{State-of-the-art}
\newacronym{ode}{ODE}{Ordinary Differential Equations}
\newacronym{e2e}{E2E}{End-to-End}

\newacronym{nn}{NN}{Neural Network}
\newacronym{dnn}{DNN}{Deep Neural Network}
\newacronym{mlp}{MLP}{MultiLayer Perceptron}
\newacronym{cnn}{CNN}{Convolutional Neural Network}
\newacronym{ae}{AE}{AutoEncoder}
\newacronym{vae}{VAE}{Variational AutoEncoder}
\newacronym{linae}{LinAE}{Linearized AutoEncoder}
\newacronym{linvae}{LinVAE}{Linearized Variational AutoEncoder}
\newacronym{vit}{ViT}{Vision Transformer}
\newacronym{gcn}{GCN}{Graph Convolutional Network}
\newacronym{gcnn}{G-CNN}{Group-Equivariant Convolutional Neural Network}

\newacronym{mse}{MSE}{Mean Squared Error}
\newacronym{cka}{CKA}{Centered Kernel Alignment}
\newacronym{cca}{CCA}{Canonical Correlation Analysis}
\newacronym{svcca}{SVCCA}{Singular Value CCA}
\newacronym{pwcca}{PWCCA}{Projection Weighted CCA}
\newacronym{svd}{SVD}{Singular Value Decomposition}
\newacronym{pca}{PCA}{Principal Component Analysis}

\newacronym{abs}{Abs.}{Absolute}
\newacronym{cos}{Cos.}{Cosine}

\newacronym{at}{AT}{Affine Transformation}
\newacronym{tr}{TR}{Translation}
\newacronym{lt}{LT}{Linear Transformation}
\newacronym{is}{IS}{Isotropic Scaling}
\newacronym{ot}{OT}{Orthogonal Transformation}
\newacronym{pt}{PT}{Permutation}
\newacronym{mis}{MIS}{Manifold Isometry}

\newacronym{gflops}{GFLOPs}{Giga Floating-Point Operations}

\newcommand{\mnist}[0]{\texttt{MNIST}} 
\newcommand{\fmnist}[0]{\texttt{F-MNIST}} 
\newcommand{\cifart}[0]{\texttt{CIFAR-10}} 
\newcommand{\cifarh}[0]{\texttt{CIFAR-100}}
\newcommand{\cifarhf}[0]{\texttt{CIFAR-100F}}
\newcommand{\cifarhc}[0]{\texttt{CIFAR-100C}}
\newcommand{\imagenet}[0]{\texttt{ImageNet-1k}} 
\newcommand{\laion}[0]{\texttt{LAION-2B}}  
\newcommand{\agnews}[0]{\texttt{AG News}}
\newcommand{\scene}[0]{\texttt{SceneParse150}}

\newcommand{\vitt}[0]{\texttt{ViT-T}} 
\newcommand{\vits}[0]{\texttt{ViT-S}} 
\newcommand{\vitb}[0]{\texttt{ViT-B}} 
\newcommand{\vitl}[0]{\texttt{ViT-L}} 
\newcommand{\dinos}[0]{\texttt{DINO-S}}
\newcommand{\dinob}[0]{\texttt{DINO-B}}
\newcommand{\deits}[0]{\texttt{DeiT-S}}
\newcommand{\deitb}[0]{\texttt{DeiT-B}}
\newcommand{\clipb}[0]{\texttt{OpenCLIP-ViT-B}}
\newcommand{\clipl}[0]{\texttt{OpenCLIP-ViT-L}}
\newcommand{\cliph}[0]{\texttt{OpenCLIP-ViT-H}}
\newcommand{\mbertb}[0]{\texttt{ModernBERT-B}}

\newcommand{\dino}[0]{\texttt{DINOv2}}
\newcommand{\deit}[0]{\texttt{DeiT}}
\newcommand{\vit}[0]{\texttt{ViT}}

\newacronym{toast}{TOAST}{Transformer Optimization using Adaptive and Simple Transformations}

\definecolor{customgreen}{RGB}{223,239,202}

\definecolor{takeaway}{RGB}{255, 245, 160} 
\definecolor{border}{RGB}{230, 200, 40}   

\definecolor{goodgreen}{RGB}{200,255,200} 
\definecolor{bestapproxblue}{RGB}{200,220,255} 
\definecolor{baselinegray}{RGB}{235,235,235} 

\definecolor{softred}{RGB}{220,20,60}  
\lstdefinestyle{python}{
    language=Python,
    basicstyle=\ttfamily\small\color{black}, 
    keywordstyle=\color{cyan}, 
    stringstyle=\color{green}, 
    commentstyle=\color{yellow}, 
    showstringspaces=false,
    breaklines=true,
}

\title{TOAST \raisebox{-0.15em}{\includegraphics[height=1em]{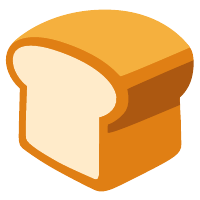}}: Transformer Optimization using Adaptive and Simple Transformations}

\author{\name Irene Cannistraci \email irene.cannistraci@inf.ethz.ch \\
      \addr Department of Computer Science, ETH Zurich
      \AND
      \name Simone Antonelli \\
      \addr CISPA Helmholtz Center for Information Security
      \AND
      \name Emanuele Palumbo \\
      \addr Department of Computer Science, ETH Zurich
      \AND
      \name Thomas M. Sutter \\
      \addr Department of Computer Science, ETH Zurich
      \AND
      \name Emanuele Rodolà \\
      \addr Sapienza University of Rome
      \AND
      \name Bastian Rieck$^{*}$  \\
      \addr University of Fribourg
      \AND
      \name Julia E. Vogt$^{\thanks{Equal advising}}$  \\
      \addr Department of Computer Science, ETH Zurich     
}

\def\month{04}  
\def\year{2026} 
\def\openreview{\url{https://openreview.net/forum?id=fSwMCsBtTG}} 

\begin{document}

\maketitle

\begin{abstract}
Foundation models achieve state-of-the-art performance across different tasks, but their size and computational demands raise concerns about accessibility and sustainability. Existing efficiency methods often require additional retraining or finetuning, limiting their practicality.
Recent findings suggest that deep neural networks exhibit internal representation similarities.
While such similarities across different models have been exploited for enabling techniques such as model stitching and merging, intra-network redundancy remains underexplored as a source for efficiency gains.  
In this paper, we introduce \gls{toast}\footnote{The code is available at: \url{https://github.com/icannistraci/toast}}, a framework that exploits these redundancies to approximate entire transformer blocks with lightweight closed-form mappings, such as linear transformations or even the identity function, without any additional training.  
Across state-of-the-art pretrained vision models (e.g., ViT, DINOv2, DeiT) and datasets ranging from MNIST to ImageNet-1k, \gls{toast} reduces parameters and computation while preserving, and in some cases improving, downstream performance. These results show that large portions of transformer depth can be replaced by trivial functions, opening a new perspective on efficient foundation models.
\end{abstract}

\section{Introduction} \label{sec:intro}

As \glspl{nn} continue to grow in size and complexity, their demand for computational resources has become a critical bottleneck. While larger models consistently achieve \gls{sota} performance, this comes at the cost of substantial memory usage and power consumption, limiting their accessibility and deployment. This challenge is, for instance, most relevant in on-device scenarios, where saving memory, latency, and energy, even by little margins, is critical \citep{pan2022edgevit,li2022efficientformer}.
This has motivated a growing body of work on reducing model complexity.
However, most existing approaches either require additional, resource-intensive training phases or lead to significant drops in accuracy.
Recent studies reveal that there exist strong representational similarities both within and between \glspl{nn}.
In other words, when focusing on intra-network similarities, different blocks often perform overlapping functions or produce highly correlated outputs.\looseness-1

This redundancy suggests an opportunity: \emph{instead of retraining or pruning, can we approximate these blocks with simpler transformations?}
To address this question, we propose \glsentryfull{toast}, a novel framework that exploits block-level representational redundancy to replace transformer blocks with lightweight transformations. By doing so, \gls{toast} reduces parameter count and computational cost, while maintaining (and in some cases even improving) downstream task performance. Crucially, our method is training-free, making it simple, efficient, and widely applicable, even in resource-constrained scenarios such as deployment on edge devices, where even the smallest available models may exceed memory or power budgets. Our main contributions are as follows: 
\begin{itemize}
    \item We propose \gls{toast}, a simple yet effective framework that replaces transformer blocks with lightweight transformations (e.g., linear maps or even the identity), significantly reducing parameters and computational cost while preserving downstream performance (\Cref{fig:method}).
    \item We introduce linear approximation error as a stable and computationally lightweight criterion for identifying redundant transformer blocks (\Cref{tab:skip_recommendations,tab:skip_by_budget,alg:layer_skip_selection,alg:linear_error}) and we present a systematic analysis of block-wise representational similarities in pretrained vision transformers, revealing consistent redundancy patterns across diverse models and motivating the possibility of approximating entire blocks (\Cref{fig:latent-analysis-cka,fig:app-latent-analysis-cka-cls}).
    \item We empirically demonstrate that accurate block approximations can be obtained from only a few hundred samples, showing that block redundancy can be exploited without requiring large-scale retraining (\Cref{table:classification-results-preliminary,fig:num-samples-ablation,table:transformation-ablation}).
    \item We extensively validate our approach across a wide spectrum of vision models (e.g., \dinob{}, \vitl{}, \deits{}, \vits{}, \dinos{}, \vitt{}) and datasets ranging from \mnist{} to \imagenet{}, confirming both the generality and efficiency of the method (\Cref{table:classification-results-cifar100,table:classification-results-preliminary,table:classification-results-imagenet,table:app-vit-s-classification,table:app-dino-s-classification,table:app-vit-t-classification,table:app-vitb-classification-results,table:clip-zeroshot}).
    \item We preliminarily validate the application of \gls{toast} beyond vision classification, including semantic segmentation using \vits{} and \dinob{} on \scene{}, and text classification using \mbertb{} on \agnews{} (\Cref{table:segmentation,app:text-classification,table:text-classification}).
    \end{itemize}
\section{Related Work} \label{sec:related}

\paragraph{Measuring Similarities} A range of metrics have been introduced to assess the similarity between latent spaces generated by different \glspl{nn} \citep{klabunde2023similarity,Ballester2023}. One established approach is \gls{cca} \citep{hotelling1992relations}, known for its invariance to linear transformations. Variants of \gls{cca}, such as \gls{svcca} \citep{raghu2017svcca}, aim to enhance robustness, while techniques like \gls{pwcca} \citep{morcos2018insights} mitigate sensitivity to small perturbations. Another widely used metric, \gls{cka} \citep{kornblith2019similarity}, captures the similarity between latent spaces while ignoring orthogonal transformations. However, recent work \citep{davari2022reliability} highlights that this metric can be sensitive to shifts in the latent space. Additionally, \citet{barannikov2021representation} proposes a method to compare two data representations by measuring the multi-scale topological dissimilarity, while \citet{NEURIPS2024_79be41d8} leverages the principles of spectral geometry to model and analyze the relationships between distinct latent spaces.

\paragraph{Leveraging Similarities} \citet{valeriani2024geometry} examines the intrinsic dimensionality and neighbor compositions of representations in transformer models. \citet{Kvinge2022} investigates how models process variations in data points across layers, while \cite{nguyen2020wide} assesses the impact of network depth and width on hidden representations. Additionally, \citet{crisostomi2023from} studies the conditions under which two latent spaces can be merged into a unified one. \citet{moschella2022relative} constructs a unified space shared by different \glspl{nn}, enabling zero-shot stitching of independently trained models across different modalities \citep{norelli2023asif}. More recently, \citet{cannistraci2024from} enables model stitching without explicit assumptions about the transformation class connecting the latent manifold embeddings, or with only partial correspondence between latent spaces \citep{DBLP:conf/iclr/CannistraciMMFN23}. Finally, \citet{pmlr-v243-lahner24a, maiorca2024latent} demonstrate that representations learned by distinct \glspl{nn} can be aligned using simple transformations.\looseness=-1

\paragraph{Architectural Efficiency} While large-scale models with billions or even trillions of parameters continue to achieve state-of-the-art performance, their growth comes with trade-offs, such as slower inference times and significantly higher computational costs.
Improving the efficiency of \gls{dnn} has been a long-standing area of research. For instance, \citet{NIPS2016_37bc2f75} shows that removing residual blocks from deep \glspl{cnn} only marginally impacts performance, which inspired approaches to reduce inference time by dynamically deciding which layers to execute based on the input \citep{Wu_2018_CVPR, Veit_2018_ECCV}. Additionally, various techniques to enhance efficiency have emerged, such as early exiting and model pruning. Early exit strategies, which introduce intermediate output layers at different stages of the network, have been shown to reduce inference time \citep{xin2020deebert, zhou2020bert, yu2022width, tang2023you}. However, these approaches require the training of intermediate classifiers to enable exits at predefined layers. Alternatively, model pruning reduces computational load by either removing individual weights based on specific criteria, such as gradient information \citep{ma2023llm}, entropy \citep{liao2023can}, or second-order information \citep{NEURIPS2020_d1ff1ec8}, or by eliminating larger structural components, like channels or residual blocks in ResNets \citep{bai2023unified, wang2023practical}, weights in LLMs \citep{sun2023simple} and self-attention layers in Transformers \citep{zhang2020accelerating, sajjad2023effect, venkataramanan2024skipattention, Zhang_2024_CVPR}. Although effective, these approaches require training the model from scratch and, in the best case, finetuning. However, \citet{bai2023unified} shows that for \glspl{cnn}, this additional training step can sometimes be avoided.

Unlike other methods, \gls{toast} leverages intra-network similarities to reduce vision transformer complexity \emph{without the need for additional training steps} while maintaining competitive performance.

\section{Block Approximation} \label{sec:method}

The central idea of our approach is that it is possible to leverage representation similarities within transformer-based architectures to replace entire blocks with simpler transformations. In this work, a \emph{block} refers to a sequence of layers including multi-head self-attention, normalization, and feed-forward layers, that function together as a cohesive unit. By replacing these blocks with simpler transformations, we can reduce the computational complexity of the network while maintaining its core functionality.

\begin{figure}[htb]
\vspace{1em}
    \begin{centering}
        \begin{overpic}[width=0.7\textwidth,tics=10]{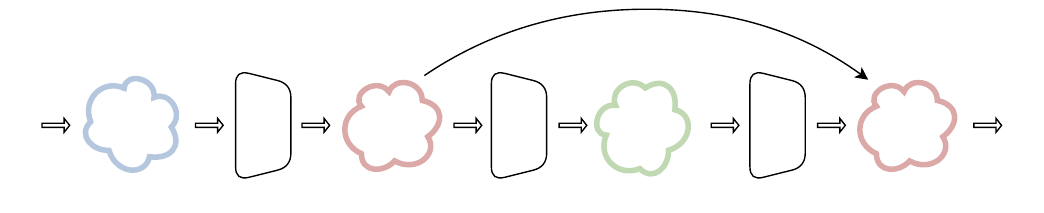} 
            \put(-2,8.15){$\dots$}  
            \put(98,8.15){$\dots$} 
            
            \put(60,21){$\mathcal{T}$}
            
            \put(24.3,0.5){$s$}
            
            \put(49.75,8.5){\makebox(0,0){\color{softred}\scalebox{8}{$\scriptscriptstyle\times$}}}  
            \put(74.6,8.5){\makebox(0,0){\color{softred}\scalebox{8}{$\scriptscriptstyle\times$}}}  
            
            \put(35,0){$\mathbf{X}^{(s)}$}
            \put(84,0){$\mathbf{X}^{(e)}$}

            \put(73.5,0.5){$e$}
        \end{overpic}
        \caption{\textbf{Framework Description}. Given two latent spaces $\mathbf{X}^{(s)}$ and $\mathbf{X}^{(e)}$ corresponding to the outputs of blocks $s$ and $e$ for a random subset of 500 training samples, \gls{toast} estimates a lightweight transformation $\mathcal{T}$ such that $\mathbf{X}^{(e)} \approx \mathcal{T}(\mathbf{X}^{(s)})$. This allows \emph{entire} transformer blocks to be approximated by simple closed-form mappings (e.g., linear or identity), reducing parameters and computation without retraining.}
        \label{fig:method}
    \end{centering}
\end{figure}

\paragraph{Approximating Transformer Blocks} Given two blocks $s$ and $e$, our goal is to replace the intermediate blocks $s+1, \dots, e$ with a single, lightweight transformation that maps the output of block $s$ directly to an approximation of the output of block $e$. This approach allows us to skip the computation of blocks $s+1, \dots, e$, effectively reducing the overall computational costs. This approximation can be repeated for multiple, non-overlapping blocks, i.e., blocks $(s_i, e_i)$ and $(s_j, e_j)$ with $e_i < s_j$.
An overview of the method is provided in \Cref{fig:method}.

Let $\mathbf{X}^{(s)} \in \mathbb{R}^{| \mathcal{D}_\text{sub} | \times d_s}$ and $\mathbf{X}^{(e)} \in \mathbb{R}^{| \mathcal{D}_\text{sub} | \times d_e}$ represent the output representations from block $s$ and $e$ respectively, for the data points in $\mathcal{D}_\text{sub} \subset \mathcal{D}$, sampled uniformly at random from the full training dataset $\mathcal{D}$. Our objective is to find a transformation $\mathcal{T}: \mathbb{R}^{d_s} \to \mathbb{R}^{d_e}$ such that:
\[
\mathbf{X}^{(e)} \approx \mathcal{T}(\mathbf{X}^{(s)})
\]
In this work, we consider $\mathcal{T}$ to be the \emph{identity} or a \emph{linear transformation} $\mathbf{T}$.
We can compute the linear transformation $\mathbf{T}$ by minimizing the squared error between the transformed output $\mathcal{T}(\mathbf{X}^{(s)})$ and the actual $\mathbf{X}^{(e)}$:\looseness=-1 
\[
\mathbf{T} = \underset{\mathcal{T}}{\arg\min} \| \mathbf{X}^{(e)} - \mathcal{T}( \mathbf{X}^{(s)} ) \|_2^2
\]
This optimization problem allows for a closed-form solution that efficiently computes the optimal transformation $\mathbf{T}$. The solution bypasses the computation of \emph{all} layers between any two blocks $s$ and $e$, replacing them with $\mathbf{T}$. 
This approximation reduces computational complexity while minimally affecting internal representations, as illustrated in \Cref{fig:app-pca-approx-dino-1,fig:app-pca-approx-dino-11,fig:app-pca-approx-vit-1,fig:app-pca-approx-vit-11,fig:app-pca-approx-deit-11}, and preserves compatibility with downstream classifiers, achieving significant compression as shown in \Cref{table:classification-results-preliminary,table:classification-results-imagenet,table:classification-results-cifar100,table:app-vit-s-classification,table:app-dino-s-classification,table:app-vit-t-classification,table:app-vitb-classification-results}.

\paragraph{Patterns of Similarity between Transformer Blocks}
Inspired by existing results \cite{venkataramanan2024skipattention}, which show that multi-head attention modules exhibit similarity in learned representations, we investigate whether pretrained foundation models contain \emph{entire blocks} that produce highly similar representations. Rather than using \gls{cka} to measure representational similarity, we quantify how well the output of a later block can be reconstructed from an earlier one using a simple linear transformation. All representations are computed using only the \texttt{[CLS]} token, providing a consistent and semantically aligned basis for comparing blocks.

Given representations $\mathbf{H}_s$ and $\mathbf{H}_e$ extracted from blocks $s < e$, we learn the optimal linear map $\mathbf{W}^*$ that solves\looseness=-1
\[
\mathbf{W}^* = \operatorname*{arg\,min}_{\mathbf{W}} \|\mathbf{H}_e - \mathbf{H}_s \mathbf{W}\|_F^2.
\]
We measure similarity via the normalized residual error
\[
\epsilon(s,e) = \frac{\|\mathbf{H}_e - \mathbf{H}_s \mathbf{W}^*\|_F}{\|\mathbf{H}_e\|_F},
\]
where lower values indicate that block~$e$'s representations are well explained by a linear transformation of block~$s$.

By computing the metric for all block pairs, using only a small random subset of the training data (i.e., 50 samples), and ranking them, we can automatically identify blocks whose computations contribute minimally beyond a near-linear mapping. We additionally perform an ablation study comparing several candidate similarity metrics for block selection, and we report these results in \Cref{sec:app-metric-ablation}. The procedure used to automatically extract the top-$k$ skip candidates is summarized in \Cref{alg:layer_skip_selection}, and the linear approximation error is detailed in \Cref{alg:linear_error}. 
\section{Experiments} \label{sec:experiments}

In this section, we first analyze the similarities between different transformer blocks to motivate their approximation using simple transformations. We then present comprehensive results on image classification across various models and datasets to demonstrate the effectiveness and efficiency of the proposed method. Beyond these core results, we further study the robustness of \gls{toast} through ablations on the number of samples required for approximation and the choice of translator architecture, and we additionally examine \gls{e2e} finetuning after \gls{toast} is applied, showing that additional training compute can recover most of the accuracy lost. Overall, our findings show that \gls{toast} achieves strong performance while producing lighter and faster models. Due to space constraints, additional results on zero-shot image classification, as well as further qualitative and quantitative analyses, are provided in the Appendix (\Cref{sec:app-image-zeroshot,sec:app-image-classification}).

\subsection{Latent Analysis} \label{sec:latent-analysis} 

In this section we investigate similarities in the latent representations of \dinob{} and \deits{} on five datasets: \cifart{}, \cifarh{}, \mnist{}, \fmnist{}, and \imagenet{}. We compute the linear approximation error using only the \texttt{[CLS]} token, averaged over a small subset of 50 training samples. This is sufficient to reveal block-level similarity patterns while remaining computationally efficient. Additional results with other pretrained vision transformers (\vitt{}, \vits{}, \dinos{}, \vitb{}) are provided in \Cref{sec:app-similarities}, showing consistent patterns for each model across different datasets.

\begin{figure}[h]
    \centering
    \vspace{1.5em}
    \begin{minipage}[t]{.18\textwidth}
        \centering
        \begin{overpic}[width=\textwidth]{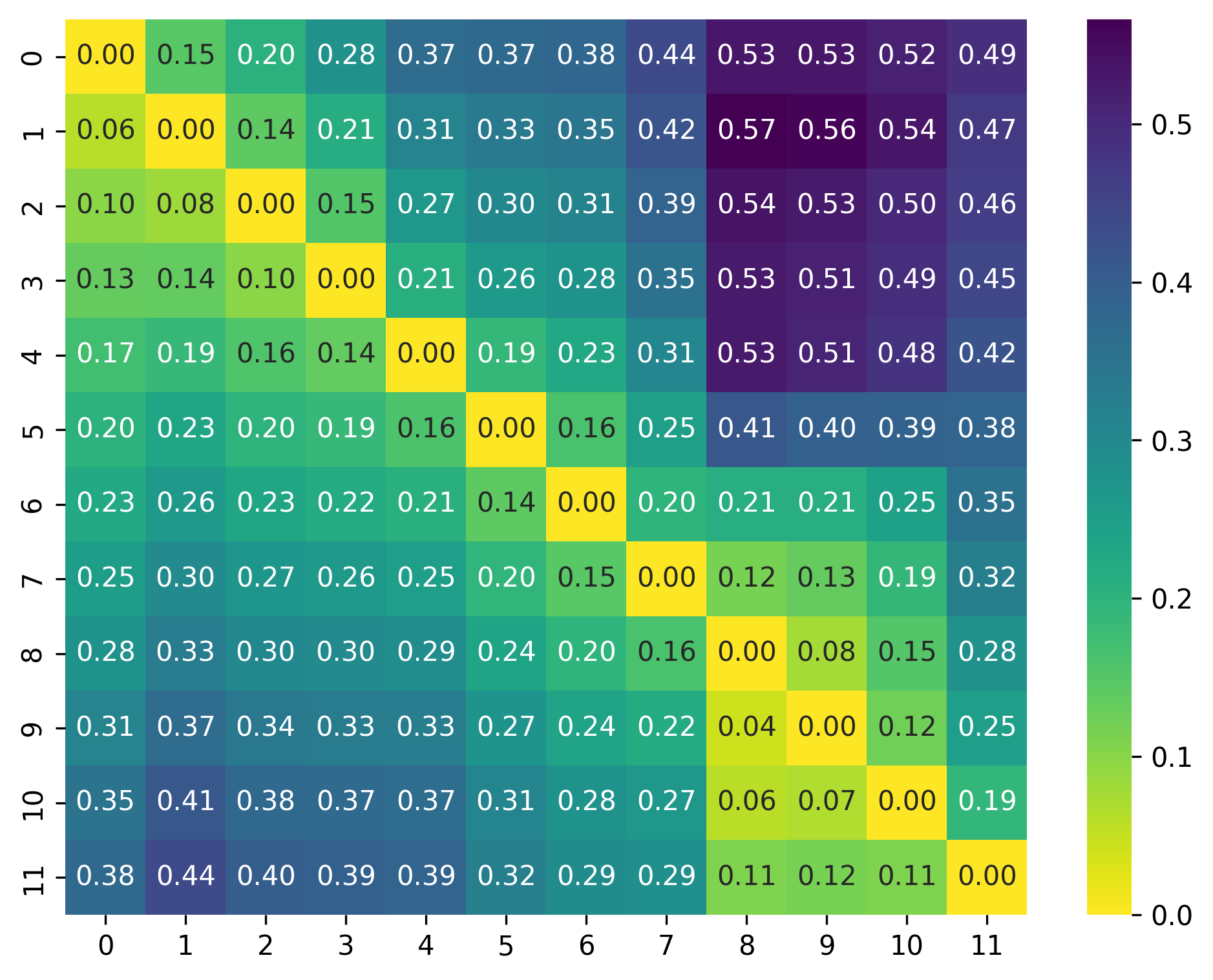}
            \put(-10,22){\rotatebox{90}{\dinob{}}}
            \put(30,83){\mnist{}}
        \end{overpic}
    \end{minipage}
    \begin{minipage}[t]{.18\textwidth}
        \centering
        \begin{overpic}[width=\textwidth]{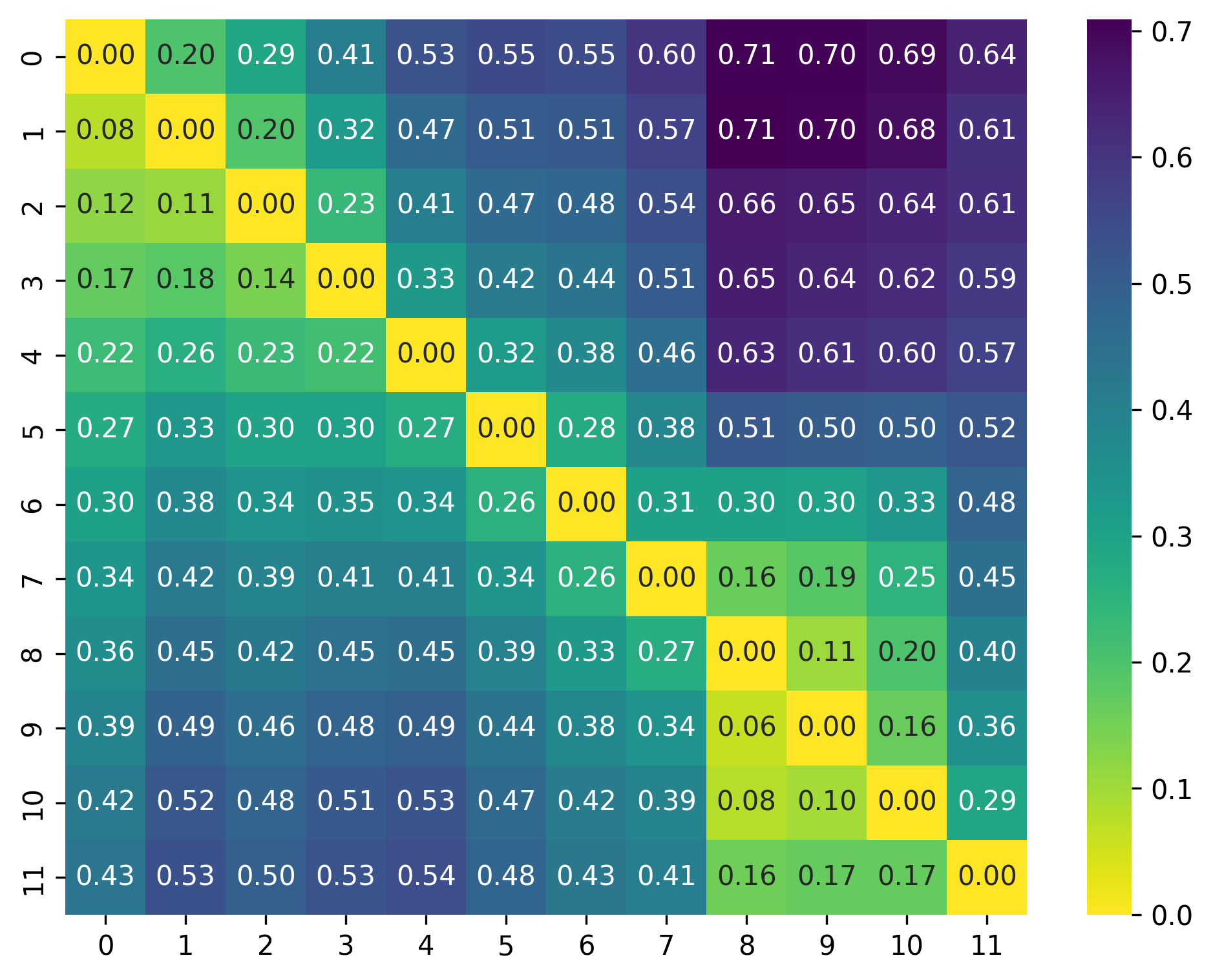}
            \put(25,83){\fmnist{}}
        \end{overpic}
    \end{minipage}
    \begin{minipage}[t]{.18\textwidth}
        \centering
        \begin{overpic}[width=\textwidth]{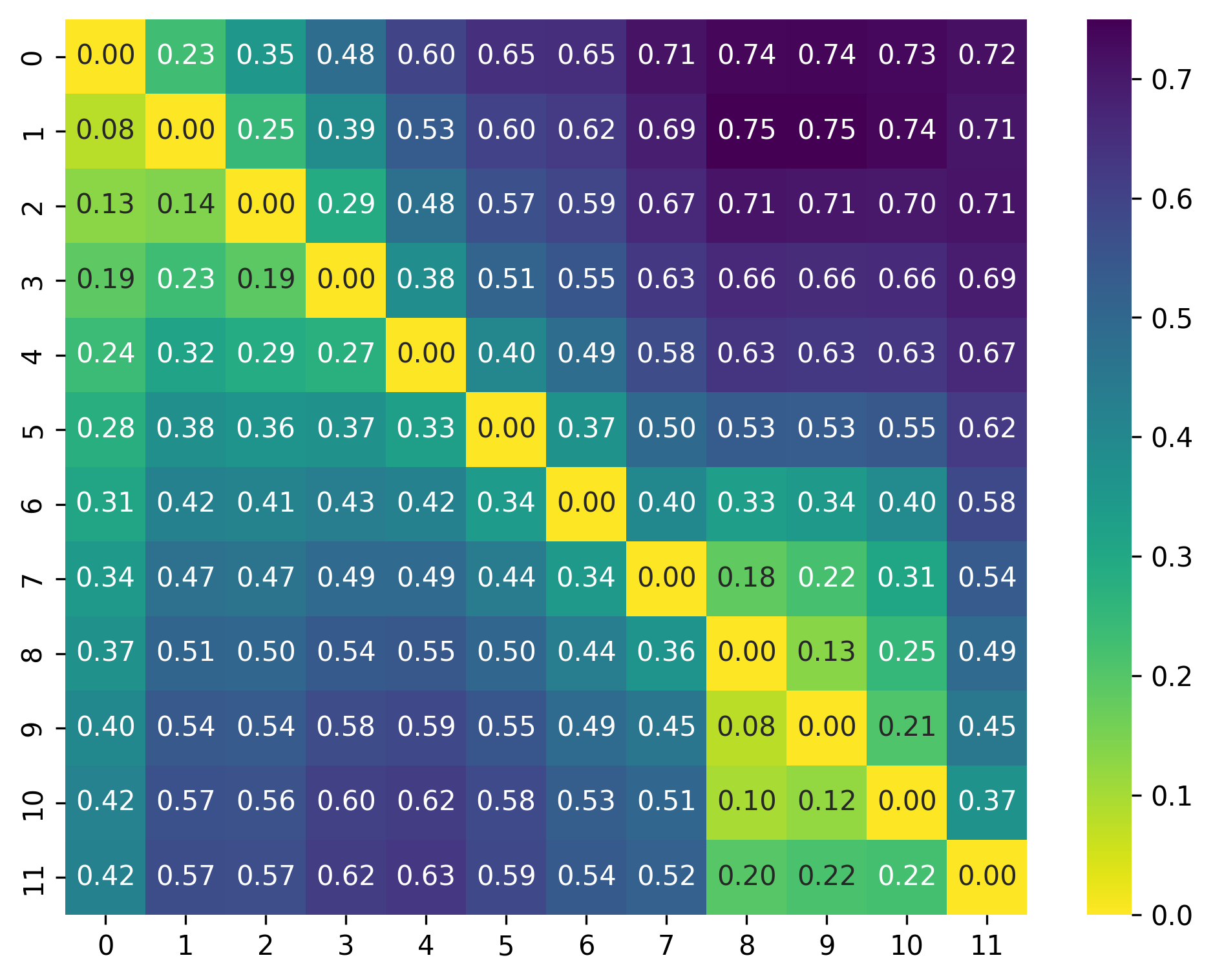}
            \put(21,83){\cifart{}}
        \end{overpic}
    \end{minipage}
    \begin{minipage}[t]{.18\textwidth}
        \centering
        \begin{overpic}[width=\textwidth]{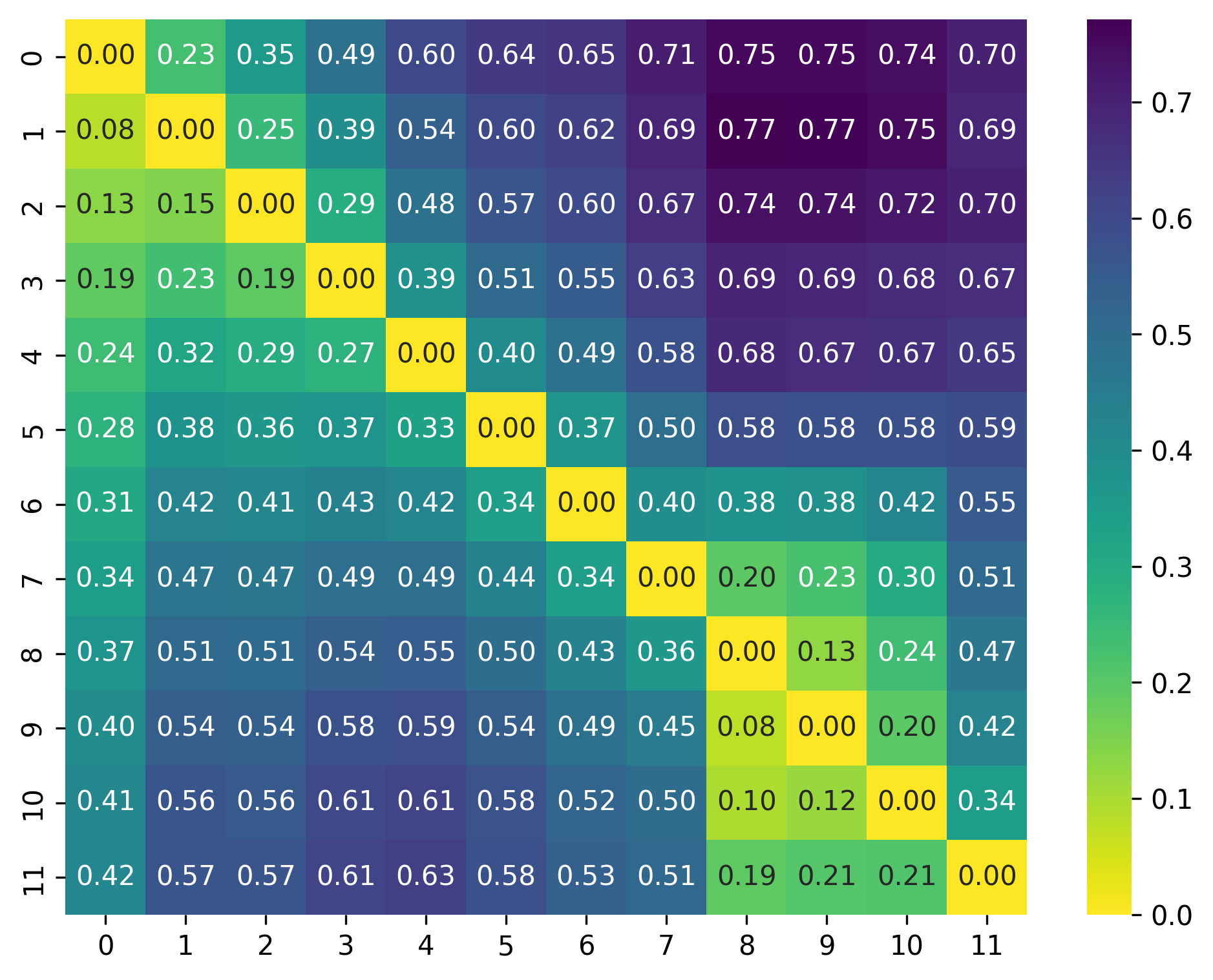}
            \put(17,83){\cifarh{}}
        \end{overpic}
    \end{minipage}
    \begin{minipage}[t]{.18\textwidth}
        \centering
        \begin{overpic}[width=\textwidth]{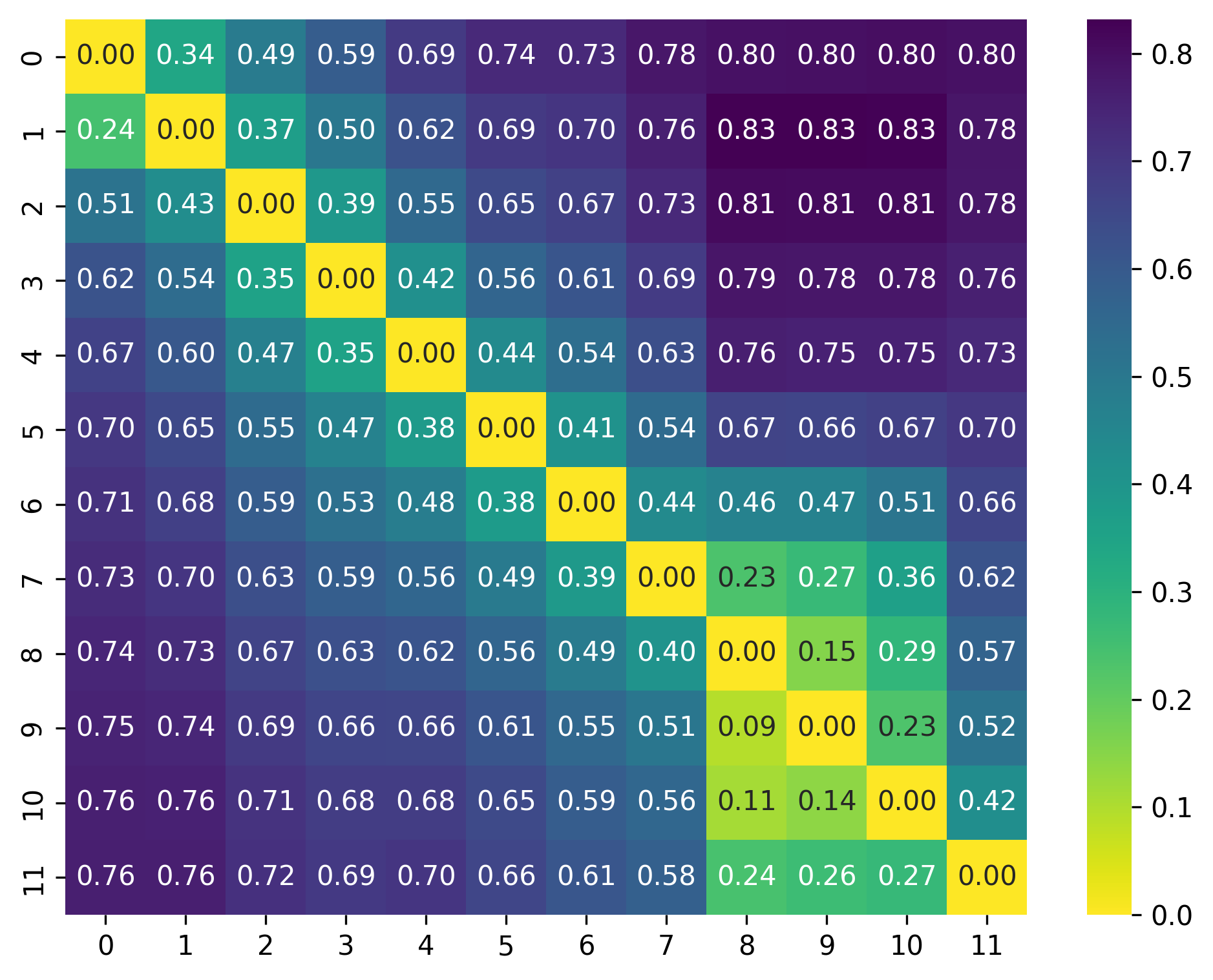}
            \put(11,83){\imagenet{}}
        \end{overpic}
    \end{minipage}

    \begin{minipage}[t]{.18\textwidth}
        \centering
        \begin{overpic}[width=\textwidth]{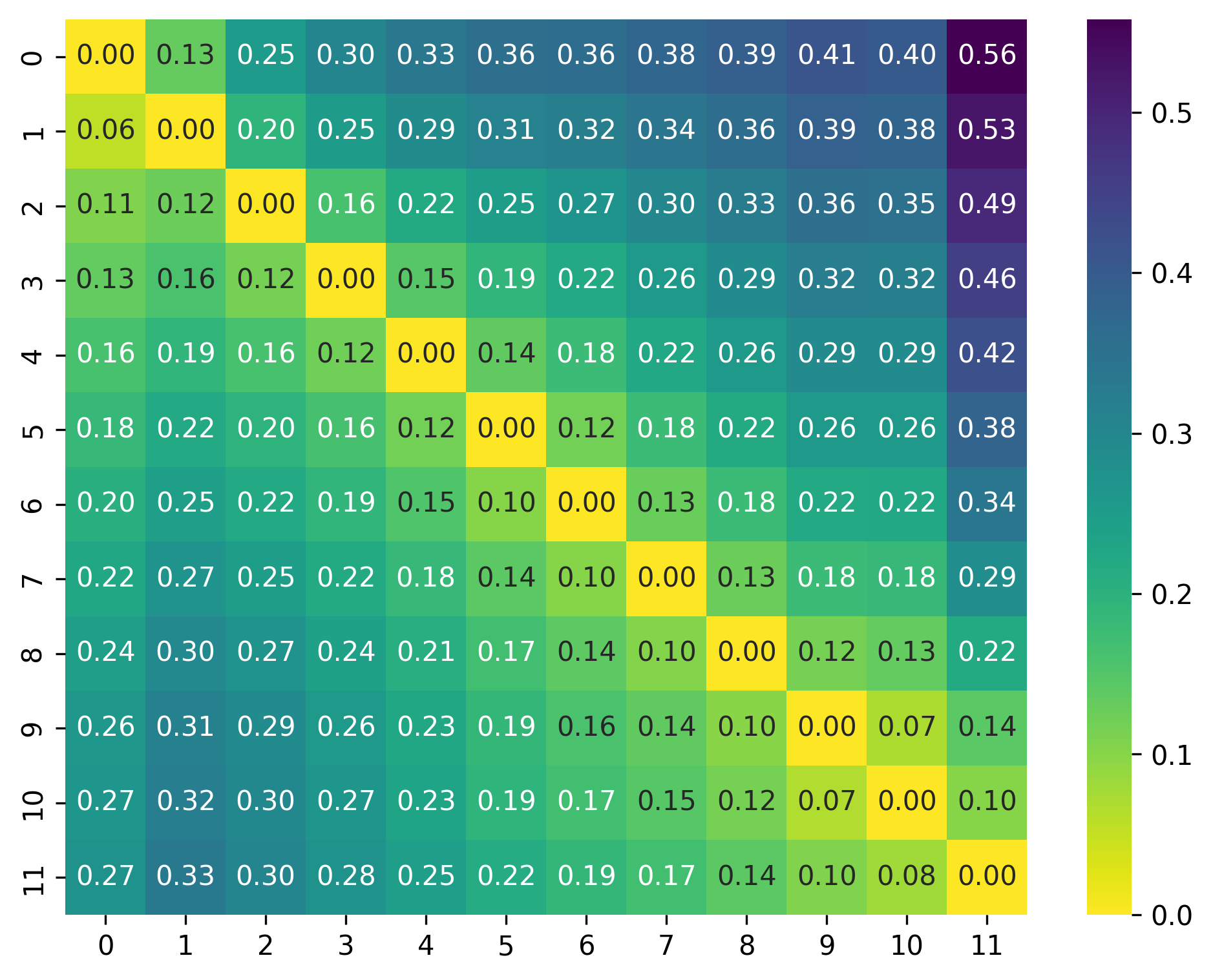}
            \put(-10,22){\rotatebox{90}{\deits{}}}
        \end{overpic}
    \end{minipage}
    \begin{minipage}[t]{.18\textwidth}
        \centering
        \includegraphics[width=\textwidth]{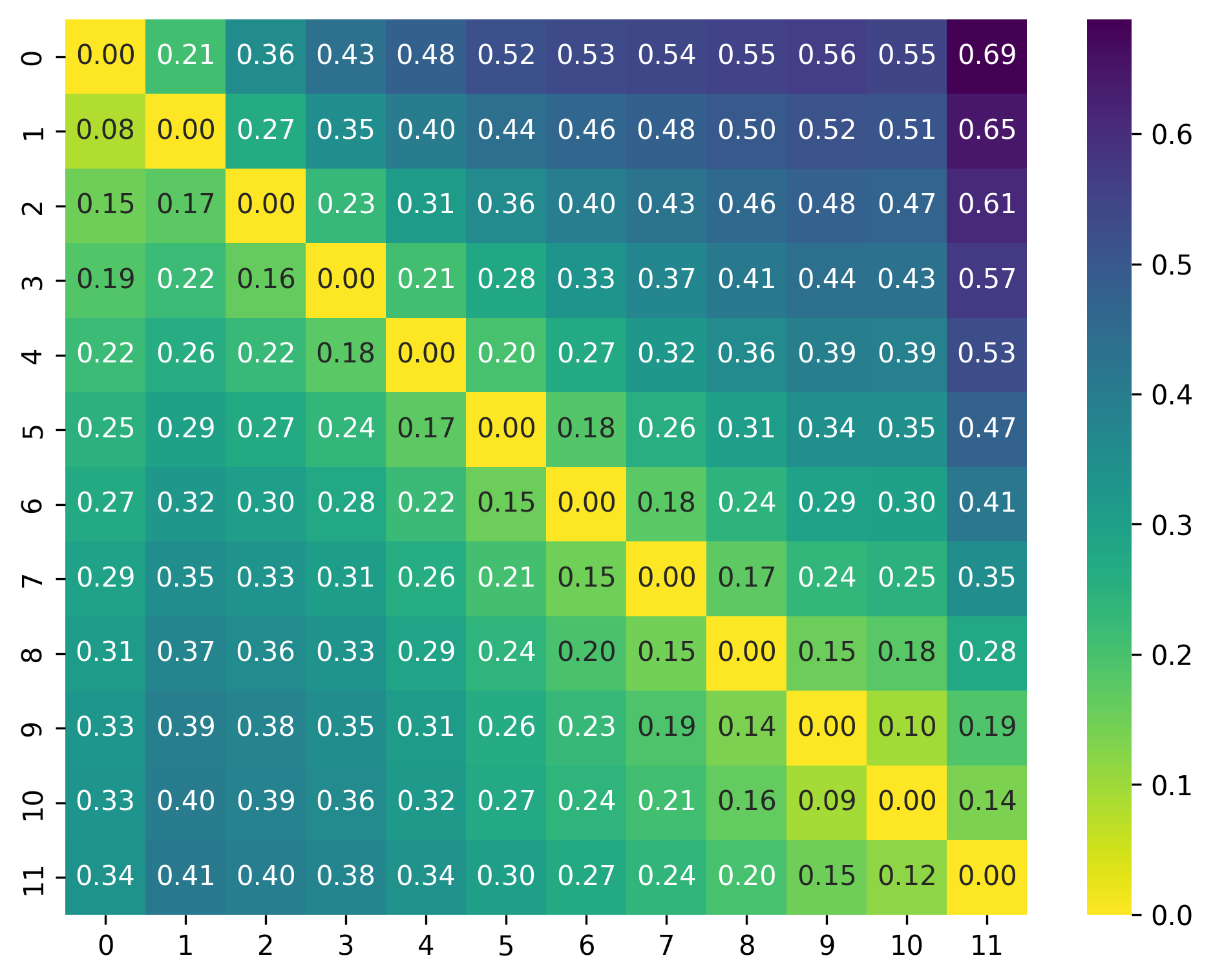}
    \end{minipage}
    \begin{minipage}[t]{.18\textwidth}
        \centering
        \includegraphics[width=\textwidth]{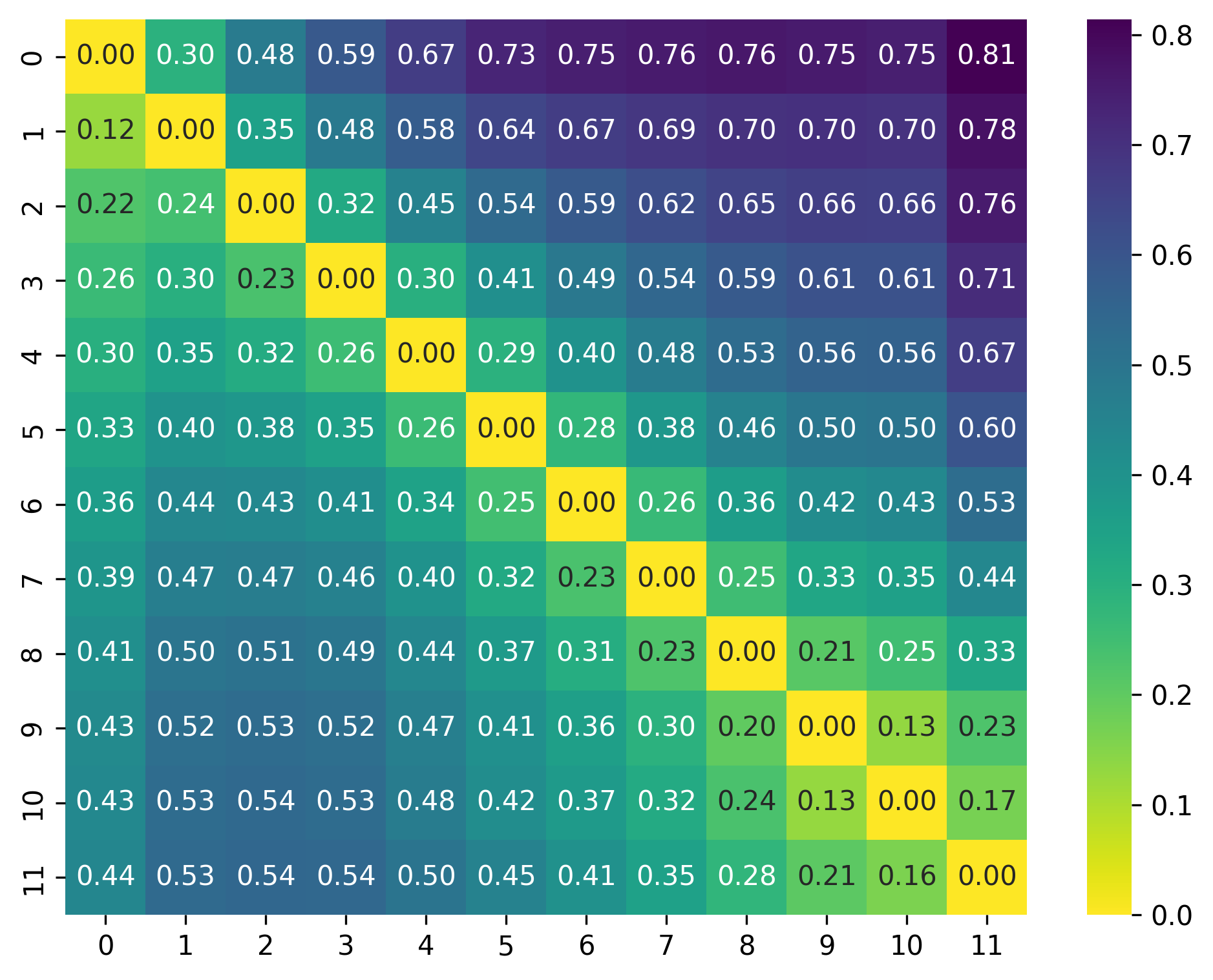}
    \end{minipage}
    \begin{minipage}[t]{.18\textwidth}
        \centering
        \includegraphics[width=\textwidth]{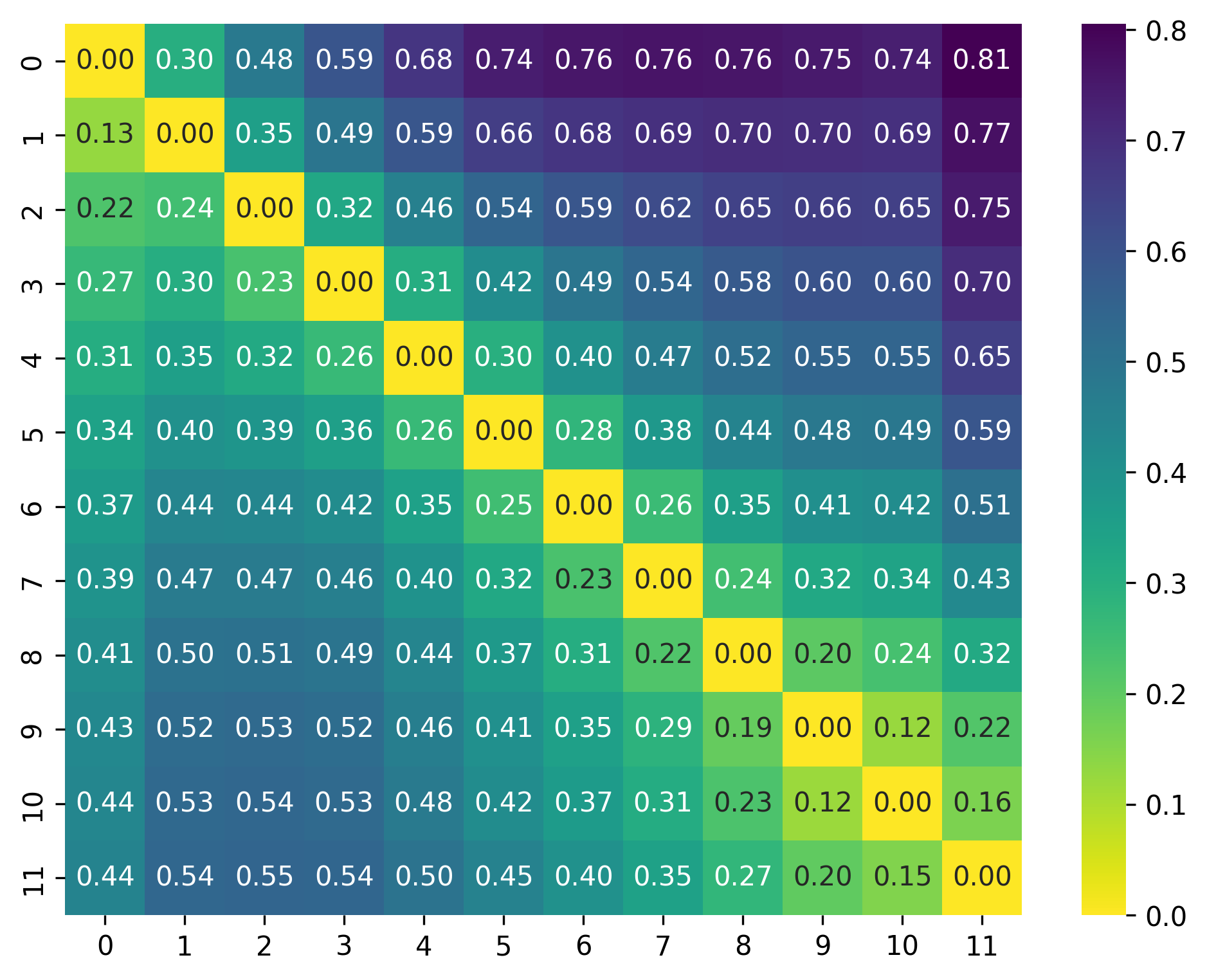}
    \end{minipage}
    \begin{minipage}[t]{.18\textwidth}
        \centering
        \includegraphics[width=\textwidth]{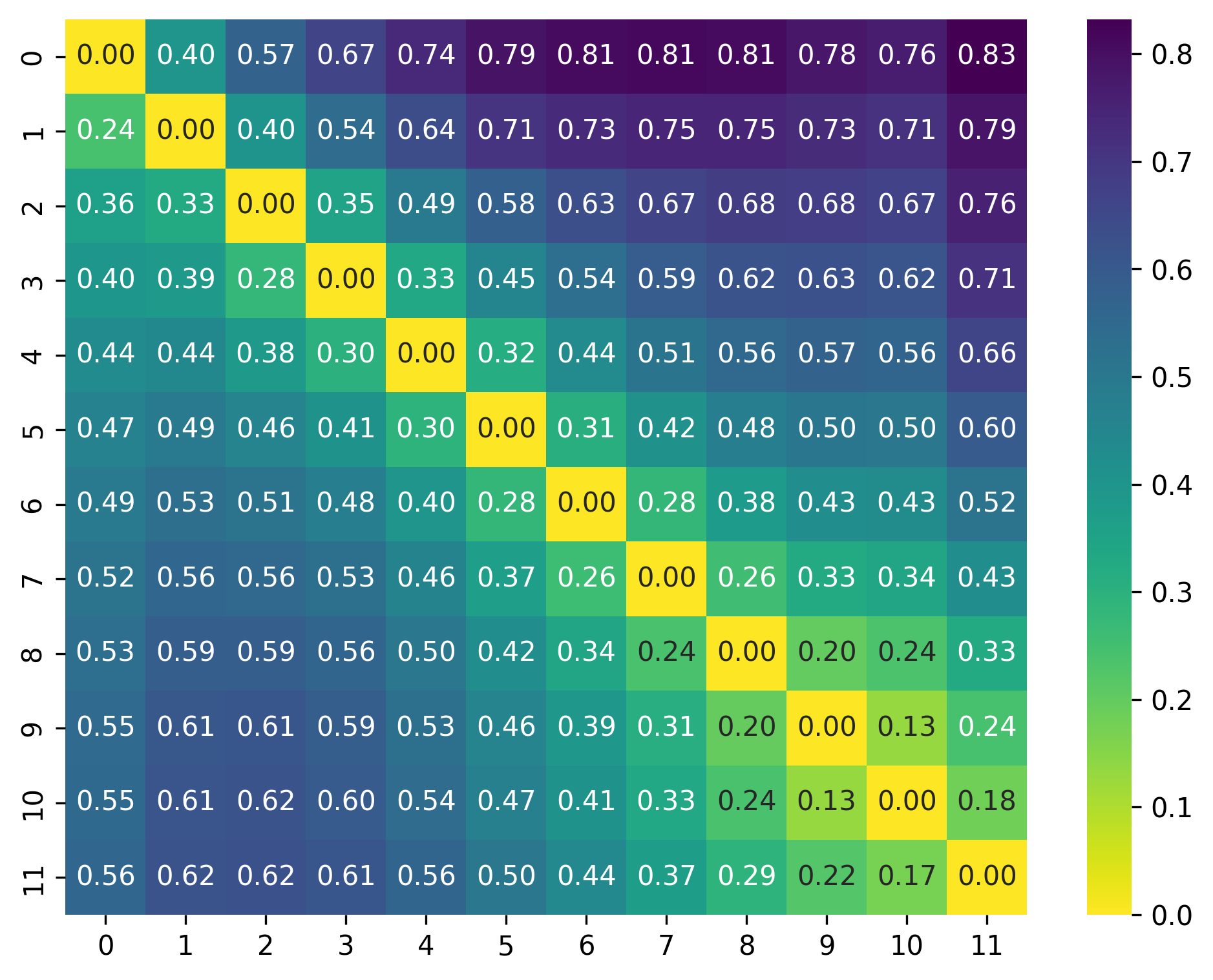}
    \end{minipage}
    \caption{\textbf{Block Similarities}. Block-by-block similarities in \dinob{} and \deits{} models across five datasets: \mnist{}, \fmnist{}, \cifart{}, \cifarh{}, and \imagenet{}. Each matrix quantifies the linear approximation error using only the \texttt{[CLS]} token, averaged over a small subset of 50 training samples. The matrices reveal that the similarity between blocks is predominantly influenced by the model rather than the specific dataset. Additional results in \Cref{sec:app-similarities}.}
    \label{fig:latent-analysis-cka}
\end{figure}

\paragraph{Do pretrained vision transformer models exhibit block-wise similarity patterns?} The results in \Cref{fig:latent-analysis-cka} reveal that while the similarity patterns differ across models, they remain largely consistent for the same model across different datasets. This suggests that the similarity structure between computational blocks is predominantly influenced by the model itself.
Although the general similarity pattern remains the same, the differences in values become more pronounced (i.e., the block structure becomes more evident) as the complexity of the dataset increases (e.g., from \mnist{} to \imagenet{}).
These findings align with observations from \cite{nguyen2020wide}, where \glspl{dnn} trained from scratch exhibit a distinctive ``block structure'' in their representations, which is linked to model overparameterization. Our results extend this observation to vision pretrained foundation models, showing that such a structure is primarily an intrinsic property of the model. Moreover, these consistent block-wise patterns indicate potential targets for approximation, suggesting that entire blocks may be replaced with simpler transformations without substantially altering the model's internal representations.\looseness=-1

\begin{tcolorbox}[colframe=takeaway, colback=takeaway, opacityback=1.0, boxrule=0mm, arc=2mm]
    \textbf{Takeaway} Pretrained vision foundation models present block-wise similarity patterns that are primarily determined by the model itself.
\end{tcolorbox}

\begin{figure}[h]
    \centering
    \begin{subfigure}{0.55\textwidth}
        \centering
        \begin{overpic}[width=\textwidth]{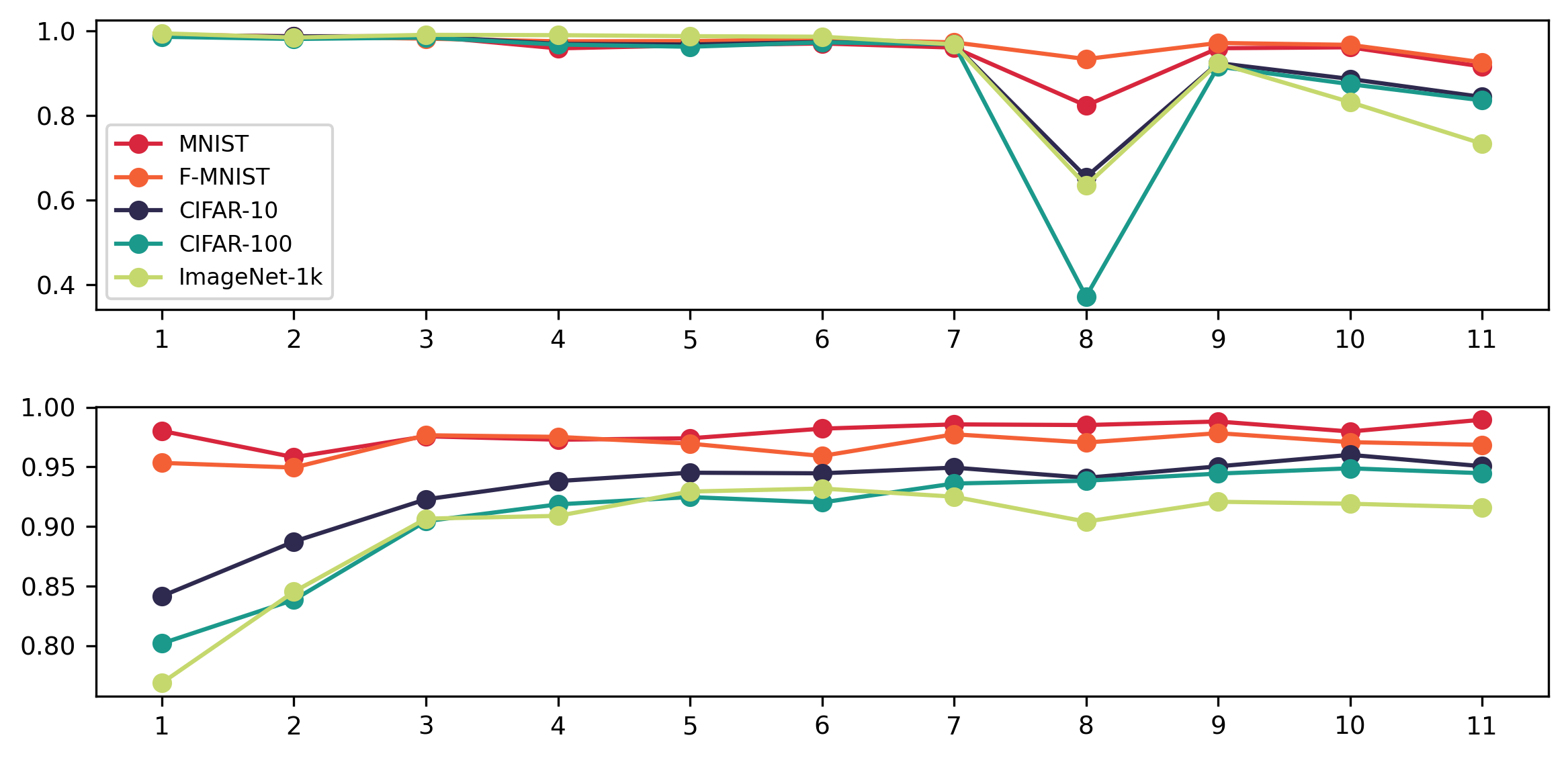}
            \put(-3,31){\rotatebox{90}{\dinob}}
            \put(-3,8){\rotatebox{90}{\deits}}
        \end{overpic}
        \caption{\textbf{Approximation vs. Representation Similarity}. \gls{cka} similarity between the last block representations of the original and the approximated model when approximating the $i^{\text{th}}$ block.}
        \label{fig:cka-vs-lastlayer}
    \end{subfigure}
    \hfill
    \begin{subfigure}{0.4\textwidth}
        \centering
        \begin{overpic}[width=\textwidth]{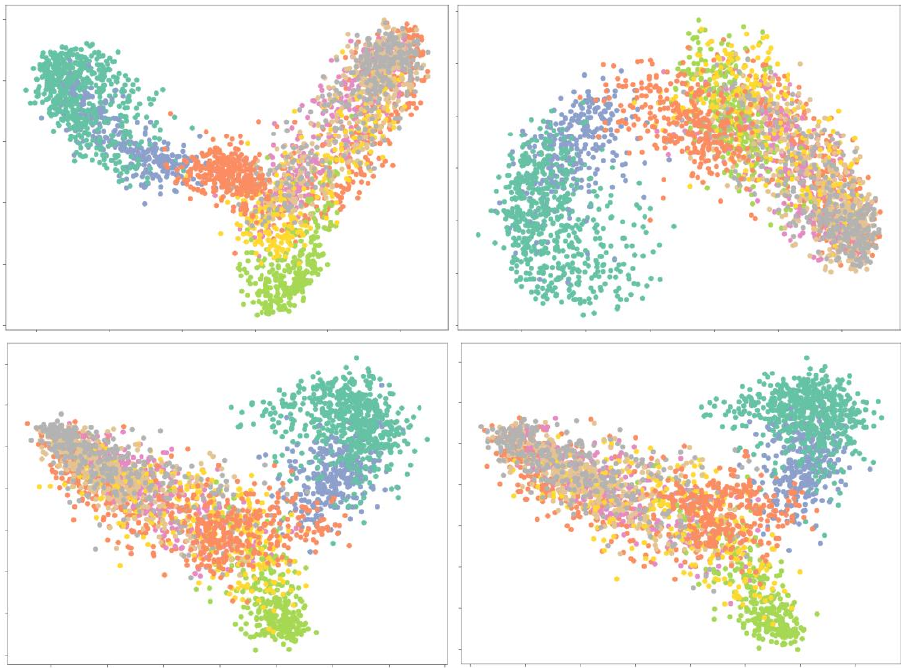}
            \put(14, 75){\texttt{Original}}
            \put(67, 75){\texttt{TOAST}}
            \put(-4,47){\rotatebox{90}{\dinob}}
            \put(-4,10){\rotatebox{90}{\deits}}
        \end{overpic}
        \caption{\textbf{PCA Visualization}. Final block representations for the original and \gls{toast} models on \fmnist{}.}
        \label{fig:pca-approx-main}
    \end{subfigure} 
\end{figure}

\paragraph{How does TOAST affect latent representations?} We next analyze the impact of the proposed transformations on the final block's latent representations, which are used for downstream classification.
We approximate these blocks using a shared linear transformation applied across all tokens, estimated on a subset of 500 training samples. For consistency, we use the same models and datasets as in \Cref{fig:latent-analysis-cka}.
To quantify the effect of the approximation, following \citep{venkataramanan2024skipattention,kornblith2019similarity} we compute the \gls{cka} similarity between the final block representations of the original and the \gls{toast}-approximated model for each block $k$ using its preceding block as input. As shown in \Cref{fig:cka-vs-lastlayer}, the model-specific similarity patterns re-emerge after approximation.
The plots highlight more specific trends. Approximating blocks is easier on simpler tasks (e.g., image classification on \mnist{} or \fmnist{}), yielding representations that closely match the originals, whereas on more complex datasets (e.g., \imagenet{} or \cifarh{}), the approximated representations deviate more from the original ones. Furthermore, the final blocks of \deits{} exhibit high similarity, suggesting that approximating these layers preserves the final representations, while earlier blocks remain more critical.
To provide a more intuitive view, \Cref{fig:pca-approx-main} visualizes the final-layer representations using \gls{pca}.
We compare the original representations with those obtained after approximating the final block (10 $\rightarrow$ 11) using \gls{toast} on \fmnist{}, with colors indicating the 10 classes. The visualization confirms that approximating the final block of \dinob{} results in noticeable deviations from the original representations, whereas for \deits{} the approximated representations remain highly similar. These observations align with the \gls{cka} analysis in \Cref{fig:cka-vs-lastlayer}, highlighting that the effect of block approximation depends strongly on the model and its internal block structure. Additional results across other models and datasets are provided in \Cref{sec:app-similarities}.\looseness=-1

\begin{tcolorbox}[colframe=takeaway, colback=takeaway, opacityback=1.0, boxrule=0mm, arc=2mm]
    \textbf{Takeaway} Transformer blocks can be approximated using simple transformations, without compromising representation fidelity.
\end{tcolorbox}

\paragraph{Which blocks are most suitable for linear replacement?} We conduct a systematic layer-wise sensitivity analysis across six models spanning three families (\vits{}, \vitb{}, \deits{}, \deitb{}, \dinos{}, \dinob{}). For each model, we replace one block at a time with a linear translator and measure the accuracy drop relative to the unmodified baseline on both \cifarhf{} and \imagenet{}. Additional results, including inference-only results, are detailed in \Cref{sec:app-layer-sensitivity}.
\begin{figure}[h]
    \centering
    \begin{overpic}[width=.7\textwidth]{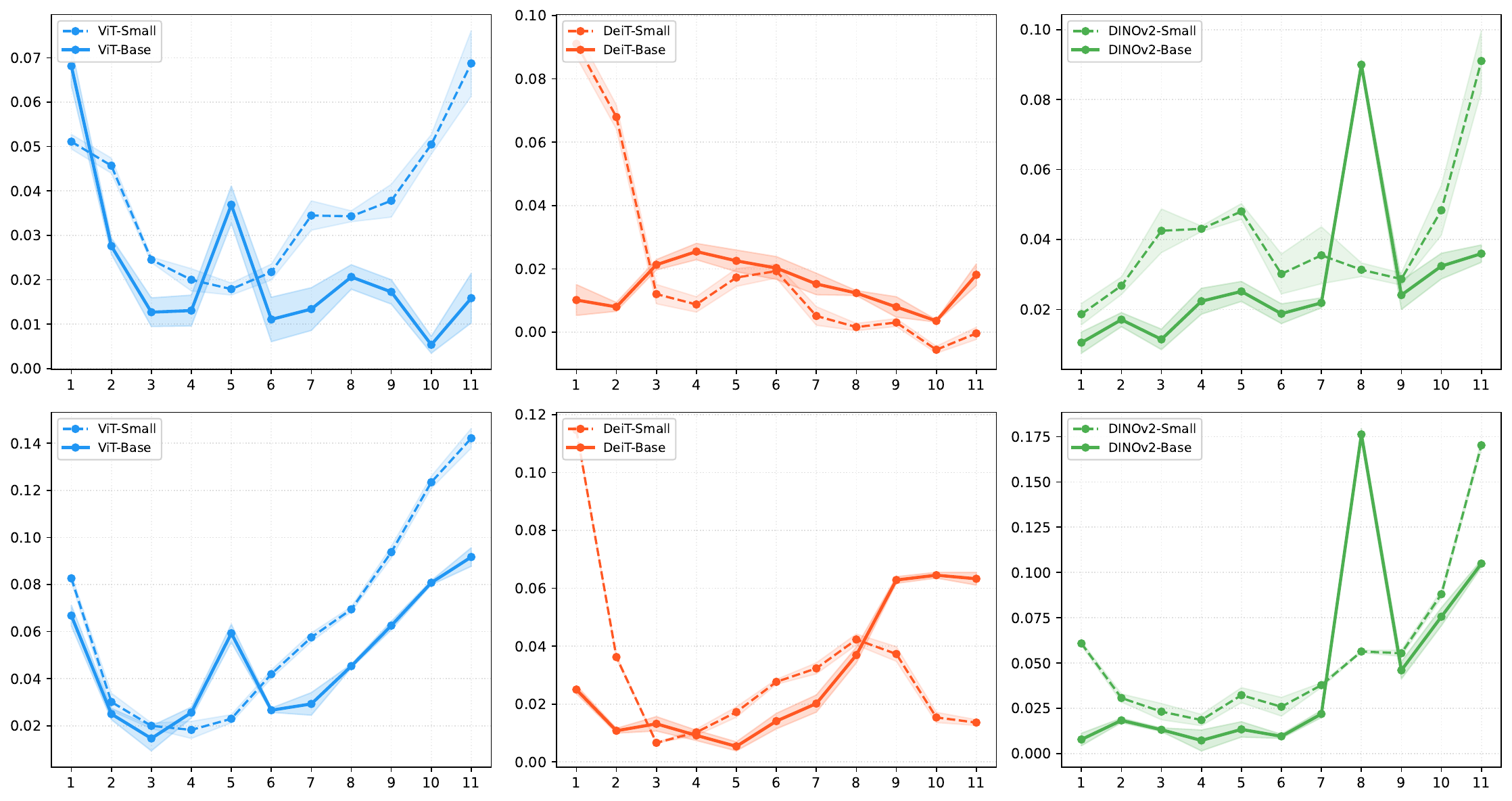}
        \put(-2,32){\rotatebox{90}{\cifarhf{}}}
        \put(-2,05){\rotatebox{90}{\imagenet{}}} 
        \put(-7,17){\rotatebox{90}{Accuracy drop ($\uparrow$)}}
    \end{overpic}
    \caption{\textbf{Layer-wise Sensitivity.}
        Accuracy drop ($\uparrow$ = worse) when replacing each individual block with a linear translator fitted on 500 samples, for \cifarhf{} (top) and \imagenet{} (bottom), grouped by model family. Solid lines = base models; dashed lines = small models. Results averaged over 3 seeds.}
    \label{fig:layer-skip-delta}
\end{figure}
As shown in \Cref{fig:layer-skip-delta} (and previously in \Cref{fig:latent-analysis-cka}), sensitivity profiles are strongly architecture-dependent, but their shapes are remarkably consistent regardless of the dataset. For instance, \deit{} models are consistently most sensitive at the earliest blocks (layer~1 of \deits{}: $+9.11\%$ on \cifarhf{}, $+11.47\%$ on \imagenet{}) and easiest at the latest. Conversely, \dino{} models always remain easiest to linearize at early-to-middle blocks and hardest at late blocks (\dinos{} layer~11: $+9.10\%$/$+17.02\%$). \vit{} models, meanwhile, show a U-shaped profile across datasets, with the minimum in the middle (e.g.\ \vitb{} layer~3: $+1.46\%$). These results confirm the analysis in \Cref{fig:cka-vs-lastlayer,fig:pca-approx-main}.
Despite these differences, three conclusions hold across all models: (i) Block replaceability is an intrinsic property of the architecture rather than the data, as easiest/hardest blocks are preserved across datasets and evaluation protocols (linear probe vs.\ frozen head). (ii) Bigger models tolerate block skipping better than small models at nearly every position (e.g.\ worst-case on \imagenet{}: $9.17\%$ for base vs.\ $17.02\%$ for small, excluding the \dinob{} layer-8 outlier). (iii) No fixed layer position is universally safe to skip, which directly motivates the data-driven heuristic introduced in \Cref{sec:method} for automatically identifying redundant blocks.

\begin{tcolorbox}[colframe=takeaway, colback=takeaway, opacityback=1.0, boxrule=0mm, arc=2mm]
    \textbf{Takeaway} Block replaceability is heavily architecture-dependent and cannot be determined a priori, yet its patterns remain remarkably consistent across datasets.
\end{tcolorbox}

\paragraph{Can entire transformer blocks be approximated without losing accuracy?} Initial results, reported in \Cref{table:classification-results-preliminary}, support the qualitative analysis and empirically demonstrate that \emph{entire vision transformer blocks} can be effectively approximated using simpler transformations (e.g., linear projections or, in some cases, the identity function). Such approximations reduce both the number of parameters and \gls{gflops}, thereby improving throughput (images per second), while incurring only a slight to negligible decrease in downstream task performance.
For instance, consistent with our earlier analysis, we find that approximating the final block of \deits{} when using \imagenet{} (e.g., approximating blocks 10 $\rightarrow$ 11 or 9 $\rightarrow$ 11 with a linear transformation) yields modest performance drops going from 73.85\% to 73.78\% and 70.01\%, respectively, while providing substantial efficiency gains. Importantly, we also show that even the identity transformation achieves competitive results, with accuracy drops as small as -0.24\% and -5.44\%, respectively.
However, the choice of translator naturally depends on the efficiency-accuracy trade-off: the linear transformation generally guarantees the most reliable accuracy–efficiency balance, whereas the identity yields the leanest training-free approximation when maximum simplicity is required.
Further methodological details and the full evaluation are presented in \Cref{sec:app-implementation-details} and \Cref{sec:classification}, respectively, while details on the efficiency metrics and additional analysis on those are reported in \Cref{sec:app-efficiency,app:acc-eff-tradeoff}.

\begin{table}[h!]
    \centering
    \caption{\textbf{\gls{toast} Image Classification Performance}. Performance comparison using the Identity translator and the Linear Translator for \deits{} and \imagenet{} across 3 seeds. The ``Approx.'' column specifies the blocks used for approximation, where the first one represents the block whose output is used to approximate the second block's output. Additional results in \Cref{table:classification-results-imagenet,table:classification-results-cifar100,sec:app-image-classification}.}
    \label{table:classification-results-preliminary}
    \resizebox{\textwidth}{!}{%
        \begin{tabular}{lccccccccc}
            \toprule
                                 &                     &                 & \multicolumn{3}{c}{Identity Translator} & \multicolumn{3}{c}{Linear Translator}                                                                                                     \\
            \cmidrule(l){4-6} \cmidrule(l){7-9}
                                 & Approx.             & Params.         & Accuracy \% $\uparrow$                  & GFLOPS $\downarrow$                   & imgs/s $\uparrow$ & Accuracy \% $\uparrow$              & GFLOPS $\downarrow$ & imgs/s $\uparrow$ \\
            \midrule
            \multirow[c]{6}{*}{} & 2 $\rightarrow$ 4   & \texttt{-3.25M} & $63.74 \pm 0.19 (-13.69\%)$             & $4.15$                                & $7222.5$          & $69.87 \pm 0.14 (-5.39\%)$          & $4.18$              & $7187.6$          \\
                                 & 9 $\rightarrow$ 11  & \texttt{-3.25M} & $69.83 \pm 0.33 (-5.44\%)$              & $4.15$                                & $7224.6$          & $70.01 \pm 0.27 (-5.20\%)$          & $4.18$              & $7203.8$          \\
            \midrule
                                 & 0 $\rightarrow$ 1   & \texttt{-1.62M} & $64.02 \pm 0.08 (-13.31\%)$             & $4.56$                                & $6755.8$          & $62.32 \pm 0.15 (-15.61\%)$         & $4.59$              & $6748.9$          \\
                                 & 10 $\rightarrow$ 11 & \texttt{-1.62M} & $\mathbf{73.67 \pm 0.26 (-0.24\%)}$     & $4.56$                                & $6751.7$          & $\mathbf{73.78 \pm 0.28 (-0.10\%)}$ & $4.59$              & $6756.3$          \\
            \cmidrule(l){2-9}
                                 & original            & \texttt{21.81M} & $73.85 \pm 0.39$                        & 4.97                                  & 6349.2            & $73.85 \pm 0.39$                    & 4.97                & 6325.6            \\
            \bottomrule
        \end{tabular}
    }
\end{table}

\begin{tcolorbox}[colframe=takeaway, colback=takeaway, opacityback=1.0, boxrule=0mm, arc=2mm]
    \textbf{Takeaway} \gls{toast} effectively reduces model parameters and improves model efficiency without significantly compromising the downstream task performance.
\end{tcolorbox}

\subsection{Image Classification Performance} \label{sec:classification}

We evaluate \gls{toast} on image classification tasks using pretrained models of varying sizes (\vitl{}, \dinob{}, and \deits{}) and two benchmark datasets (\cifarhf{} and \imagenet{}). 
Additional results with a broader set of models (\vitt{}, \vits{}, \vitb{}, \vitl{}, \dinos{}, \dinob{}, \deits{}) and datasets (\mnist{}, \fmnist{}, \cifart{}, \cifarhc{}) are provided in \Cref{sec:app-image-classification}.
In \Cref{app:misclassification-analysis}, we complement the quantitative evaluations with qualitative analyses of misclassifications after block approximation, providing further insight into model behavior under \gls{toast}.
Additional implementation details, including model and dataset specifications, computational resources, and software tools, are provided in \Cref{tab:pretrained-info,tab:dataset-info} and in \Cref{sec:app-computation,sec:app-efficiency,sec:app-tools}.
Block approximations in \gls{toast} are calculated via a shared linear, or identity, transformation applied across all tokens and are estimated using a subset of 500 training samples. A linear classifier is then trained on top of the frozen backbone with the Adam optimizer (learning rate $0.001$), batch size 256, for 5 epochs, over 3 different seeds. 
This setup simulates a realistic scenario where a pretrained feature extractor is adapted to a new dataset unseen during pretraining. 
However, to assess the robustness of our method, we also report the results using the original classification heads (\Cref{app:original-heads}), which confirm the consistency of our findings.

\begin{table}[h]
    \centering
    \caption{\textbf{\gls{toast} Classification Performance on \imagenet{}}. Image classification accuracy, \gls{gflops}, and throughput for \deits{}, \dinob{}, and \vitl{} using \imagenet{}. The ``Approx.'' column indicates the block pairs where the first block approximates the second. Additional results using other models and datasets are provided in \Cref{table:classification-results-cifar100,sec:app-image-classification}.}
    \label{table:classification-results-imagenet}
    \resizebox{.95\textwidth}{!}{%
        \begin{tabular}{ccccccccc}
            \toprule
                      &           &         & \multicolumn{3}{c}{Identity} & \multicolumn{3}{c}{Linear} \\
            \cmidrule(l){4-6} \cmidrule(l){7-9}
                      & Approx.   & Params. & Accuracy \% $\uparrow$ & GFLOPs $\downarrow$ & imgs/s $\uparrow$ & Accuracy \% $\uparrow$ & GFLOPs $\downarrow$ & imgs/s $\uparrow$ \\
            \midrule
            \multirow[c]{5}{*}{\rotatebox[origin=c]{90}{\deits{}}} & 3 $\rightarrow$ 4, 9 $\rightarrow$ 11 & \texttt{-4.88M} & $66.96 \pm 0.34 (-9.33\%)$ & 3.74 & 7751.4 & $68.39 \pm 0.13 (-7.39\%)$ & 3.80 & 7718.4 \\
                      & 3 $\rightarrow$ 4, 9 $\rightarrow$ 10 & \texttt{-3.25M} & $69.22 \pm 0.13 (-6.27\%)$ & 4.15 & 7210.9 & $71.35 \pm 0.22 (-3.38\%)$ & 4.21 & 7188.4 \\
                      & 2 $\rightarrow$ 3 & \texttt{-1.62M} & $70.80 \pm 0.05 (-4.12\%)$ & 4.56 & 6754.2 & $73.19 \pm 0.19 (-0.88\%)$ & 4.59 & 6736.7 \\
                      & 10 $\rightarrow$ 11 & \texttt{-1.62M} & $\mathbf{73.67 \pm 0.26 (-0.24\%)}$ & 4.56 & 6752.6 & $\mathbf{73.78 \pm 0.28 (-0.09\%)}$ & 4.59 & 6740.5 \\
            \cmidrule(l){2-9}
                      & original & \texttt{21.81M} & $73.85 \pm 0.39$ & 4.97 & 6349.2 & $73.85 \pm 0.39$ & 4.97 & 6325.6 \\
            \midrule
            \multirow[c]{6}{*}{\rotatebox[origin=c]{90}{\dinob{}}} & 0 $\rightarrow$ 4 & \texttt{-26M} & $3.58 \pm 0.06 (-95.20\%)$ & 16.32 & 3230.9 & $27.70 \pm 0.19 (-62.71\%)$ & 16.47 & 3227.7 \\
                      & 0 $\rightarrow$ 1, 2 $\rightarrow$ 3, 4 $\rightarrow$ 5 & \texttt{-19.5M} & $6.98 \pm 0.18 (-90.63\%)$ & 18.34 & 2947.0 & $61.02 \pm 0.36 (-17.86\%)$ & 18.80 & 2929.6 \\
                      & 0 $\rightarrow$ 1, 2 $\rightarrow$ 3 & \texttt{-13M} & $13.28 \pm 0.46 (-82.18\%)$ & 20.37 & 2703.9 & $70.82 \pm 0.49 (-4.66\%)$ & 20.67 & 2681.2 \\
                      & 0 $\rightarrow$ 1 & \texttt{-6.5M} & $\mathbf{65.47 \pm 0.43 (-12.14\%)}$ & 22.39 & 2506.6 & $\mathbf{73.43 \pm 0.02 (-1.15\%)}$ & 22.54 & 2487.0 \\
                      & 5 $\rightarrow$ 6 & \texttt{-6.5M} & $28.84 \pm 0.51 (-61.30\%)$ & 22.39 & 2503.1 & $73.01 \pm 0.41 (-1.71\%)$ & 22.54 & 2490.6 \\
            \cmidrule(l){2-9}
                      & original & \texttt{86.58M} & $74.52 \pm 0.26$ & 24.42 & 2321.3 & $74.52 \pm 0.26$ & 24.42 & 2316.5 \\
            \midrule
            \multirow[c]{10}{*}{\rotatebox[origin=c]{90}{\vitl{}}} & 2 $\rightarrow$ 4, 18 $\rightarrow$ 23 & \texttt{-80.83M} & $62.92 \pm 0.21 (-19.89\%)$ & 45.05 & 1654.9 & $67.43 \pm 0.05 (-14.16\%)$ & 45.47 & 1652.8 \\
                      & 17 $\rightarrow$ 23 & \texttt{-69.28M} & $66.81 \pm 0.34 (-14.95\%)$ & 47.70 & 1572.4 & $66.87 \pm 0.52 (-14.87\%)$ & 47.90 & 1567.0 \\
                      & 3 $\rightarrow$ 4, 19 $\rightarrow$ 23 & \texttt{-57.74M} & $70.97 \pm 0.42 (-9.65\%)$ & 50.34 & 1509.9 & $71.50 \pm 0.14 (-8.98\%)$ & 50.75 & 1499.5 \\
                      & 3 $\rightarrow$ 4, 20 $\rightarrow$ 23 & \texttt{-46.19M} & $73.49 \pm 0.18 (-6.44\%)$ & 52.98 & 1440.4 & $74.03 \pm 0.43 (-5.76\%)$ & 53.39 & 1436.8 \\
                      & 3 $\rightarrow$ 4, 21 $\rightarrow$ 23 & \texttt{-34.64M} & $75.80 \pm 0.26 (-3.50\%)$ & 55.62 & 1377.2 & $76.30 \pm 0.14 (-2.86\%)$ & 56.03 & 1345.6 \\
                      & 7 $\rightarrow$ 8, 15 $\rightarrow$ 16 & \texttt{-23.09M} & $76.81 \pm 0.28 (-2.21\%)$ & 58.26 & 1318.2 & $77.32 \pm 0.48 (-1.56\%)$ & 58.67 & 1316.4 \\
                      & 16 $\rightarrow$ 17, 22 $\rightarrow$ 23 & \texttt{-23.09M} & $77.64 \pm 0.32 (-1.16\%)$ & 58.26 & 1318.8 & $77.64 \pm 0.02 (-1.16\%)$ & 58.67 & 1312.3 \\
                      & 3 $\rightarrow$ 4 & \texttt{-11.55M} & $77.32 \pm 0.29 (-1.57\%)$ & 60.90 & 1269.2 & $\mathbf{78.36 \pm 0.26 (-0.24\%)}$ & 61.11 & 1270.0 \\
                      & 22 $\rightarrow$ 23 & \texttt{-11.55M} & $\mathbf{78.32 \pm 0.09 (-0.29\%)}$ & 60.90 & 1267.5 & $78.21 \pm 0.19 (-0.43\%)$ & 61.11 & 1270.9 \\
            \cmidrule(l){2-9}
                      & original & \texttt{304.35M} & $78.55 \pm 0.20$ & 63.54 & 1219.8 & $78.55 \pm 0.20$ & 63.54 & 1225.2 \\
            \bottomrule
        \end{tabular}
    }
\end{table}

\begin{table}[h]
    \centering
    \caption{\textbf{\gls{toast} Classification Performance on \cifarhf{}}. Image classification accuracy, \gls{gflops}, and throughput for \deits{}, \dinob{}, and \vitl{} using \cifarhf{}. The ``Approx.'' column indicates the block pairs where the first block approximates the second. Additional results using other models and datasets are provided in \Cref{sec:app-image-classification}.}
    \label{table:classification-results-cifar100}
    \resizebox{.95\textwidth}{!}{%
        \begin{tabular}{ccccccccc}
            \toprule
                                                                   &                                                         &                  & \multicolumn{3}{c}{Identity}        & \multicolumn{3}{c}{Linear}                                                                                                     \\
            \cmidrule(l){4-6} \cmidrule(l){7-9}
                                                                   & Approx.                                                 & Params.          & Accuracy \% $\uparrow$              & GFLOPs $\downarrow$        & imgs/s $\uparrow$ & Accuracy \% $\uparrow$              & GFLOPs $\downarrow$ & imgs/s $\uparrow$ \\
            \midrule
            \multirow[c]{5}{*}{\rotatebox[origin=c]{90}{\deits{}}} & 3 $\rightarrow$ 4, 9 $\rightarrow$ 11                   & \texttt{-4.88M}  & $68.48 \pm 0.34 (-3.44\%)$          & 3.74                       & 7755.1            & $70.64 \pm 0.37 (-0.39\%)$          & 3.80                & 7713.7            \\
                                                                   & 9 $\rightarrow$ 11                                      & \texttt{-3.25M}  & $\mathbf{72.28 \pm 0.36 (+1.92\%)}$ & 4.15                       & 7226.6            & $\mathbf{72.04 \pm 0.42 (+1.57\%)}$ & 4.18                & 6791.7            \\
                                                                   & 8 $\rightarrow$ 9                                       & \texttt{-1.62M}  & $71.34 \pm 0.10 (+0.60\%)$          & 4.56                       & 6755.2            & $70.80 \pm 0.12 (-0.17\%)$          & 4.59                & 6739.9            \\
                                                                   & 9 $\rightarrow$ 10                                      & \texttt{-1.62M}  & $71.66 \pm 0.39 (+1.04\%)$          & 4.56                       & 6692.1            & $71.49 \pm 0.20 (+0.80\%)$          & 4.59                & 6741.3            \\
            \cmidrule(l){2-9}
                                                                   & original                                                & \texttt{21.81M}  & $70.92 \pm 0.18$                    & 4.97                       & 6349.0            & $70.92 \pm 0.18$                    & 4.97                & 6249.4            \\
            \midrule
            \multirow[c]{6}{*}{\rotatebox[origin=c]{90}{\dinob{}}} & 0 $\rightarrow$ 4                                       & \texttt{-26M} & $18.29 \pm 0.86 (-79.09\%)$         & 16.32                      & 3233.8            & $62.25 \pm 0.54 (-28.83\%)$         & 16.47               & 3204.9            \\
                                                                   & 0 $\rightarrow$ 1, 2 $\rightarrow$ 3, 4 $\rightarrow$ 5 & \texttt{-19.5M} & $29.05 \pm 0.31 (-66.79\%)$         & 18.34                      & 2943.1            & $79.06 \pm 0.27 (-9.60\%)$          & 18.80               & 2922.6            \\
                                                                   & 0 $\rightarrow$ 1, 2 $\rightarrow$ 3                    & \texttt{-13M} & $33.25 \pm 0.18 (-61.99\%)$         & 20.37                      & 2705.6            & $84.18 \pm 0.18 (-3.76\%)$          & 20.67               & 2690.1            \\
                                                                   & 0 $\rightarrow$ 1                                       & \texttt{-6.5M}  & $\mathbf{78.83 \pm 0.22 (-9.87\%)}$ & 22.39                      & 2492.8            & $\mathbf{86.64 \pm 0.37 (-0.94\%)}$ & 22.54               & 2493.8            \\
                                                                   & 2 $\rightarrow$ 3                                       & \texttt{-6.5M}  & $47.51 \pm 0.52 (-45.68\%)$         & 22.39                      & 2484.2            & $86.06 \pm 0.20 (-1.60\%)$          & 22.54               & 2484.6            \\
            \cmidrule(l){2-9}
                                                                   & original                                                & \texttt{86.58M}  & $87.46 \pm 0.04$                    & 24.42                      & 2315.5            & $87.46 \pm 0.04$                    & 24.42               & 2317.3            \\
            \midrule
            \multirow[c]{11}{*}{\rotatebox[origin=c]{90}{\vitl{}}} & 2 $\rightarrow$ 4, 18 $\rightarrow$ 23                  & \texttt{-80.83M} & $74.41 \pm 0.44 (-13.79\%)$         & 45.05                      & 1655.7            & $84.02 \pm 0.39 (-2.66\%)$          & 45.47               & 1649.6            \\
                                                                   & 17 $\rightarrow$ 23                                     & \texttt{-69.28M} & $85.32 \pm 0.45 (-1.16\%)$          & 47.70                      & 1578.8            & $84.55 \pm 0.44 (-2.05\%)$          & 47.90               & 1552.1            \\
                                                                   & 3 $\rightarrow$ 4, 19 $\rightarrow$ 23                  & \texttt{-57.74M} & $84.23 \pm 0.08 (-2.43\%)$          & 50.34                      & 1503.6            & $85.81 \pm 0.39 (-0.59\%)$          & 50.75               & 1497.4            \\
                                                                   & 3 $\rightarrow$ 4, 20 $\rightarrow$ 23                  & \texttt{-46.19M} & $84.68 \pm 0.18 (-1.90\%)$          & 52.98                      & 1445.2            & $86.30 \pm 0.11 (-0.03\%)$          & 53.39               & 1431.0            \\
                                                                   & 20 $\rightarrow$ 23                                     & \texttt{-34.64M} & $\mathbf{86.61 \pm 0.07 (+0.33\%)}$ & 55.62                      & 1381.2            & $86.55 \pm 0.22 (+0.27\%)$          & 55.82               & 1372.6            \\
                                                                   & 3 $\rightarrow$ 4, 21 $\rightarrow$ 23                  & \texttt{-34.64M} & $84.86 \pm 0.28 (-1.70\%)$          & 55.62                      & 1376.7            & $86.37 \pm 0.28 (+0.06\%)$          & 56.03               & 1372.7            \\
                                                                   & 20 $\rightarrow$ 22                                     & \texttt{-23.09M} & $86.30 \pm 0.23 (-0.03\%)$          & 58.26                      & 1317.5            & $86.52 \pm 0.12 (+0.24\%)$          & 58.47               & 1314.6            \\
                                                                   & 3 $\rightarrow$ 4, 21 $\rightarrow$ 22                  & \texttt{-23.09M} & $84.58 \pm 0.19 (-2.02\%)$          & 58.26                      & 1315.8            & $86.20 \pm 0.11 (-0.14\%)$          & 58.67               & 1317.6            \\
                                                                   & 20 $\rightarrow$ 21                                     & \texttt{-11.55M} & $86.44 \pm 0.24 (+0.14\%)$          & 60.90                      & 1268.5            & $86.39 \pm 0.08 (+0.08\%)$          & 61.11               & 1266.7            \\
                                                                   & 21 $\rightarrow$ 22                                     & \texttt{-11.55M} & $86.55 \pm 0.01 (+0.26\%)$          & 60.90                      & 1270.7            & $\mathbf{86.72 \pm 0.24 (+0.46\%)}$ & 61.11               & 1269.2            \\
            \cmidrule(l){2-9}
                                                                   & original                                                & \texttt{304.35M} & $86.32 \pm 0.08$                    & 63.54                      & 1223.1            & $86.32 \pm 0.08$                    & 63.54               & 1224.3            \\
            \bottomrule
        \end{tabular}
    }
\end{table}

\paragraph{Are \gls{toast} results competitive?} As shown in \Cref{table:classification-results-cifar100,table:classification-results-imagenet}, \gls{toast} consistently reduces model size and \gls{gflops} while maintaining, and in some cases improving, image classification accuracy. This aligns with our representational analyses in \Cref{sec:latent-analysis}: for instance, approximating the final block of \deits{} produces latent representations nearly identical to the original (\Cref{fig:cka-vs-lastlayer,fig:pca-approx-main}), making it an ideal candidate for approximation.  
Even when multiple consecutive blocks are approximated (e.g., 9$\rightarrow$ 11), models maintain performance comparable to or exceeding the original while significantly reducing parameters. This demonstrates that a simple linear transformation, or even the identity in certain cases, is sufficient to capture the functionality of full transformer blocks without additional training, provided the transformation is shared across all tokens.  
Additionally, efficiency gains are notable: throughput (imgs/s) increases while \gls{gflops} decreases, highlighting practical benefits for deployment, as also shown in \Cref{app:acc-eff-tradeoff}. Additional results across other models (\dinob{}, \vitl{}) and datasets confirm that \gls{toast} generalizes across architectures and scales (\Cref{sec:app-image-classification}). Finally, while approximations are easier for simpler datasets (e.g., \cifarhf{}), \gls{toast} still achieves meaningful compression with minimal accuracy loss on complex datasets like \imagenet{}.
Additional results across models and datasets are provided in \Cref{table:app-vit-s-classification,table:app-dino-s-classification,table:app-vit-t-classification,table:app-vitb-classification-results}. To assess scalability, we applied \gls{toast} to \vitl{}. Approximating selected blocks, e.g., 17 $\rightarrow$ 23, reduces the parameter count by 69.3M, lowers \gls{gflops} from 63.54 to 47.70, and increases throughput from 1223.1 to 1578.8 imgs/s, while incurring a minimal accuracy drop of 1.16\%. This shows \gls{toast}'s utility in balancing substantial computational savings with a modest performance trade-off, even in large models.

\begin{tcolorbox}[colframe=border, colback=takeaway, opacityback=0.95, boxrule=0mm]
\textbf{Takeaway} Approximating selected blocks enables efficiency gains with minimal impact on the accuracy.
\end{tcolorbox}

\begin{figure}[h]
    \centering  
    \begin{overpic}[width=.9\textwidth]{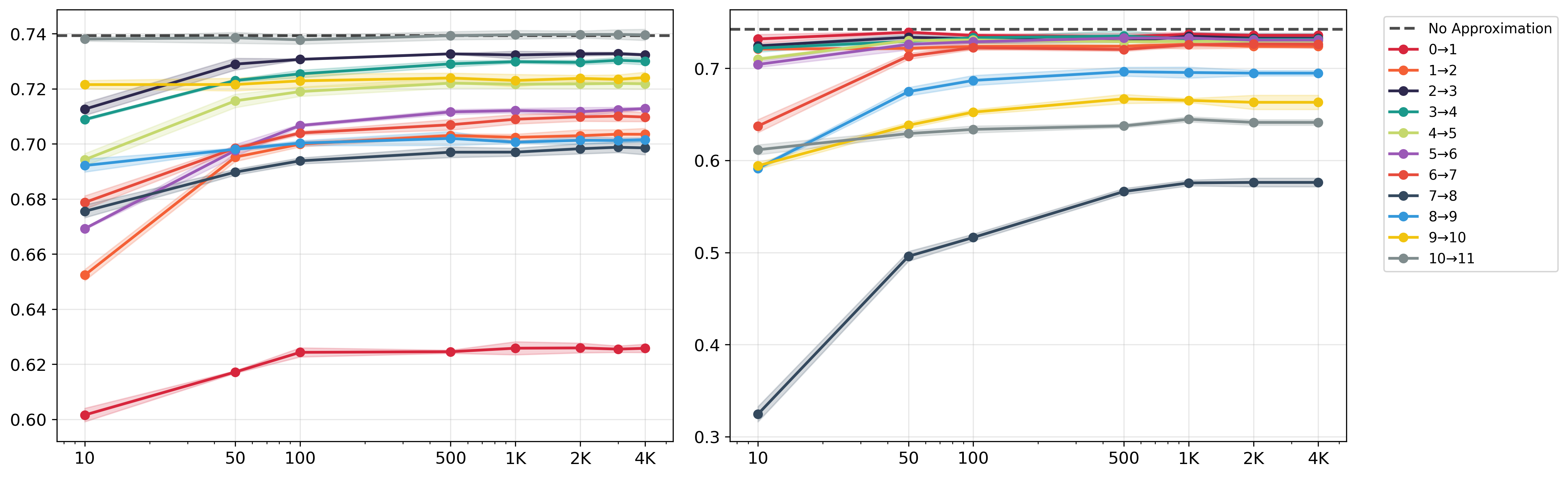}
        \put(-2.5,9){\rotatebox{90}{\texttt{Accuracy ($\uparrow$)}}} 
        \put(20,31){\deits{}}
        \put(63,31){\dinob{}}
    \end{overpic}
\caption{\textbf{Sample Size Ablation}. Classification accuracy as a function of the number of training samples used for approximating different layers of \dinob{} and \deits{} with a linear transformation using \imagenet{}. Accuracy stabilizes after approximately 500 samples.}
    \label{fig:num-samples-ablation}
\end{figure}

\paragraph{Are 500 training samples enough?} We study the sensitivity of block approximation to the number of training samples using \dinob{} and \deits{} on \imagenet{}. As shown in \Cref{fig:num-samples-ablation}, performance typically plateaus quickly: 500 samples are sufficient to obtain stable and reliable approximations. 
Increasing the sample count beyond this threshold provides only marginal gains, while substantially fewer samples lead to noticeable degradation.
Interestingly, when the representational spaces of consecutive blocks are already highly aligned, even as few as 10 or 50 samples suffice to achieve competitive approximations. Conversely, for blocks that are harder to approximate, such as the early layers of \deits{} (e.g., 0$\rightarrow$ 1), even 4000 samples are insufficient to estimate a linear transformation that maintains competitive performance.  
We highlight that these results are obtained on \imagenet{}, which contains 1000 classes. The 500 samples represent only a small subset of the class space, yet reliable approximations are still achieved. This indicates that \gls{toast} primarily captures the block-level structure of representations rather than requiring exhaustive coverage of all classes. Consequently, \gls{toast} could also be practical in scenarios where large labeled datasets are limited.
\begin{tcolorbox}[colframe=border, colback=takeaway, opacityback=0.95, boxrule=0mm]
\textbf{Takeaway} A small number of samples is sufficient to achieve stable and reliable representations when approximating transformer blocks, balancing efficiency and accuracy.
\end{tcolorbox}

\paragraph{What if a more complex transformation is used?} We evaluate whether deeper approximators improve downstream task performance.
Specifically, we compare \gls{toast} (Identity and Linear) to \gls{mlp} and Residual \gls{mlp}, trained for 300 steps with Adam (learning rate $10^{-3}$).
These more complex transformations, as for Identity and Linear, are applied across all tokens, and estimated using a subset of 500 training samples.
Results in \Cref{table:transformation-ablation} show a consistent trend for \vitl{} on both \imagenet{} and \cifarhf{}: the linear transformation provides the most reliable trade-off across datasets. 
\begin{table}[h]
    \centering
    \caption{\textbf{Transformations Comparison}. Classification accuracy on \cifarhf{} and \imagenet{} using \vitl{}. The ``Approx.'' column specifies the block mapping (output of the first block is used to approximate the output of the second). \gls{mlp} and Res-\gls{mlp} are trained approximators, while Identity and Linear are closed-form and training-free. Results are averaged over three seeds.}
    \label{table:transformation-ablation}
    \resizebox{.65\textwidth}{!}{%
        \begin{tabular}{ccccccc}
            \toprule
             &                                        &                  & \multicolumn{4}{c}{Accuracy $\uparrow$}                                                             \\
            \cmidrule(l){4-7}
             & Approx.                                & Params.          & Identity                                & Linear            & \gls{mlp}         & Res-\gls{mlp}     \\
            \midrule
            \multirow[c]{10}{*}{\rotatebox{90}{\cifarhf{}}}
             & 17 $\rightarrow$ 23                    & \texttt{-69.28M} & $85.32 \pm 0.45$                        & $84.55 \pm 0.44$  & $81.90 \pm 0.36$  & $84.04 \pm 0.46$  \\
             & 3 $\rightarrow$ 4, 19 $\rightarrow$ 23 & \texttt{-57.74M} & $84.23 \pm  0.08$                       & $85.81 \pm 0.39$  & $83.63 \pm 0.42$  & $85.58 \pm 0.06$  \\
            \cmidrule(l){2-7}
             & 3 $\rightarrow$ 4, 20 $\rightarrow$ 23 & \texttt{-46.19M} & $84.68 \pm  0.18$                       & $86.30 \pm  0.11$ & $84.36 \pm  0.48$ & $86.10 \pm  0.39$ \\
            \cmidrule(l){2-7}
             & 20 $\rightarrow$ 23                    & \texttt{-34.64M} & $86.61 \pm 0.07$                        & $86.55 \pm 0.22$  & $84.68 \pm 0.39$  & $86.19 \pm 0.02$  \\
             & 3 $\rightarrow$ 4, 21 $\rightarrow$ 23 & \texttt{-34.64M} & $84.86 \pm 0.28$                        & $86.37 \pm 0.28$  & $84.90 \pm 0.71$  & $86.10 \pm 0.37$  \\
            \cmidrule(l){2-7}
             & 20 $\rightarrow$ 22                    & \texttt{-23.09M} & $86.30 \pm 0.23$                        & $86.52 \pm 0.12$  & $84.97 \pm 0.18$  & $86.71 \pm 0.28$  \\
             & 3 $\rightarrow$ 4, 21 $\rightarrow$ 22 & \texttt{-23.09M} & $84.58 \pm 0.19$                        & $86.20 \pm 0.11$  & $84.83 \pm 0.31$  & $86.49 \pm 0.08$  \\
            \cmidrule(l){2-7}
             & 20 $\rightarrow$ 21                    & \texttt{-11.55M} & $86.44 \pm 0.24$                        & $86.39 \pm 0.08$  & $84.40 \pm 0.70$  & $86.63 \pm 0.06$  \\
             & 21 $\rightarrow$ 22                    & \texttt{-11.55M} & $86.55 \pm 0.01$                        & $86.72 \pm 0.24$  & $85.20 \pm 0.26$  & $86.82 \pm 0.31$  \\
            \cmidrule(l){2-7}
             & original                               & \texttt{304.35M} & $86.32 \pm 0.08$                        & $86.32 \pm 0.08$  & $86.32 \pm 0.08$  & $86.32 \pm 0.08$  \\
            \midrule
            \multirow[c]{10}{*}{\rotatebox{90}{\imagenet{}}}
             & 17 $\rightarrow$ 23                    & \texttt{-69.28M} & $66.81 \pm 0.34$                        & $66.87 \pm 0.52$  & $57.60 \pm 0.11$  & $65.04 \pm 0.12$  \\
             & 3 $\rightarrow$ 4, 19 $\rightarrow$ 23 & \texttt{-57.74M} & $70.97 \pm 0.42$                        & $71.50 \pm 0.14$  & $66.19 \pm 0.17$  & $70.75 \pm 0.07$  \\
            \cmidrule(l){2-7}
             & 3 $\rightarrow$ 4, 20 $\rightarrow$ 23 & \texttt{-46.19M} & $ 73.49 \pm 0.18$                       & $74.03 \pm  0.43$ & $69.49 \pm  0.24$ & $73.68 \pm  0.12$ \\
            \cmidrule(l){2-7}
             & 20 $\rightarrow$ 23                    & \texttt{-34.64M} & $74.45 \pm 0.07$                        & $74.45 \pm 0.24$  & $70.19 \pm 0.30$  & $74.46 \pm 0.22$  \\
             & 3 $\rightarrow$ 4, 21 $\rightarrow$ 23 & \texttt{-34.64M} & $75.80 \pm 0.26$                        & $76.30 \pm 0.14$  & $73.23 \pm 0.29$  & $76.14 \pm 0.22$  \\
            \cmidrule(l){2-7}
             & 20 $\rightarrow$ 22                    & \texttt{-23.09M} & $75.49 \pm 0.19$                        & $74.84 \pm 0.21$  & $70.56 \pm 0.25$  & $75.59 \pm 0.18$  \\
             & 3 $\rightarrow$ 4, 21 $\rightarrow$ 22 & \texttt{-23.09M} & $76.25 \pm 0.02$                        & $76.61 \pm 0.29$  & $73.52 \pm 0.40$  & $76.43 \pm 0.21$  \\
            \cmidrule(l){2-7}
             & 20 $\rightarrow$ 21                    & \texttt{-11.55M} & $77.00 \pm 0.27$                        & $77.19 \pm 0.25$  & $72.72 \pm 0.31$  & $76.24 \pm 0.21$  \\
             & 21 $\rightarrow$ 22                    & \texttt{-11.55M} & $77.24 \pm 0.28$                        & $77.06 \pm 0.24$  & $74.20 \pm 0.48$  & $77.14 \pm 0.27$  \\
            \cmidrule(l){2-7}
             & original                               & \texttt{304.35M} & $78.55 \pm 0.20$                        & $78.55 \pm 0.20$  & $78.55 \pm 0.20$  & $78.55 \pm 0.20$  \\
            \bottomrule
        \end{tabular}
    }
\end{table}
On \cifarhf{}, linear often achieves the best or near-best accuracy (e.g., 21$\rightarrow$ 22: $86.72\%$ vs. $86.82\%$ for Res-\gls{mlp} and $85.20\%$ for \gls{mlp}), while remaining training-free, thus more efficient. On \imagenet{}, the gap becomes even clearer: for the same blocks Identity reaches $77.24\%$, while Res-\gls{mlp} and \gls{mlp} reach $77.14\%$ and $74.20\%$, respectively. Linear ($77.06\%$) is also competitive, confirming that closed-form translators match deeper trained alternatives. \gls{toast} operates in closed form, requires no optimization, and consistently achieves strong efficiency–accuracy trade-offs. These findings confirm that a simple linear transformation is sufficient to approximate transformer blocks in most settings, with deeper translators offering little benefit despite their higher cost.

\begin{tcolorbox}[colframe=border, colback=takeaway, opacityback=0.95, boxrule=0mm]
\textbf{Takeaway} \gls{toast} consistently matches or outperforms deeper trained approximators while requiring no gradient-based training.
\end{tcolorbox}

\paragraph{What if you finetune the entire network instead of just the head?} We evaluate whether finetuning the entire network after applying \gls{toast} improves downstream task performance. Specifically, following the setting in \Cref{table:classification-results-cifar100}, we first apply \gls{toast} (Identity and Linear) using 500 samples. Subsequently, we finetune the entire network for 20 epochs with a batch size of 256 using Adam. We use a learning rate of $10^{-4}$ for \deits{}; for \dinob{}, we employ a discriminative learning rate to avoid catastrophic forgetting (setting the encoder learning rate to $10^{-6}$ and the classification head to $2\times10^{-4}$). We report results averaged over 3 random seeds. To ensure a direct comparison, we maintain the same block skip configurations used in \Cref{table:classification-results-cifar100}.
Results in \Cref{table:classification-e2e-results-cifar100} highlight that \gls{e2e} finetuning is critical for recovering performance when making aggressive structural approximations, whereas head-only finetuning often struggles in these scenarios. For example, with \deits{}, the approximation $3 \rightarrow 4, 9 \rightarrow 11$ causes a notable drop in head-only accuracy ($-3.44\%$); however, \gls{e2e} training allows the network to adapt, recovering performance to within $\sim1\%$ of the original model. This recovery capability is even more pronounced in \dinob{}. When approximating $0 \rightarrow 4$ (removing 26M parameters), head-only performance collapses (dropping nearly $80\%$ in accuracy); yet, \gls{e2e} finetuning successfully recovers the vast majority of this loss, achieving $89.54\%$ accuracy. Similarly, for the multi-block approximation $0 \rightarrow 1, 2 \rightarrow 3, 4 \rightarrow 5$ (removing 19.5M parameters), \gls{e2e} finetuning restores accuracy to $91.34\%$, a drop of only $1.63\%$ compared to the baseline. Complete results are reported in \Cref{table:classification-e2e-results-cifar100-complete}.
\begin{table}[h]
    \centering
    \caption{\textbf{E2E vs Head-only Finetuning Classification Performance.} Image classification accuracy when end-to-end finetuning the network vs only finetuning the head for \deits{}, \dinob{}, and \vitl{} using \cifarhf{}. The ``Approx.'' column indicates the block pairs where the first block approximates the second. Bold marks the best result among the rows shown in this table; the full ranking across all approximations is reported in \Cref{table:classification-results-cifar100}.}
    \label{table:classification-e2e-results-cifar100}
    \setlength{\tabcolsep}{2pt}
    \resizebox{.8\textwidth}{!}{
        \begin{tabular}{cccccccc}
            \toprule
             &                                           &                                                         &                                           & \multicolumn{2}{c}{Head-only finetuning (Acc \%)} & \multicolumn{2}{c}{End-to-end finetuning (Acc \%)}                                                                                                         \\
            \cmidrule(lr){5-6}
            \cmidrule(lr){7-8}

             &                                           & Approx.                                                 & Params.                                   & Identity                                          & Linear                                             & Identity                                          & Linear                                            \\
            \midrule
             & \multirow{4}{*}{\rotatebox{90}{\deits{}}}
             & 3 $\rightarrow$ 4, 9 $\rightarrow$ 11     & \texttt{-4.88M}                                         & $68.48 \pm 0.34 \scriptstyle{(-3.44\%)}$  & $70.64 \pm 0.37 \scriptstyle{(-0.39\%)}$          & $86.03 \pm 0.03 \scriptstyle{(-1.05\%)}$           & $85.99 \pm 0.00 \scriptstyle{(-1.09\%)}$                                                              \\
             &                                           & 9 $\rightarrow$ 11                                      & \texttt{-3.25M}                           & $72.28 \pm 0.36 \mathbf{\scriptstyle{(+1.92\%)}}$ & $72.04 \pm 0.42 \mathbf{\scriptstyle{(+1.57\%)}}$  & $86.81 \pm 0.01 \scriptstyle{(-0.15\%)}$          & $86.86 \pm 0.01 \scriptstyle{(-0.09\%)}$          \\
             &                                           & 9 $\rightarrow$ 10                                      & \texttt{-1.62M}                           & $71.66 \pm 0.39 \scriptstyle{(+1.04\%)}$          & $71.49 \pm 0.20 \scriptstyle{(+0.80\%)}$           & $87.01 \pm 0.02 \scriptstyle{(+0.08\%)}$          & $86.85 \pm 0.01 \scriptstyle{(-0.10\%)}$          \\
            \cmidrule(l){3-8}
             &                                           & original                                                & \texttt{21.81M}                           & $70.92 \pm 0.18$                                  & $70.92 \pm 0.18$                                   & $86.94 \pm 0.03$                                  & $86.94 \pm 0.03$                                  \\
            \cmidrule(l){2-8}

             & \multirow{5}{*}{\rotatebox{90}{\dinob{}}}
             & 0 $\rightarrow$ 4                         & \texttt{-26M}                                        & $18.29 \pm 0.86 \scriptstyle{(-79.09\%)}$ & $62.25 \pm 0.54 \scriptstyle{(-28.83\%)}$         & $56.15 \pm 0.01 \scriptstyle{(-39.52\%)}$          & $89.54 \pm 0.00 \scriptstyle{(-3.56\%)}$                                                              \\
             &                                           & 0 $\rightarrow$ 1, 2 $\rightarrow$ 3, 4 $\rightarrow$ 5 & \texttt{-19.5M}                          & $29.05 \pm 0.31 \scriptstyle{(-66.79\%)}$         & $79.06 \pm 0.27 \scriptstyle{(-9.60\%)}$           & $70.06 \pm 0.00 \scriptstyle{(-24.54\%)}$         & $91.34 \pm 0.00 \scriptstyle{(-1.63\%)}$          \\
             &                                           & 0 $\rightarrow$ 1, 2 $\rightarrow$ 3                    & \texttt{-13M}                          & $33.25 \pm 0.18 \scriptstyle{(-61.99\%)}$         & $84.18 \pm 0.18 \scriptstyle{(-3.76\%)}$           & $78.69 \pm 0.22 \scriptstyle{(-15.25\%)}$         & $92.34 \pm 0.00 \scriptstyle{(-0.55\%)}$          \\
             &                                           & 0 $\rightarrow$ 1                                       & \texttt{-6.5M}                           & $78.83 \pm 0.22 \mathbf{\scriptstyle{(-9.87\%)}}$ & $86.64 \pm 0.37 \mathbf{\scriptstyle{(-0.94\%)}}$  & $92.39 \pm 0.00 \mathbf{\scriptstyle{(-0.50\%)}}$ & $92.78 \pm 0.00 \mathbf{\scriptstyle{(-0.08\%)}}$ \\
            \cmidrule(l){3-8}
             &                                           & original                                                & \texttt{86.58M}                           & $87.46 \pm 0.04$                                  & $87.46 \pm 0.04$                                   & $92.85 \pm 0.00$                                  & $92.85 \pm 0.00$                                  \\
            \cmidrule(l){2-8}

             & \multirow{5}{*}{\rotatebox{90}{\vitl{}}}
             & 2 $\rightarrow$ 4, 18 $\rightarrow$ 23    & \texttt{-80.83M}                                        & $74.41 \pm 0.44 \scriptstyle{(-13.79\%)}$ & $84.02 \pm 0.39 \scriptstyle{(-2.66\%)}$          & $89.19 \pm 0.01 \scriptstyle{(-1.94\%)}$           & $88.73 \pm 0.00 \scriptstyle{(-2.44\%)}$                                                              \\
             &                                           & 17 $\rightarrow$ 23                                     & \texttt{-69.28M}                          & $85.32 \pm 0.45 \scriptstyle{(-1.16\%)}$          & $84.55 \pm 0.44 \scriptstyle{(-2.05\%)}$           & $90.73 \pm 0.00 \scriptstyle{(-0.24\%)}$          & $90.36 \pm 0.00 \scriptstyle{(-0.65\%)}$          \\
             &                                           & 3 $\rightarrow$ 4, 19 $\rightarrow$ 23                  & \texttt{-57.74M}                          & $84.23 \pm 0.08 \scriptstyle{(-2.43\%)}$          & $85.81 \pm 0.39 \scriptstyle{(-0.59\%)}$           & $90.34 \pm 0.00 \scriptstyle{(-0.67\%)}$          & $90.12 \pm 0.00 \scriptstyle{(-0.91\%)}$          \\
             &                                           & 3 $\rightarrow$ 4, 20 $\rightarrow$ 23                  & \texttt{-46.19M}                          & $84.68 \pm 0.18 \scriptstyle{(-1.90\%)}$          & $86.30 \pm 0.11 \scriptstyle{(-0.03\%)}$           & $90.44 \pm 0.00 \scriptstyle{(-0.56\%)}$          & $90.03 \pm 0.00 \scriptstyle{(-1.01\%)}$          \\
             &                                           & 20 $\rightarrow$ 23                                     & \texttt{-34.64M}                          & $86.61 \pm 0.07 \mathbf{\scriptstyle{(+0.33\%)}}$ & $86.55 \pm 0.22 \mathbf{\scriptstyle{(+0.27\%)}}$           & $91.04 \pm 0.00 \mathbf{\scriptstyle{(+0.10\%)}}$ & $90.74 \pm 0.00 \mathbf{\scriptstyle{(-0.23\%)}}$          \\
            \cmidrule(l){3-8}
             &                                           & original                                                & \texttt{304.35M}                          & $86.32 \pm 0.08$                                  & $86.32 \pm 0.08$                                   & $90.95 \pm 0.00$                                  & $90.95 \pm 0.00$                                  \\

            \bottomrule
        \end{tabular}
    }
\end{table}

\begin{tcolorbox}[colframe=border, colback=takeaway, opacityback=0.95, boxrule=0mm]
\textbf{Takeaway} When additional resources are available, \gls{toast} enables effective performance recovery.
\end{tcolorbox}
\subsection{TOAST Applicability to Other Tasks or Domains} \label{sec:add-experiments}
We further evaluate \gls{toast} beyond vision classification by applying it to text classification and semantic segmentation tasks. 
Additional implementation details, including model and dataset specifications, computational resources, and software tools, are provided in \Cref{tab:pretrained-info,tab:dataset-info} and \Cref{sec:app-computation,sec:app-efficiency,sec:app-tools}, with complete results in \Cref{app:text-classification}.
For text classification, we use \mbertb{} on the \agnews{} dataset,
\begin{wraptable}{r}{0.4\textwidth}
    \vspace{-0.2cm}
    \centering
    \caption{\textbf{Segmentation Performance.} mIoU results for each single configuration using \vits{} and \dinob{}.}
    \label{table:segmentation}
    \resizebox{.3\textwidth}{!}{%
        \begin{tabular}{ccccc}
        \toprule
                     & \multicolumn{4}{c}{mIoU $\uparrow$} \\
            \cmidrule(lr){2-5}
                     & \multicolumn{2}{c}{\vits{}} & \multicolumn{2}{c}{\dinob{}} \\
            \cmidrule(lr){2-3} \cmidrule(lr){4-5}
        Approx.      & Linear & \gls{mlp} & Linear & \gls{mlp} \\
            \midrule
            0 $\rightarrow$ 1   & 0.27 & 0.26 & 0.29 & 0.29 \\
            1 $\rightarrow$ 2   & 0.29 & 0.29 & 0.29 & 0.29 \\
            2 $\rightarrow$ 3   & 0.30 & 0.30 & 0.29 & 0.29 \\
            3 $\rightarrow$ 4   & 0.30 & 0.29 & 0.29 & 0.29 \\
            4 $\rightarrow$ 5   & 0.30 & 0.29 & 0.29 & 0.29 \\
            5 $\rightarrow$ 6   & 0.28 & 0.27 & 0.29 & 0.29 \\
            6 $\rightarrow$ 7   & 0.28 & 0.27 & 0.29 & 0.29 \\
            7 $\rightarrow$ 8   & 0.29 & 0.28 & 0.26 & 0.23 \\
            8 $\rightarrow$ 9   & 0.28 & 0.27 & 0.28 & 0.27 \\
            9 $\rightarrow$ 10  & 0.29 & 0.29 & 0.27 & 0.26 \\
            10 $\rightarrow$ 11 & 0.30 & 0.29 & 0.27 & 0.26 \\
            \midrule
            original            & \multicolumn{2}{c}{0.31} & \multicolumn{2}{c}{0.29} \\
            \bottomrule
        \end{tabular}
    }
\end{wraptable}
 while for segmentation we employ the same backbone on the \scene{} dataset. 
For both domains, we adopt the same setup as in the vision experiments: block approximations are implemented via a shared linear map, identity, or small MLP transformation applied across all tokens, estimated using a subset of 500 training samples.
In the text domain, a linear classifier is trained on top of the frozen backbone for 5 epochs over 3 seeds.
For segmentation, a segmentation head is trained on the frozen backbone for 10 epochs over 3 seeds. 
The results in \Cref{table:text-classification} show that, in this setting as well, the linear transformation outperforms the more complex MLP. Moreover, up to 10 blocks can be approximated (i.e., $11 \rightarrow 21$), substantially reducing GFLOPs, improving throughput, and decreasing model size, while incurring only a minimal drop in accuracy. Results in \Cref{table:segmentation} further demonstrate that a linear transformation is sufficient even for a more complex task such as segmentation, indicating that appropriately selecting which layers to approximate enables model size reduction with minimal impact on downstream accuracy.

\begin{table}[h]
    \centering
    \caption{\textbf{\gls{toast} Text Classification Performance on \agnews{}}. Text classification accuracy, \gls{gflops}, and throughput for \mbertb{} using \agnews{}. The ``Approx.'' column specifies the block mapping (output of the first block is used to approximate the output of the second). \gls{mlp} is a trained approximator, while Linear is closed-form and training-free. Results are averaged over three seeds. }
    \label{table:text-classification}
    \centering
    \resizebox{.9\textwidth}{!}{%
        \begin{tabular}{ccccccccc}
            \toprule
                                                                        &                     & \multicolumn{3}{c}{Linear} & \multicolumn{3}{c}{\gls{mlp}}                                                                                     \\
            \cmidrule(l){3-5} \cmidrule(l){6-8}
            Approx.                                                     & Params $\downarrow$ & Accuracy \% $\uparrow$      & GFLOPs $\downarrow$           & tokens/s $\uparrow$ & Accuracy \% $\uparrow$ & GFLOPs $\downarrow$ & tokens/s $\uparrow$ \\
            \midrule
            11 $\rightarrow$ 21                                         & \texttt{92.82M}     & $0.81 \pm 0.05$            & 12.7                          & 2264.0           & $0.73 \pm 0.00$       & 12.68               & 2216.50          \\
            4 $\rightarrow$ 8, 11 $\rightarrow$ 14, 18 $\rightarrow$ 21 & \texttt{92.82M}     & $0.82 \pm 0.07$            & 12.7                          & 2220.7           & $0.73 \pm 0.01$       & 12.68               & 2155.16          \\
            \midrule
            4 $\rightarrow$ 7, 18 $\rightarrow$ 21                      & \texttt{109.68M}    & $0.82 \pm 0.07$            & 15.9                          & 1803.9           & $0.71 \pm 0.02$       & 15.85               & 1771.80          \\
            \midrule
            4 $\rightarrow$ 8                                           & \texttt{126.54M}    & $0.86 \pm 0.02$            & 19.0                          & 1636.0           & $0.82 \pm 0.01$       & 19.03               & 1632.65          \\
            \midrule
            11 $\rightarrow$ 14                                         & \texttt{132.16M}    & $0.86 \pm 0.02$            & 20.1                          & 1544.3           & $0.82 \pm 0.01$       & 20.08               & 1540.23          \\
            18 $\rightarrow$ 21                                         & \texttt{132.16M}    & $0.85 \pm 0.02$            & 20.1                          & 1472.8           & $0.82 \pm 0.01$       & 20.08               & 1467.56          \\
            \midrule
            4 $\rightarrow$ 5                                           & \texttt{143.4M}    & $0.88 \pm 0.00$            & 22.2                          & 1380.3           & $0.81 \pm 0.01$       & 22.20               & 1384.42          \\
            11 $\rightarrow$ 12                                         & \texttt{143.4M}    & $0.87 \pm 0.02$            & 22.2                          & 1378.8           & $0.82 \pm 0.01$       & 22.20               & 1394.63          \\
            20 $\rightarrow$ 21                                         & \texttt{143.4M}    & $0.87 \pm 0.02$            & 22.2                          & 1340.2           & $0.84 \pm 0.00$       & 22.20               & 1332.27          \\
            \midrule
            original                                                    & \texttt{149.01M}    & $0.88 \pm 0.00$            & 23.25                         & 1337.25          & $0.88 \pm 0.00$       & 23.25               & 1347.46          \\
            \bottomrule
        \end{tabular}
    }
\end{table}

\begin{tcolorbox}[colframe=border, colback=takeaway, opacityback=0.95, boxrule=0mm]
\textbf{Takeaway} \gls{toast} extends beyond vision and standard classification, demonstrating broader applicability across domains.
\end{tcolorbox}

\section{Limitations and Future Work} \label{sec:limitations}
While \gls{toast} efficiently approximates transformer blocks, our current investigation has primarily focused on vision transformer architectures and their application to classification tasks with preliminary results also extending to segmentation and text classification.
Additionally, our experimental evaluation largely utilizes a frozen backbone with a linear classifier trained for limited epochs. While this setting was chosen to simulate a strict computational budget and to isolate the direct impact of the approximation on the features, the resulting models are naturally less expressive than those finetuned end-to-end or equipped with complex task-specific heads.
From a theoretical standpoint, our study remains primarily empirical and does not yet provide a rigorous characterization of why linear block approximations are effective.
An important open question is whether over-parameterization in transformers induces low-dimensional or nearly linearly equivalent representations within certain blocks. Future work will aim to bridge this gap by analyzing model expressivity through the lens of linear regions and related complexity measures. 
Furthermore, we aim to expand the analysis of these block-level similarities.
This involves investigating redundancies at finer granularities, such as within individual attention heads or feed-forward layers, and developing more principled and reliable metrics for automatically selecting which blocks to approximate.
The heuristic used in the current work, while effective, is not yet fully accurate, and improving it could enable more consistent identification of approximation-friendly layers with minimal impact on downstream performance.
Such advancements may lead to more refined, context-aware approximation strategies that further enhance model efficiency.

\section{Conclusion} \label{sec:conclusion}
In this work, we first analyze the emergence of consistent block-wise representation similarities within pretrained transformer models and then propose a method to leverage these similarities to obtain smaller and more efficient yet performant models. To this end, we propose \glsentryfull{toast}, a novel method for easily approximating entire transformer blocks using a simple transformation, without requiring additional training or finetuning. Our extensive empirical evaluations across multiple pretrained vision models and datasets validate that \gls{toast} significantly reduces model parameters while maintaining, and sometimes even improving, downstream task performance. Furthermore, \gls{toast}'s straightforward linear approach often achieves better results than existing strategies like block skipping, and can be more effective than complex, trained approximations. \gls{toast} thus offers a practical and efficient method for streamlining foundation models, making them more computationally accessible, and towards deployment in resource-constrained scenarios such as on-device settings.
\subsubsection*{Acknowledgments}
The authors gratefully acknowledge Luca Moschella for the insightful discussions. 
IC was supported as part of the Swiss AI Initiative by a grant from the Swiss National Supercomputing Centre (CSCS) under project ID a135 on Alps.
TS and EP are supported by the grant \#2021-911 of the Strategic Focal Area ``Personalized Health and Related Technologies (PHRT)'' of the ETH Domain (Swiss Federal Institutes of Technology). EP is also supported by a fellowship from the ETH AI Center.
This work has received funding from the Swiss State Secretariat for Education, Research, and Innovation~(SERI).

\afterpage{\clearpage}
\bibliography{main}
\bibliographystyle{tmlr}

\appendix
\clearpage
\section{Appendix} \label{sec:appendix}

\subsection{Implementation Details} \label{sec:app-implementation-details}

This section details the experiments conducted in \Cref{sec:experiments}, providing information to reproduce them. Additionally, the code is available on \href{https://github.com/icannistraci/toast}{GitHub}.

\subsubsection{Models and Datasets}  \label{sec:app-info}

\Cref{tab:pretrained-info} contains the full list of the pretrained models, while \Cref{tab:dataset-info} contains dataset information.

\begin{table}[ht]
\caption{\textbf{Pretrained models details.} Details of the pretrained feature extractors with their HuggingFace key, their alias, and their latent space dimensionality.}  \label{tab:pretrained-info}
\centering
\resizebox{\textwidth}{!}{%
\begin{tabular}{cllc}
    \toprule
    Modality & HuggingFace Model Name & Alias & Enc. Dim \\
    \midrule
    \multirow{8}{*}{Vision}
    & WinKawaks/vit-tiny-patch16-224 & \vitt{} \citep{vit} & 192 \\
    \cmidrule{2-4}
    & WinKawaks/vit-small-patch16-224 & \vits{} \citep{vit} & 384 \\
    & facebook/dinov2-small & \dinos{} \citep{oquab2023dinov2}  & 384 \\
    & facebook/deit-small-patch16-224 & \deits{} \citep{touvron2020training}  & 384 \\
    \cmidrule{2-4}
    & google/vit-base-patch16-224 & \vitb{} \citep{vit}  & 768 \\
    & facebook/dinov2-base & \dinob{} \citep{oquab2023dinov2}  & 768 \\
    & laion/CLIP-ViT-B-16-laion2B-s34B-b88K & \clipb{} \citep{zhai2019large}  & 768 \\
    \cmidrule{2-4}
    & google/vit-large-patch16-224 & \vitl{} \citep{vit} & 1024 \\
    \midrule
    \multirow{1}{*}{Text}
    & answerdotai/ModernBERT-base & \mbertb{} \citep{warner2025smarter} & 768 \\
    \bottomrule
\end{tabular}
}
\end{table}

\begin{table}[ht]
    \caption{\textbf{Dataset details.} Details of the HuggingFace datasets used in the classification and reconstruction experiments, with the associated number of classes.}
    \label{tab:dataset-info}
\centering
\small
    \begin{tabular}{clcc}
        \toprule
        Modality & Name & Alias & $\#$ Classes \\
        \midrule
        \multirow{7}{*}{Vision}
        & MNIST \citep{mnist} & \mnist{} & 10 \\
        & Fashion-MNIST  \citep{xiao2017fashion} & \fmnist{}  & 10 \\
        & CIFAR-10 \citep{krizhevsky2009learning} & \cifart{} & 10 \\
    \cmidrule{2-4}
        & CIFAR-100 (coarse) \citep{krizhevsky2009learning} & \cifarhc{} & 20   \\
     \cmidrule{2-4}
         & CIFAR-100 (fine) \citep{krizhevsky2009learning} & \cifarhf{} & 100 \\
     \cmidrule{2-4}
         & SceneParse150 \citep{zhou2017scene,zhou2016semantic} & \scene{}     & 150   \\
     \cmidrule{2-4}
         & ImageNet-1k \citep{imagenet15russakovsky} & \imagenet{}     & 1000   \\
         \midrule
         \multirow{1}{*}{Text}
         & AG News \cite{Zhang2015CharacterlevelCN} & \agnews{}     & 4  \\
        \bottomrule
    \end{tabular}
\end{table}

\subsubsection{Tools \& Technologies} \label{sec:app-tools}
All the experiments presented in this work employ the following tools:
\begin{itemize}
    \item \textit{PyTorch Lightning}, to ensure reproducible results while also getting a clean and modular codebase;
    \item \textit{NN-Template GrokAI (2021)}, to easily bootstrap the project and enforce best practices;
    \item \textit{Transformers by HuggingFace}, to get ready-to-use transformers for both text and images;
    \item \textit{Datasets by HuggingFace}, to access most of the datasets;
    \item \textit{DVC} \citep{dvc}, for data versioning;
    \item \textit{fvcore by Facebook Research (FAIR)}, for calculating \gls{gflops};
\end{itemize}

\subsubsection{Computational Resources}\label{sec:app-computation}
Experiments involving larger models, specifically \dinob{}, \clipb{}, and \vitl{}, were conducted on an NVIDIA H100 GPU equipped with 93 GB of memory. All the other experiments utilized an NVIDIA GeForce RTX 5090 GPU with 31 GB of memory.

\subsubsection{Efficiency Metrics}\label{sec:app-efficiency}
We evaluated model efficiency using two primary metrics. \gls{gflops} were used to measure the hardware-independent theoretical complexity of a single forward pass, calculated using the \texttt{fvcore} analysis library. Throughput, measured in samples per second, was used to quantify the practical, hardware-dependent inference speed. This was benchmarked by averaging the wall-clock time over numerous iterations on a single NVIDIA H100 GPU with a consistent batch size of 256.

\subsubsection{Approximators}  \label{sec:app-approximators}
The first implementation, referred to as the Res-\gls{mlp}, is composed of two normalization layers and a feedforward submodule. The first layer normalization processes the input tensor, followed by a feedforward submodule comprising a linear transformation, a SiLU activation, a dropout layer, and a final linear transformation. The output of the feedforward submodule is added to the normalized input via a residual connection. This sum is then passed through the second normalization layer to produce the final output. The second implementation, referred to as the \gls{mlp}, is a simplified \gls{mlp} that employs a sequential architecture with an initial linear transformation that reduces the input dimensionality to half of the target dimension, followed by a GELU activation function for smooth non-linearity, and a final linear transformation that restores the reduced features to match the target dimensionality. Refer to \Cref{lst:mlp1,lst:mlp2} for the code snippet of the two translators.

\begin{lstlisting}[style=python, caption={Python Code Snippet for the Res-MLP translator}, label={lst:mlp1}]
class ResMLP(nn.Module):
    def __init__(self, num_features: int, dropout_p: float):
        super().__init__()

        self.norm1 = nn.LayerNorm(num_features)
        self.norm2 = nn.LayerNorm(num_features)

        self.ff = nn.Sequential(
            nn.Linear(num_features, num_features),
            nn.SiLU(),
            nn.Dropout(p=dropout_p),
            nn.Linear(num_features, num_features),
        )

    def forward(self, x: torch.Tensor) -> torch.Tensor:
        x_normalized = self.norm1(x)
        x_transformed = self.ff(x_normalized)
        return self.norm2(x_transformed + x_normalized)
\end{lstlisting}

\begin{lstlisting}[style=python, caption={Python Code Snippet for the MLP translator}, label={lst:mlp2}]
translation = nn.Sequential(
    nn.Linear(x.size(1), y.size(1) // 2),
    nn.GELU(),
    nn.Linear(y.size(1) // 2, y.size(1)),
)
\end{lstlisting}

\subsubsection{Metric Ablation} \label{sec:app-metric-ablation}

We introduce linear approximation error as a simple, stable, and sample-efficient criterion for identifying redundant transformer blocks, offering a practical alternative for guiding block approximation. This metric measures how well the representation of a later block can be reconstructed from an earlier one through a least-squares projection, providing a direct estimate of how much additional structure the skipped layers contribute. Importantly, the error can be estimated using as few as $50$ samples, producing substantially more stable and interpretable rankings compared to other metrics. 

\begin{table}[h]
\centering
\small
\caption{\textbf{Top-5 Block Approximation Recommendation}. Top 5 recommended blocks to be approximated based on linear approximation error using \deits{} and \cifarhf{}.}
\label{tab:skip_recommendations}
\begin{tabular}{@{}ccccc@{}}
\toprule
Rank & Approx & \# Layers & Predicted Error & Accuracy \% \\
\midrule
1 & $9 \rightarrow 10$ & 1 & 0.14 & $71.69 \pm 0.11$ \\
2 & $10 \rightarrow 11$ & 1 & 0.18 & $71.17 \pm 0.19$ \\
3 & $8 \rightarrow 9$ & 1 & 0.23 & $70.83 \pm 0.13$ \\
4 & $9 \rightarrow 11$ & 2 & 0.25 & $71.14 \pm 0.15$ \\
5 & $8 \rightarrow 10$ & 2 & 0.26 & $71.06 \pm 0.19$ \\
\midrule
- & original & 0 & - & $71.13$ \\
\bottomrule
\end{tabular}
\end{table}

As shown in \Cref{tab:skip_recommendations}, linear approximation error correlates strongly with the actual accuracy impact of skipping or approximating a block range: blocks with the lowest error consistently incur minimal or no downstream performance degradation. This makes the metric both computationally lightweight and practically reliable for identifying redundant or compressible transformer regions.

To further validate this choice, we conduct an ablation comparing several candidate similarity metrics (e.g., cosine distance, MSE, Euclidean distance, and \gls{cka}) and evaluate how well each predicts the true accuracy drop after approximation. Results, summarized in \Cref{tab:ranking_metrics}, show that linear approximation error achieves the most consistent performance across architectures, with competitive or superior Precision@5 and Recall@5 scores. Notably, metrics such as cosine distance and Euclidean distance exhibit behavior that is highly model-dependent, while \gls{cka} performs well in some cases but is less stable across architectures and budgets.

\begin{table}[h]
\caption{\textbf{Block Selection Strategy Ablation.} Ranking evaluation metrics for approximation quality prediction on CIFAR-100 using \deits{}, \dinos{}, and \dinob{}. Precision@5 and Recall@5 are shown for each model.\looseness=-1}
\label{tab:ranking_metrics}
\centering
\small
\begin{tabular}{cccccccccc}
\toprule
 & \multicolumn{2}{c}{\deits{}} &  \multicolumn{2}{c}{\dinos{}} & \multicolumn{2}{c}{\dinob{}} & \multicolumn{2}{c}{Mean} \\
\cmidrule{2-9}
& P@5 & R@5 & P@5 & R@5 & P@5 & R@5 & P@5 & R@5 \\
\midrule
Linear Error        & 0.6   & 0.6   & 0.6   & 0.6   & 0.6   & 0.6   & 0.60 & 0.60 \\
Cosine     & 1.0   & 1.0   & 0.4   & 0.4   & 0.2   & 0.2   & 0.53 & 0.53 \\
CKA        & 0.6   & 0.6   & 0.6   & 0.6   & 0.4   & 0.4   & 0.53 & 0.53 \\
MSE                 & 0.0   & 0.0   & 0.4   & 0.4   & 0.6   & 0.6   & 0.33 & 0.33 \\
Euclidean  & 0.8   & 0.8   & 0.4   & 0.4   & 0.4   & 0.4   & 0.53 & 0.53 \\
\bottomrule
\end{tabular}
\end{table}

Overall, this ablation highlights that linear approximation error provides the best trade-off between stability, computational cost, and predictive fidelity, making it a strong default metric for block selection in transformer approximation.\looseness=-1

\newpage
\subsubsection{Block Selection Pseudocode} \label{sec:app-selection-pseudocode}

\begin{algorithm}[h]
\caption{Identify Top-$k$ Layer Skip Configurations}
\label{alg:layer_skip_selection}
\begin{algorithmic}[1]
\Require Model encoder $\mathcal{M}$ with $L$ layers, dataset $\mathcal{D}$, number of top configurations $k$, skip budget $b$ (optional)
\Ensure Top-$k$ skip configurations $\mathcal{S} = \{(s_1, e_1), \ldots, (s_k, e_k)\}$

\State Extract layer representations: $\mathbf{H}_i \gets \text{encode}(\mathcal{M}, \mathcal{D}, \text{layer}_i)$ for $i \in [0, L]$
\State Initialize error matrix $\mathbf{E} \in \mathbb{R}^{L \times L}$

\For{$i = 0$ to $L-1$}
    \For{$j = i+1$ to $L$}
        \State $\mathbf{E}_{i,j} \gets \text{LinearApproximationError}(\mathbf{H}_i, \mathbf{H}_j)$
    \EndFor
\EndFor

\State Initialize candidate list $\mathcal{C} \gets \emptyset$
\For{$i = 0$ to $L-1$}
    \For{$j = i+1$ to $L$}
        \If{$b$ is specified and $j - i \neq b$}
            \State \textbf{continue} \Comment{Skip if not matching budget}
        \EndIf
        \State $\mathcal{C} \gets \mathcal{C} \cup \{(i, j, \mathbf{E}_{i,j})\}$
    \EndFor
\EndFor

\State Sort $\mathcal{C}$ by error in ascending order
\State $\mathcal{S} \gets$ top-$k$ configurations from $\mathcal{C}$
\State \Return $\mathcal{S}$
\end{algorithmic}
\end{algorithm}

\begin{algorithm}[h]
\caption{Linear Approximation Error}
\label{alg:linear_error}
\begin{algorithmic}[1]
\Require Source layer representations $\mathbf{X} \in \mathbb{R}^{n \times d}$, target layer representations $\mathbf{Y} \in \mathbb{R}^{n \times d}$
\Ensure Normalized residual error $\epsilon$

\State Solve least-squares: $\mathbf{W}^* = \arg\min_{\mathbf{W}} \|\mathbf{Y} - \mathbf{X}\mathbf{W}\|_F^2$
\State Compute prediction: $\hat{\mathbf{Y}} = \mathbf{X}\mathbf{W}^*$
\State Compute normalized error: $\epsilon = \frac{\|\mathbf{Y} - \hat{\mathbf{Y}}\|_F}{\|\mathbf{Y}\|_F}$
\State \Return $\epsilon$
\end{algorithmic}
\end{algorithm}

\begin{table}[h]
    \centering
    \caption{\textbf{Top-3 Block Approximation Recommendation}. Top 3 recommended blocks to be approximated based on linear approximation error and number of blocks to skip using \deits{} and \cifarhf{}.\looseness=-1}
    \label{tab:skip_by_budget}
    \resizebox{.5\textwidth}{!}{
        \begin{tabular}{ccccc}
            \toprule
            \# Blocks             & Rank & Approx.             & Predicted Error & Accuracy \%      \\
            \midrule
            \multirow{3}{*}{1} & 1    & $9 \rightarrow 10$  & 0.14           & $71.69 \pm 0.11$ \\
                               & 2    & $10 \rightarrow 11$ & 0.18           & $71.17 \pm 0.19$ \\
                               & 3    & $8 \rightarrow 9$   & 0.23           & $70.83 \pm 0.13$ \\
            \midrule
            \multirow{3}{*}{2} & 1    & $9 \rightarrow 11$  & 0.25           & $71.14 \pm 0.15$ \\
                               & 2    & $8 \rightarrow 10$  & 0.26           & $71.06 \pm 0.19$ \\
                               & 3    & $7 \rightarrow 9$   & 0.36           & $69.00 \pm 0.43$ \\
            \midrule
            \multirow{3}{*}{3} & 1    & $8 \rightarrow 11$  & 0.36           & $68.22 \pm 0.40$ \\
                               & 2    & $7 \rightarrow 10$  & 0.38           & $69.08 \pm 0.24$ \\
                               & 3    & $6 \rightarrow 9$   & 0.45           & $65.64 \pm 0.03$ \\
            \midrule
            0                  & -    & original            & -               & $71.13$          \\
            \bottomrule
        \end{tabular}
    }
\end{table}

\newpage
\subsection{Additional Experiments}

This section presents supplementary experiments to extend those detailed in \Cref{sec:experiments}.

\subsubsection{Latent Analysis} \label{sec:app-similarities}

\begin{figure}[h]
    \centering  
    \begin{minipage}[t]{.4\textwidth}
    \centering
        \begin{overpic}[width=\textwidth]{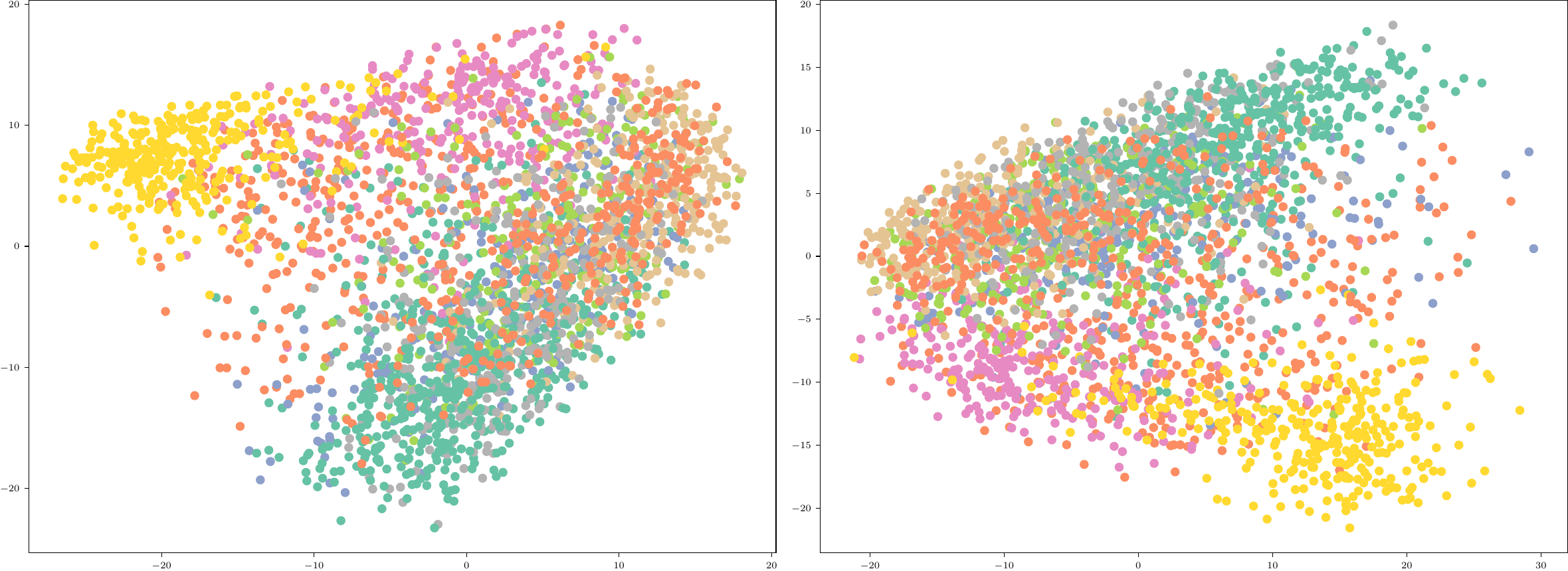}
            \put(-4,11){\rotatebox{90}{\mnist{}}}
            \put(15,38){\texttt{Original}}
            \put(68.5,38){\texttt{TOAST}} 
        \end{overpic}
    \end{minipage}
    \hspace{1em}
    \begin{minipage}[t]{.4\textwidth}
    \centering
        \begin{overpic}[width=\textwidth]{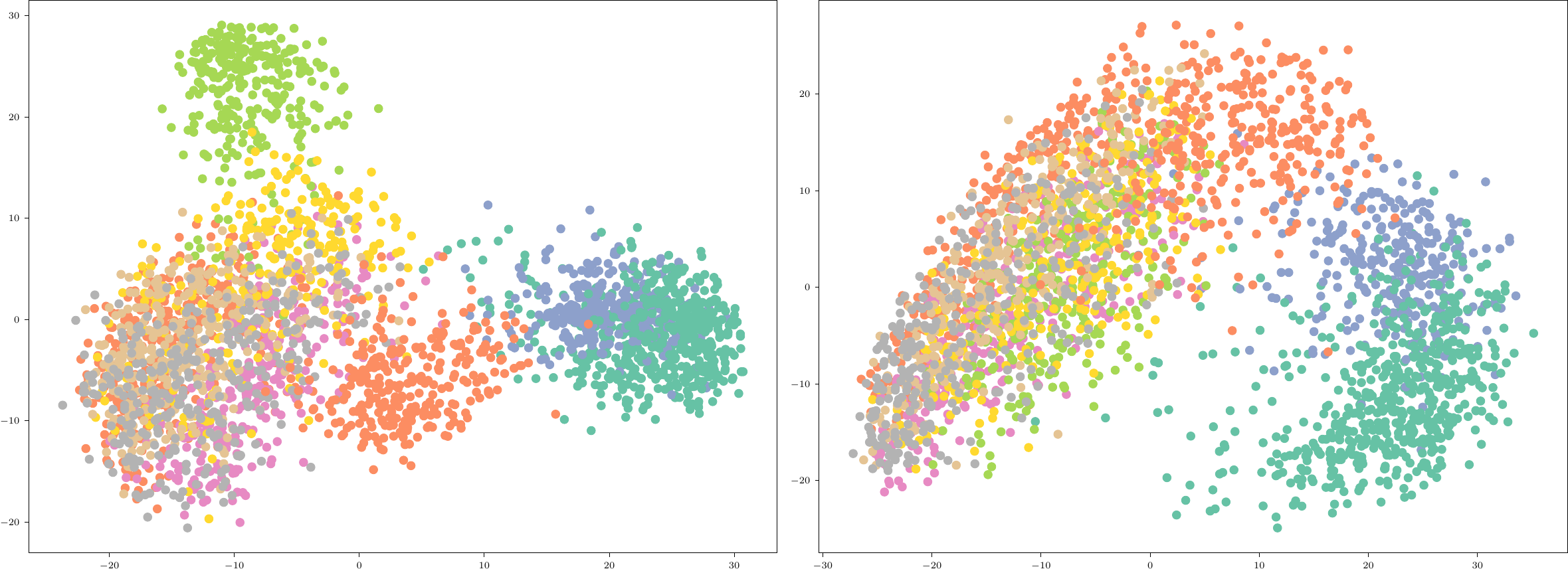}
            \put(-4,8){\rotatebox{90}{\fmnist{}}}
            \put(15,38){\texttt{Original}}
            \put(68.5,38){\texttt{TOAST}} 
        \end{overpic}
        \vspace{.1em}
    \end{minipage}
    \begin{minipage}[t]{.4\textwidth}
        \begin{overpic}[width=\textwidth]{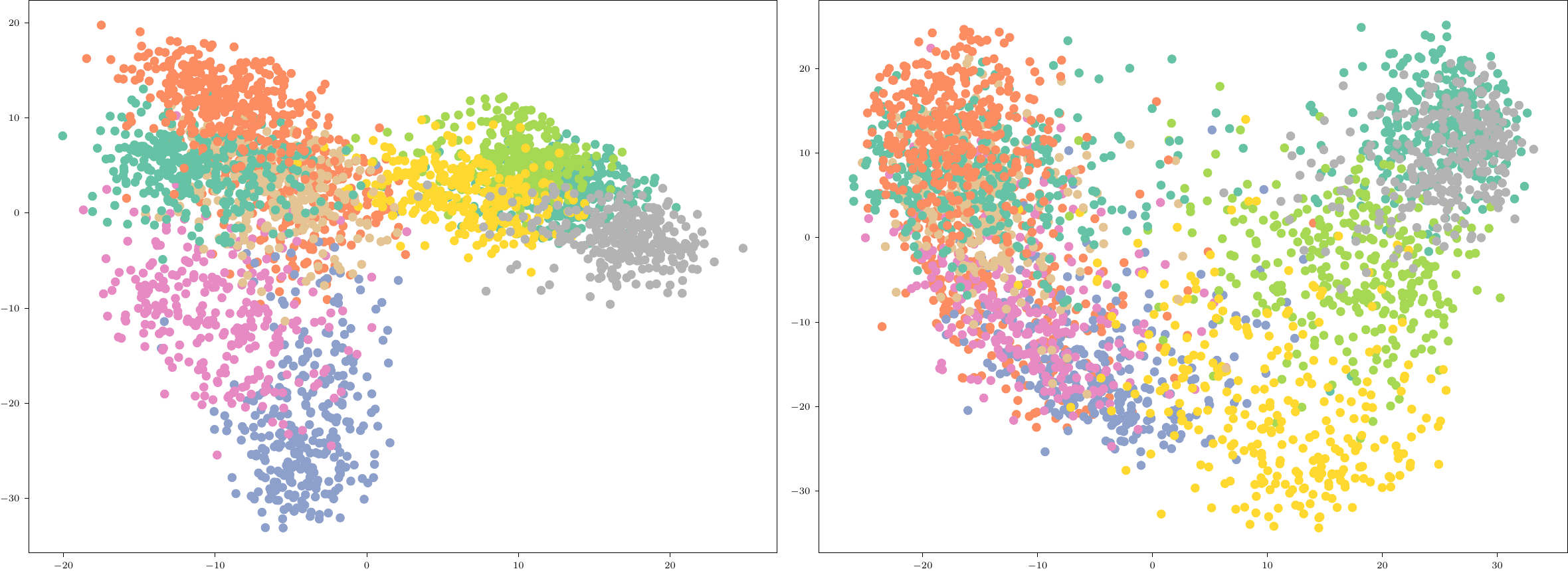}
            \put(-4,8){\rotatebox{90}{\cifart{}}}
        \end{overpic}        
    \end{minipage}
    \hspace{1em}
    \begin{minipage}[t]{.4\textwidth}
        \begin{overpic}[width=\textwidth]{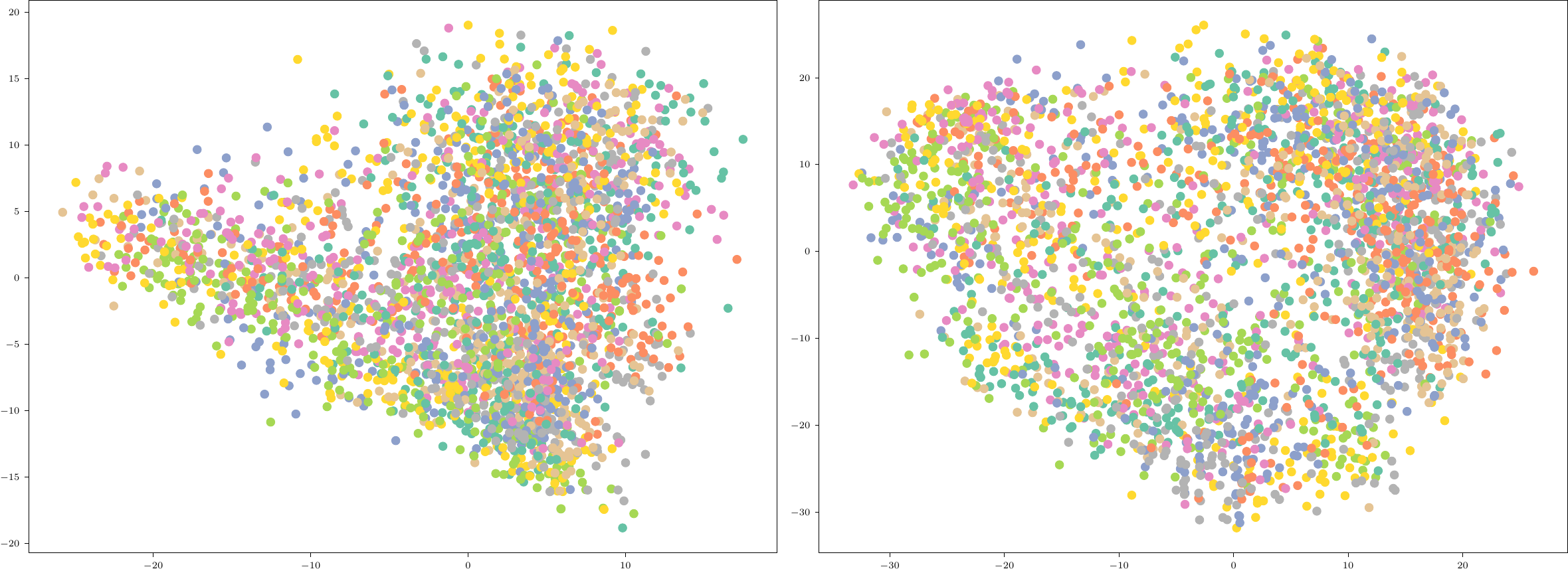}
            \put(-4,6){\rotatebox{90}{\cifarh{}}}
        \end{overpic}        
    \end{minipage}
    \caption{\textbf{Last Block Approximation.} \gls{pca} visualization of the final layer representations for both the original model and the model with its last block approximated from the preceding one. The representations are generated using the \dinos{} model across four datasets. The plots highlight that the last layer representations in this model are crucial, making it more effective to approximate earlier blocks instead. Note that for \cifarh{} (bottom right), only the overall structure of the space can be observed, as the 100 classes make it challenging to distinguish labels based on color.}
    \label{fig:app-pca-approx-dino-11}
\end{figure}

\begin{figure}[h]
    \centering  
    \vspace{1em}
    \begin{minipage}[t]{.4\textwidth}
    \centering
        \begin{overpic}[width=\textwidth]{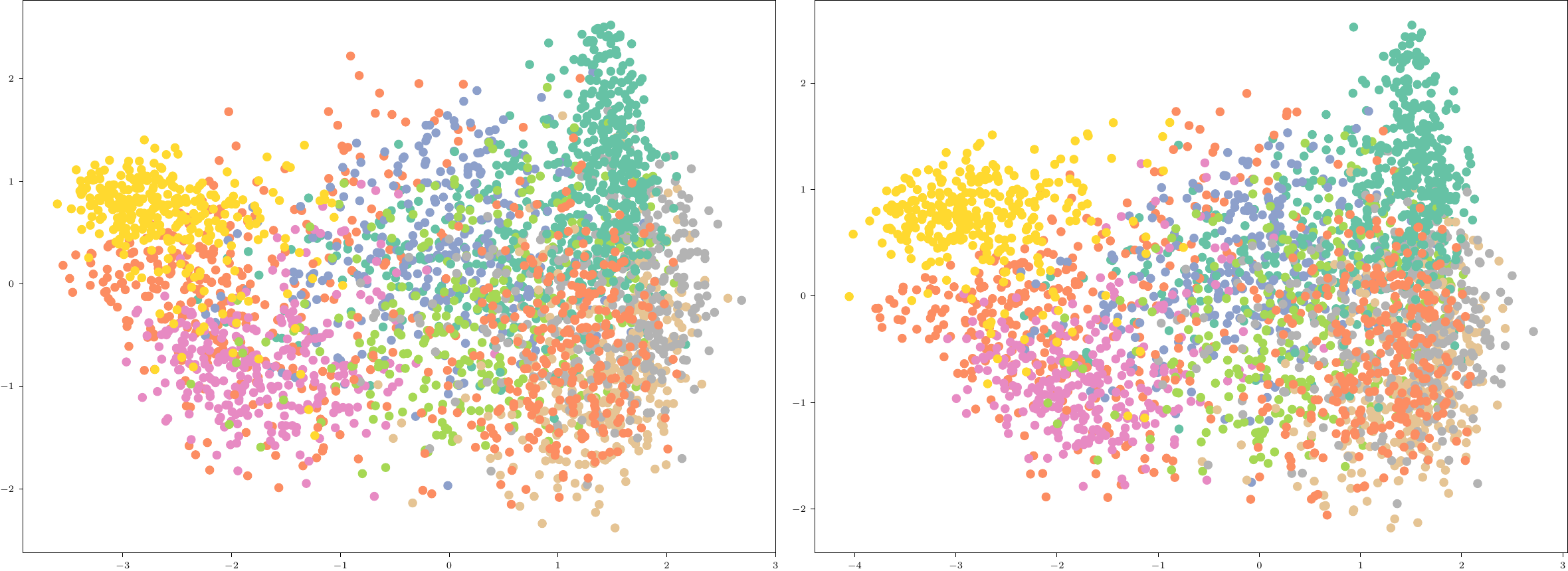}
            \put(-4,11){\rotatebox{90}{\mnist{}}}
            \put(15,38){\texttt{Original}}
            \put(68.5,38){\texttt{TOAST}} 
        \end{overpic}
    \end{minipage}
    \hspace{1em}
    \begin{minipage}[t]{.4\textwidth}
    \centering
        \begin{overpic}[width=\textwidth]{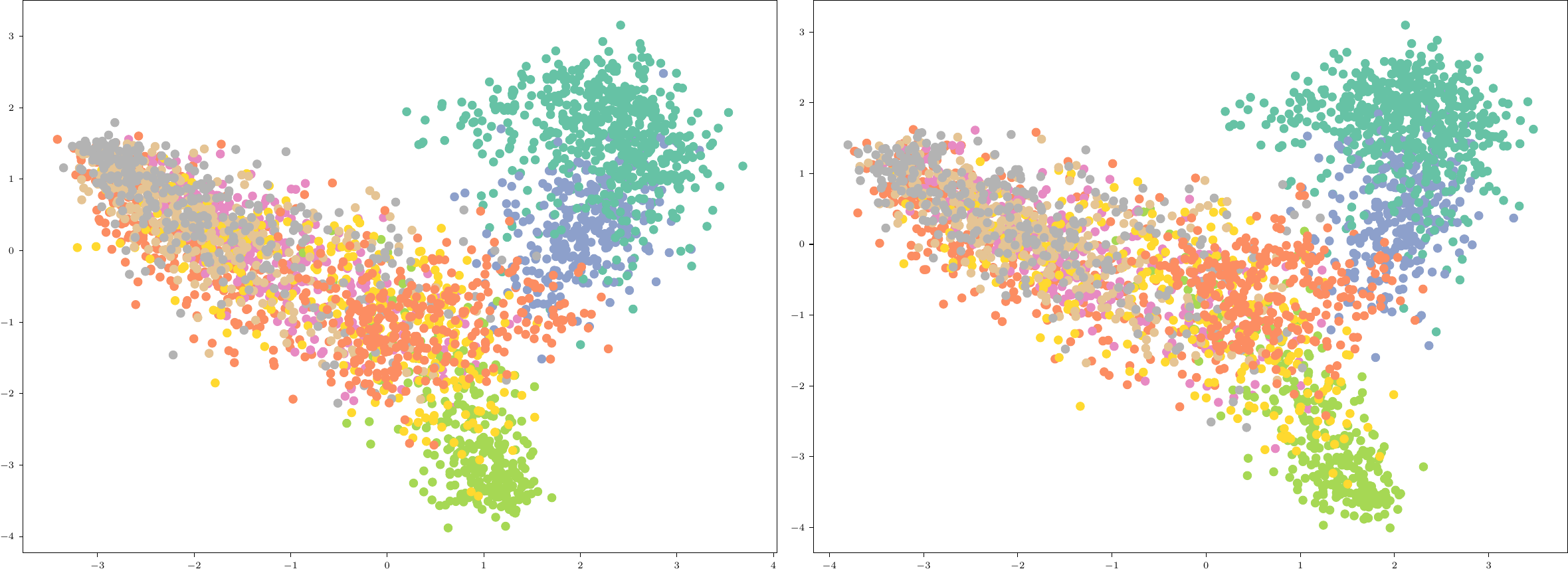}
            \put(-4,8){\rotatebox{90}{\fmnist{}}}
            \put(15,38){\texttt{Original}}
            \put(68.5,38){\texttt{TOAST}} 
        \end{overpic}
        \vspace{.1em}
    \end{minipage}
    \begin{minipage}[t]{.4\textwidth}
        \begin{overpic}[width=\textwidth]{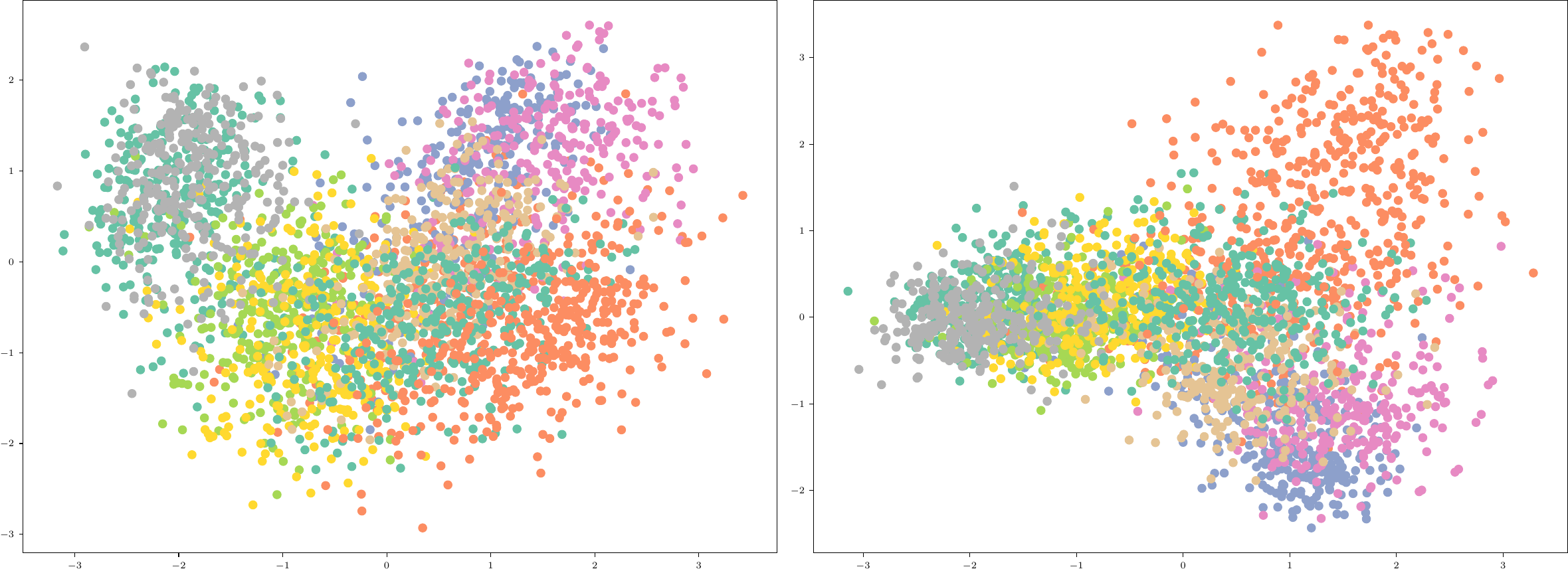}
            \put(-4,8){\rotatebox{90}{\cifart{}}}
        \end{overpic}        
    \end{minipage}
    \hspace{1em}
    \begin{minipage}[t]{.4\textwidth}
        \begin{overpic}[width=\textwidth]{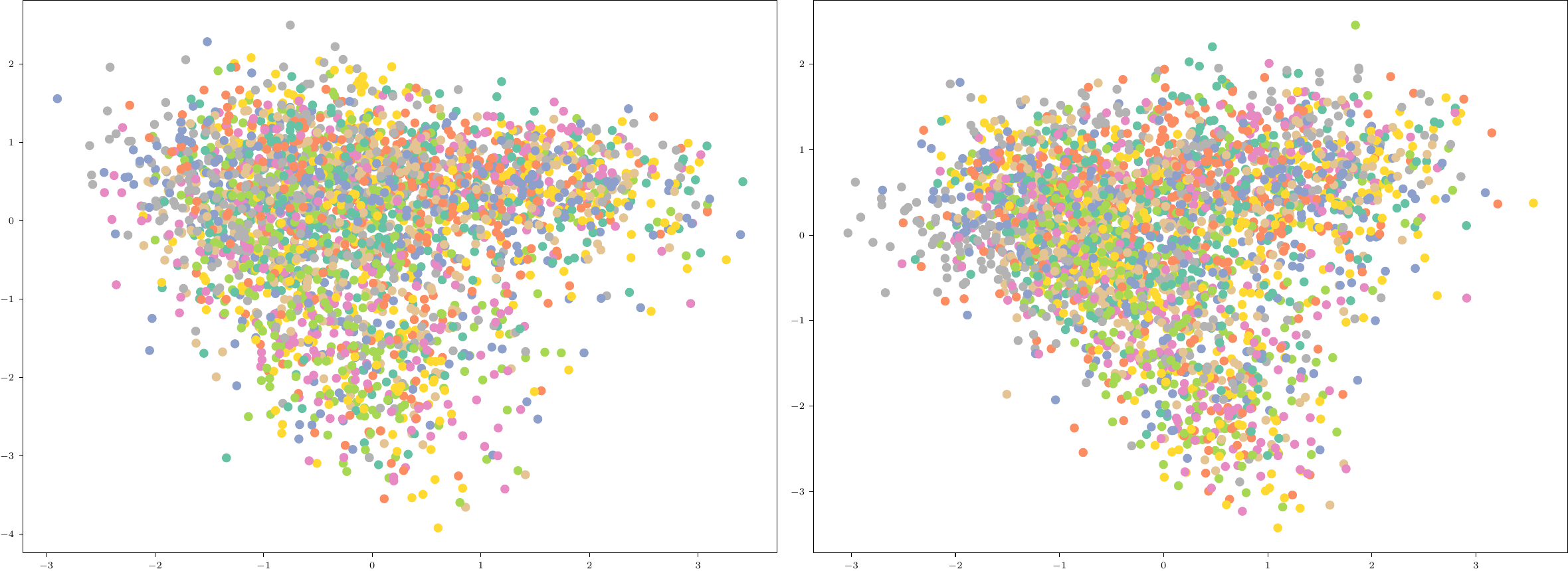}
            \put(-4,6){\rotatebox{90}{\cifarh{}}}
        \end{overpic}        
    \end{minipage}
    \caption{\textbf{Last Block Approximation.} \gls{pca} visualization of the final layer representations for both the original model and the model with its last block approximated by the preceding one. The representations are generated using the \deits{} model across four datasets. The plots highlight that in this model, the representations in the last layer are redundant and can be effectively approximated, offering potential performance improvements while reducing model complexity and parameter count. Note that for \cifarh{} (bottom right), only the overall structure of the space can be observed, as the 100 classes make it challenging to distinguish labels based on color.}
    \label{fig:app-pca-approx-deit-11}
\end{figure}

\begin{figure}[h]
    \centering  
    \begin{minipage}[t]{.4\textwidth}
    \centering
        \begin{overpic}[width=\textwidth]{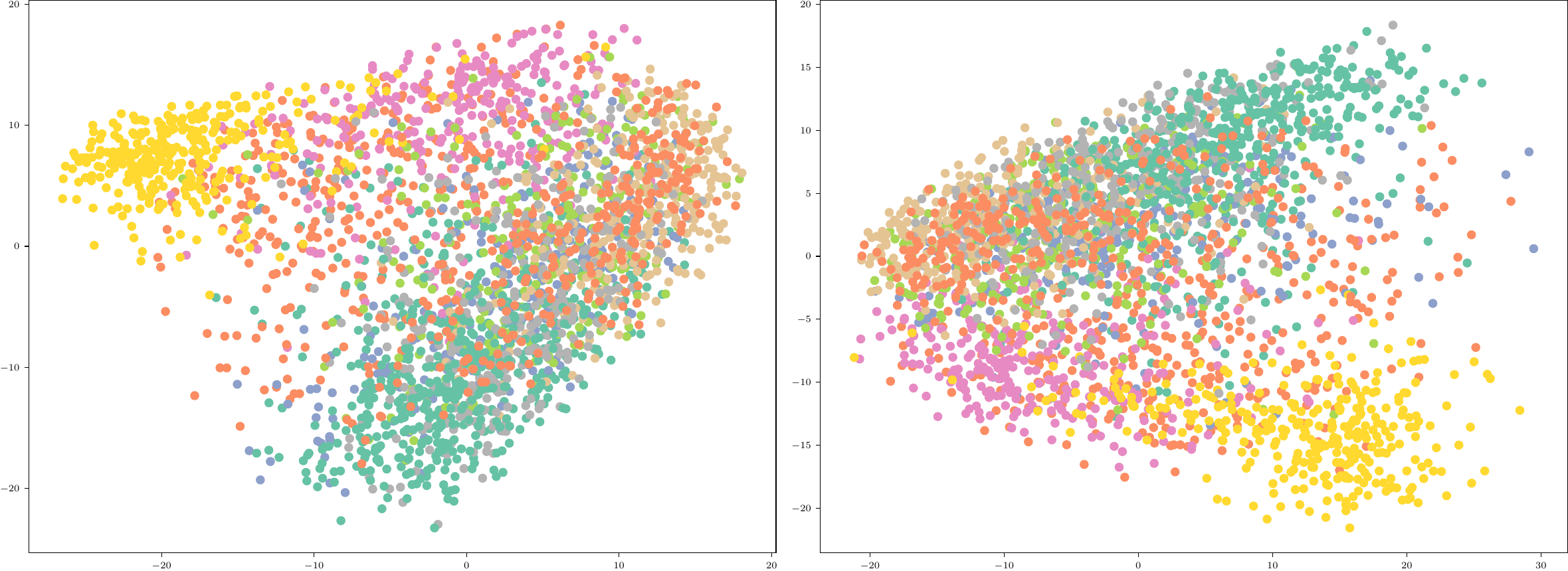}
            \put(-4,11){\rotatebox{90}{\mnist{}}}
            \put(15,38){\texttt{Original}}
            \put(68.5,38){\texttt{TOAST}} 
        \end{overpic}
    \end{minipage}
    \hspace{1em}
    \begin{minipage}[t]{.4\textwidth}
    \centering
        \begin{overpic}[width=\textwidth]{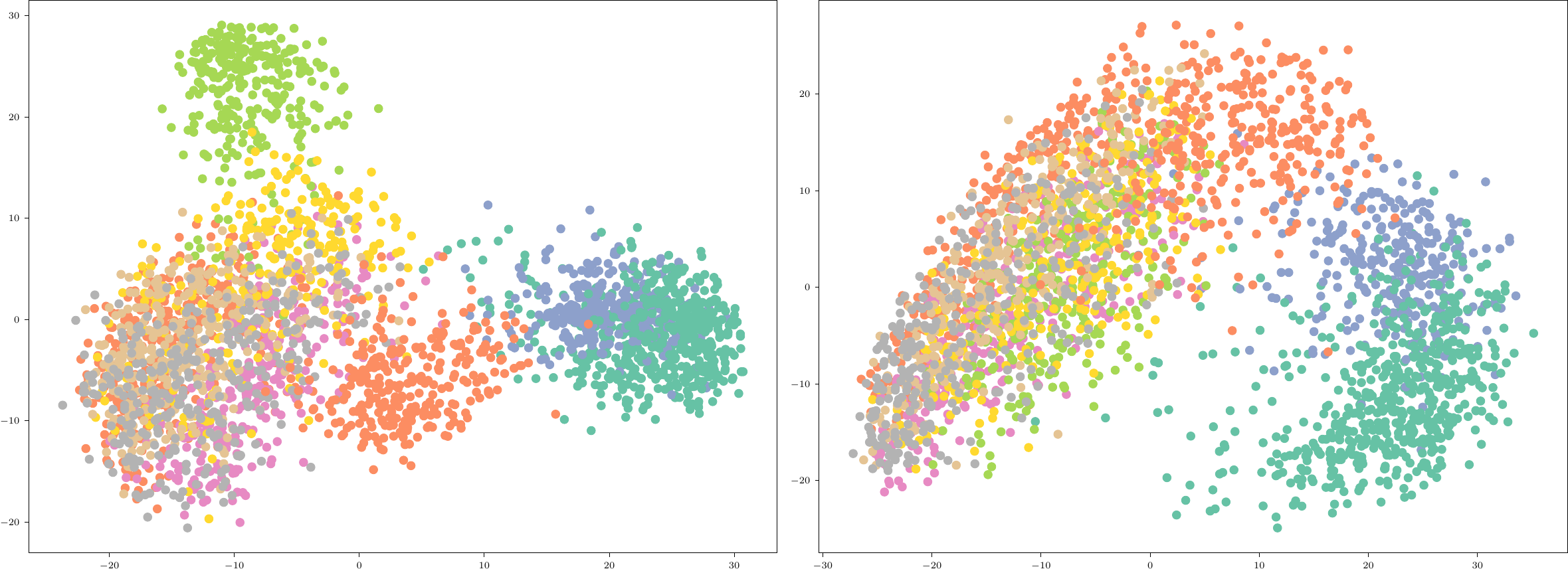}
            \put(-4,8){\rotatebox{90}{\fmnist{}}}
            \put(15,38){\texttt{Original}}
            \put(68.5,38){\texttt{TOAST}} 
        \end{overpic}
        \vspace{.1em}
    \end{minipage}
    \begin{minipage}[t]{.4\textwidth}
        \begin{overpic}[width=\textwidth]{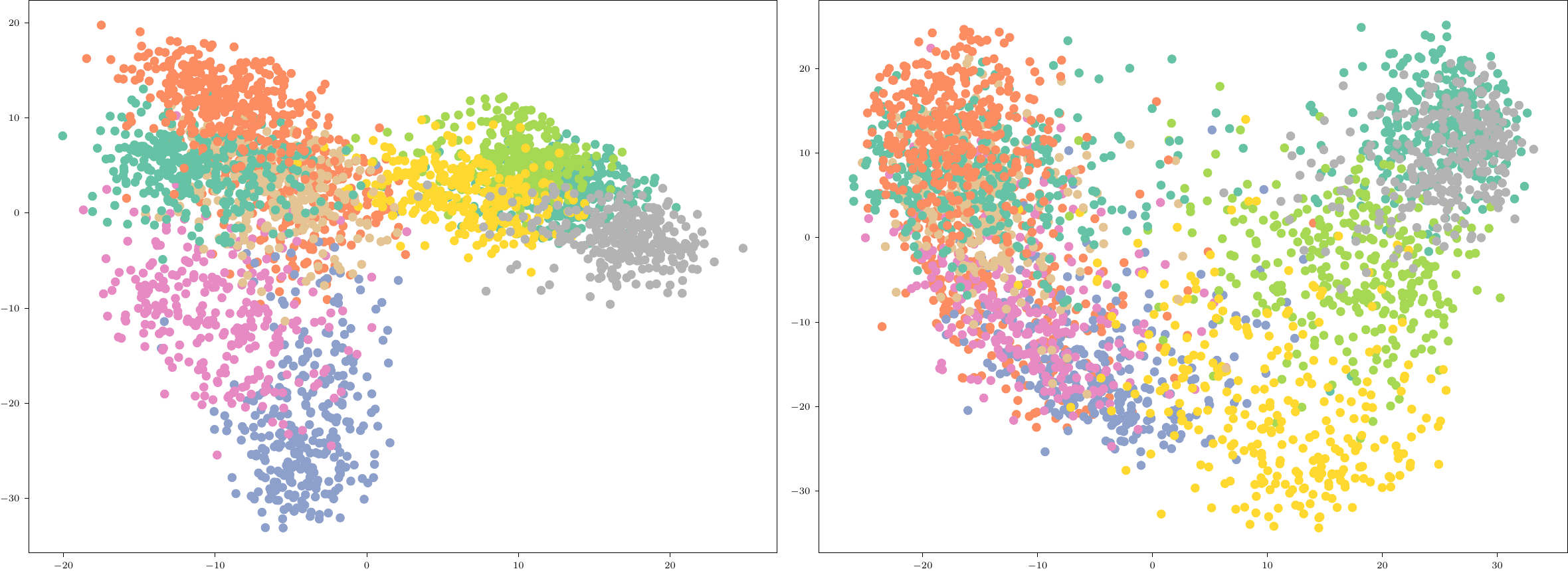}
            \put(-4,8){\rotatebox{90}{\cifart{}}}
        \end{overpic}        
    \end{minipage}
    \hspace{1em}
    \begin{minipage}[t]{.4\textwidth}
        \begin{overpic}[width=\textwidth]{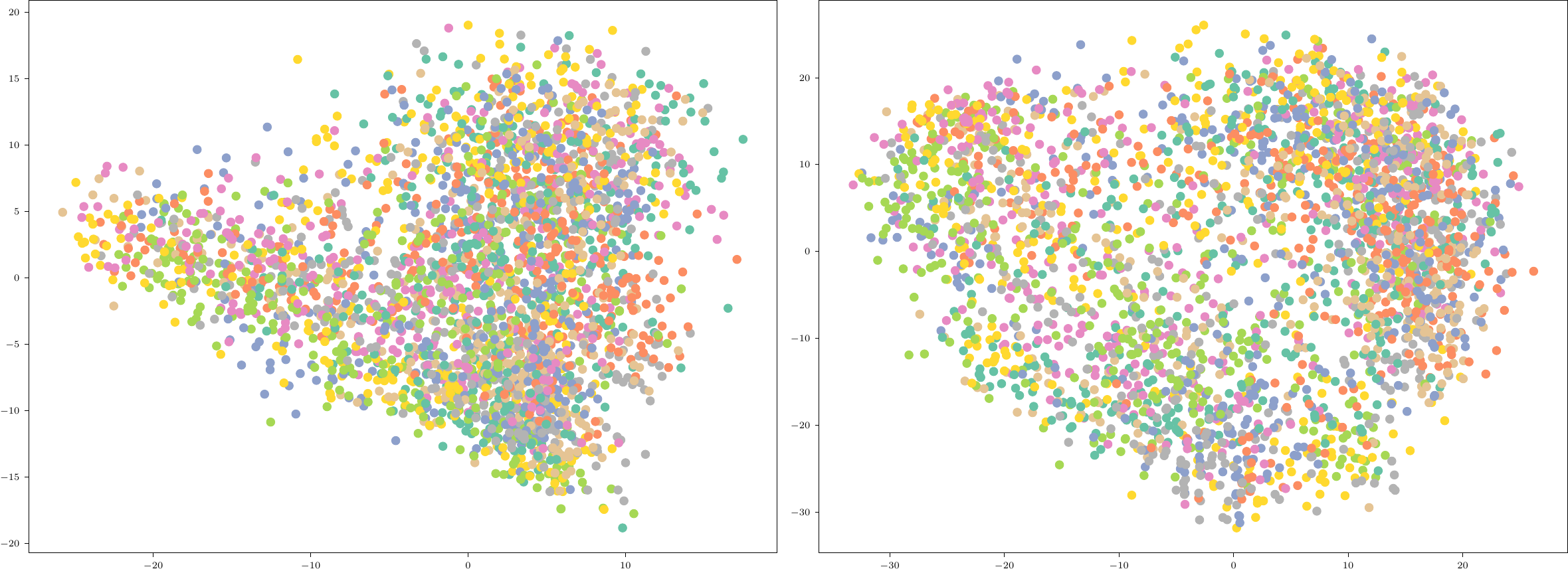}
            \put(-4,6){\rotatebox{90}{\cifarh{}}}
        \end{overpic}        
    \end{minipage}
    \caption{\textbf{Early Block Approximation.} \gls{pca} visualization of the final layer representations for both the original model and the model with its second block approximated by the preceding one. The representations are generated using the \dinos{} model across four datasets. Note that for \cifarh{} (bottom right), only the overall structure of the space can be observed, as the 100 classes make it challenging to distinguish labels based on color.}
    \label{fig:app-pca-approx-dino-1}
\end{figure}

\begin{figure}[h]
    \centering  
    \begin{minipage}[t]{.4\textwidth}
    \centering
        \begin{overpic}[width=\textwidth]{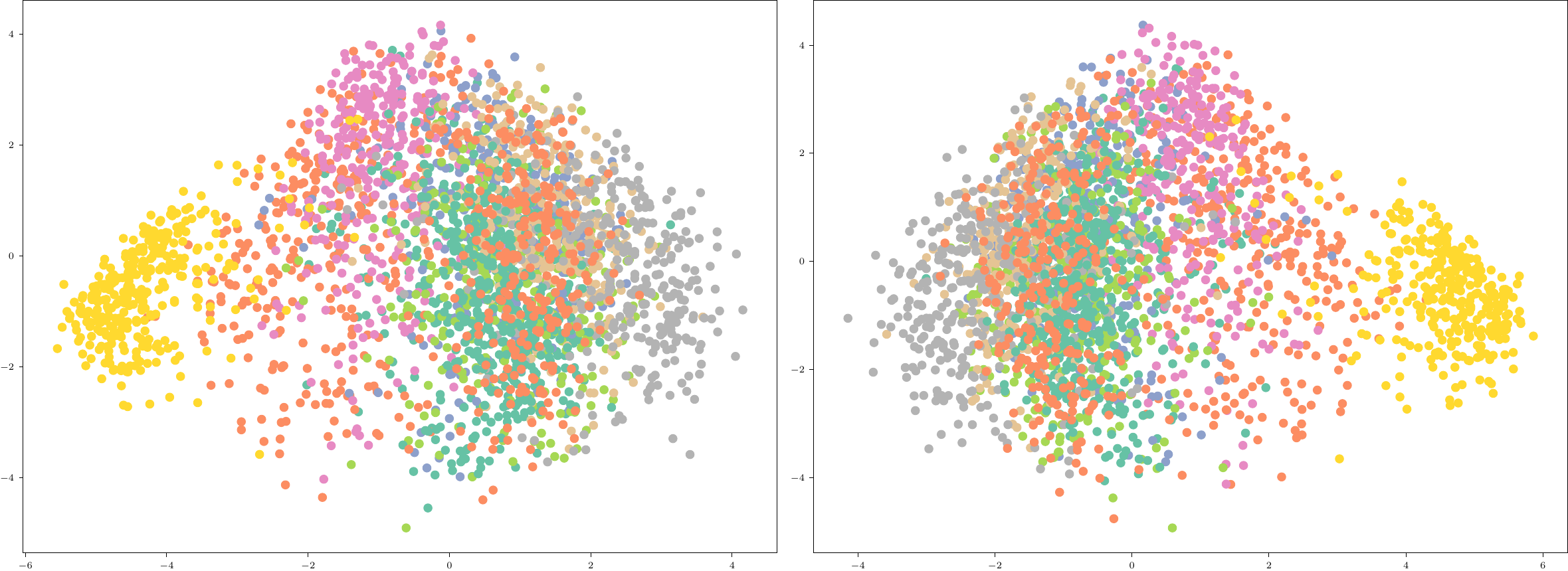}
            \put(-4,11){\rotatebox{90}{\mnist{}}}
            \put(15,38){\texttt{Original}}
            \put(68.5,38){\texttt{TOAST}} 
        \end{overpic}
    \end{minipage}
    \hspace{1em}
    \begin{minipage}[t]{.4\textwidth}
    \centering
        \begin{overpic}[width=\textwidth]{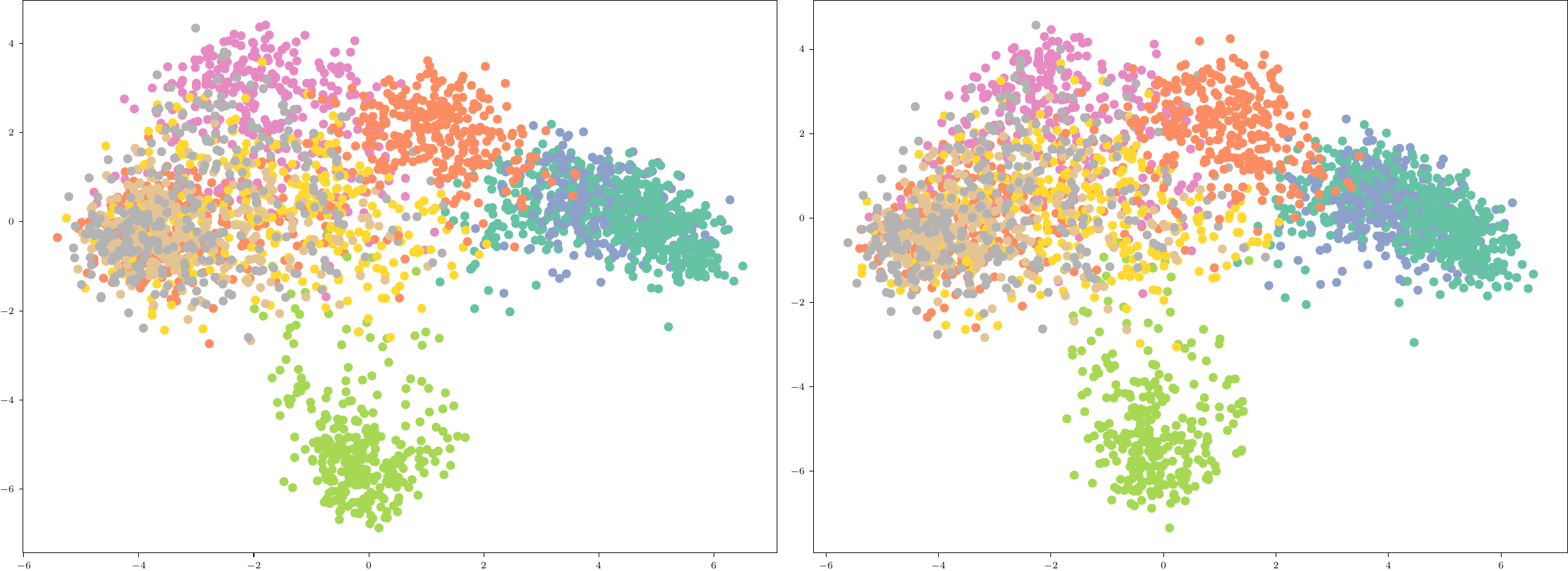}
            \put(-4,8){\rotatebox{90}{\fmnist{}}}
            \put(15,38){\texttt{Original}}
            \put(68.5,38){\texttt{TOAST}} 
        \end{overpic}
        \vspace{.1em}
    \end{minipage}
    \begin{minipage}[t]{.4\textwidth}
        \begin{overpic}[width=\textwidth]{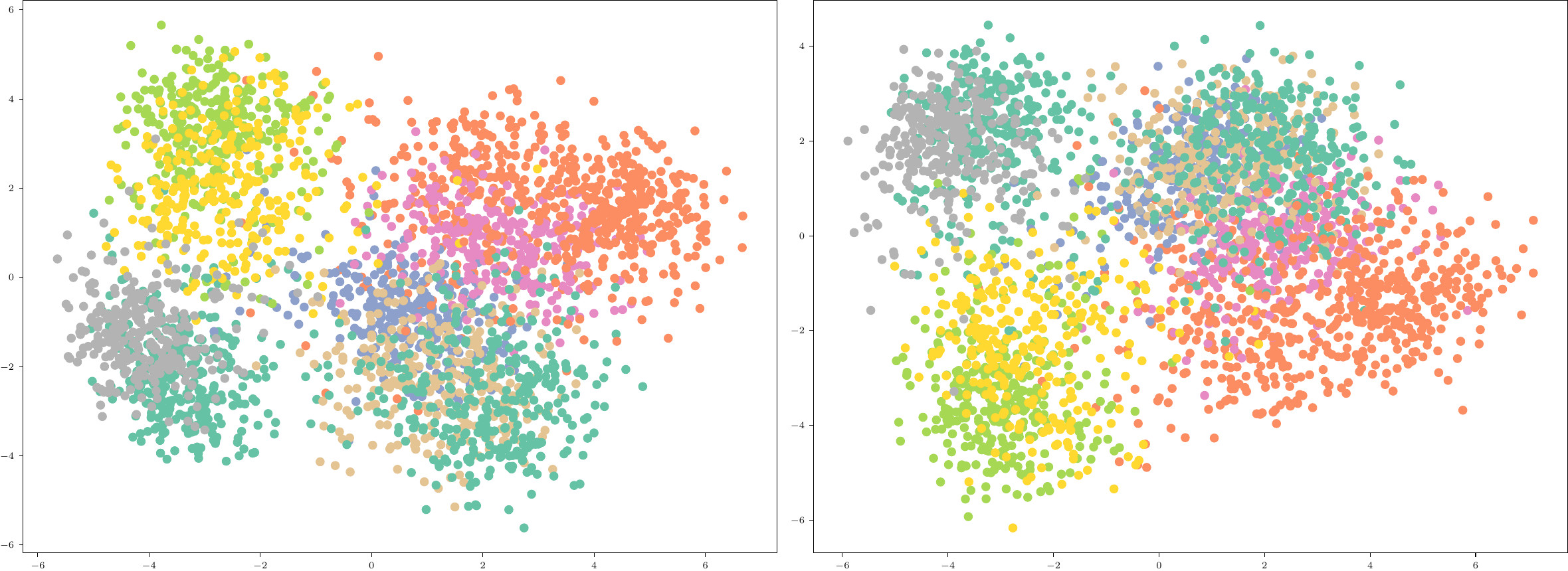}
            \put(-5,3.5){\rotatebox{90}{\cifart{}}}
        \end{overpic}        
    \end{minipage}
    \hspace{1em}
    \begin{minipage}[t]{.4\textwidth}
        \begin{overpic}[width=\textwidth]{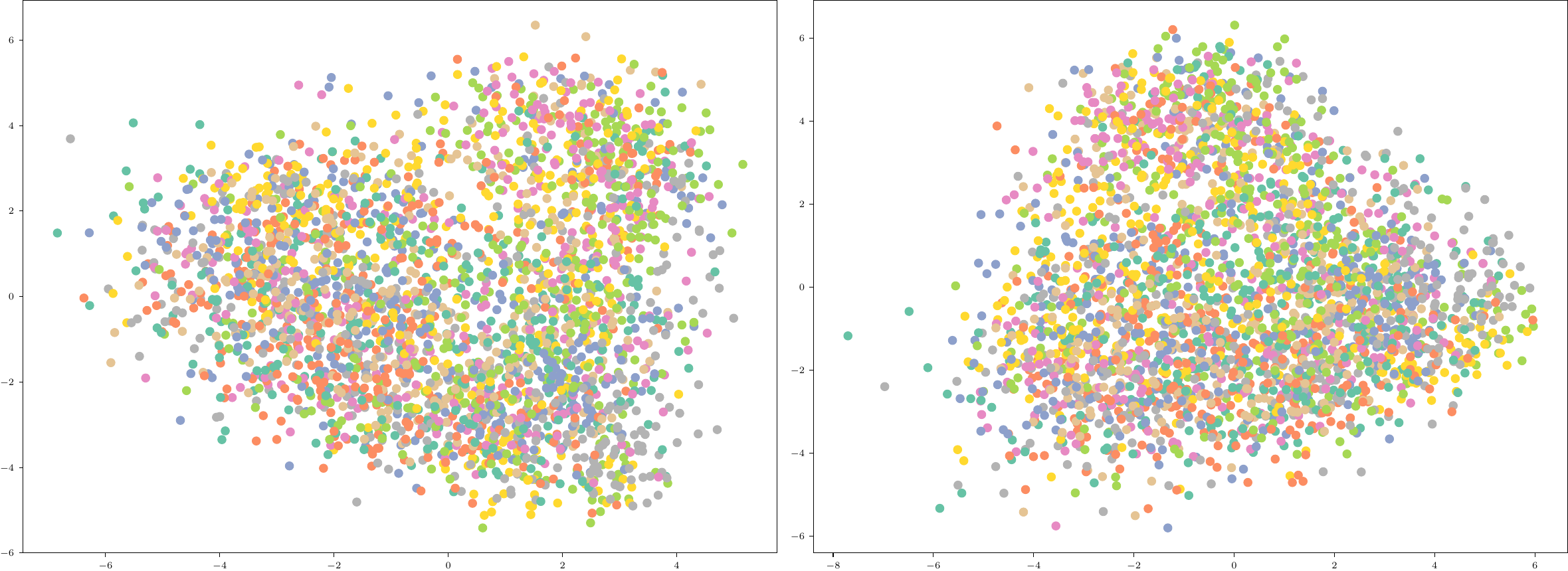}
            \put(-4,6){\rotatebox{90}{\cifarh{}}}
        \end{overpic}        
    \end{minipage}
    \caption{\textbf{Early Block Approximation.} \gls{pca} visualization of the last layer representations for both the original model and the model with its second block approximated using the previous one. Representations are obtained using the \vits{} model across four datasets.}
    \label{fig:app-pca-approx-vit-1}
\end{figure}

\begin{figure}[h]
    \centering  
    \begin{minipage}[t]{.4\textwidth}
    \centering
        \begin{overpic}[width=\textwidth]{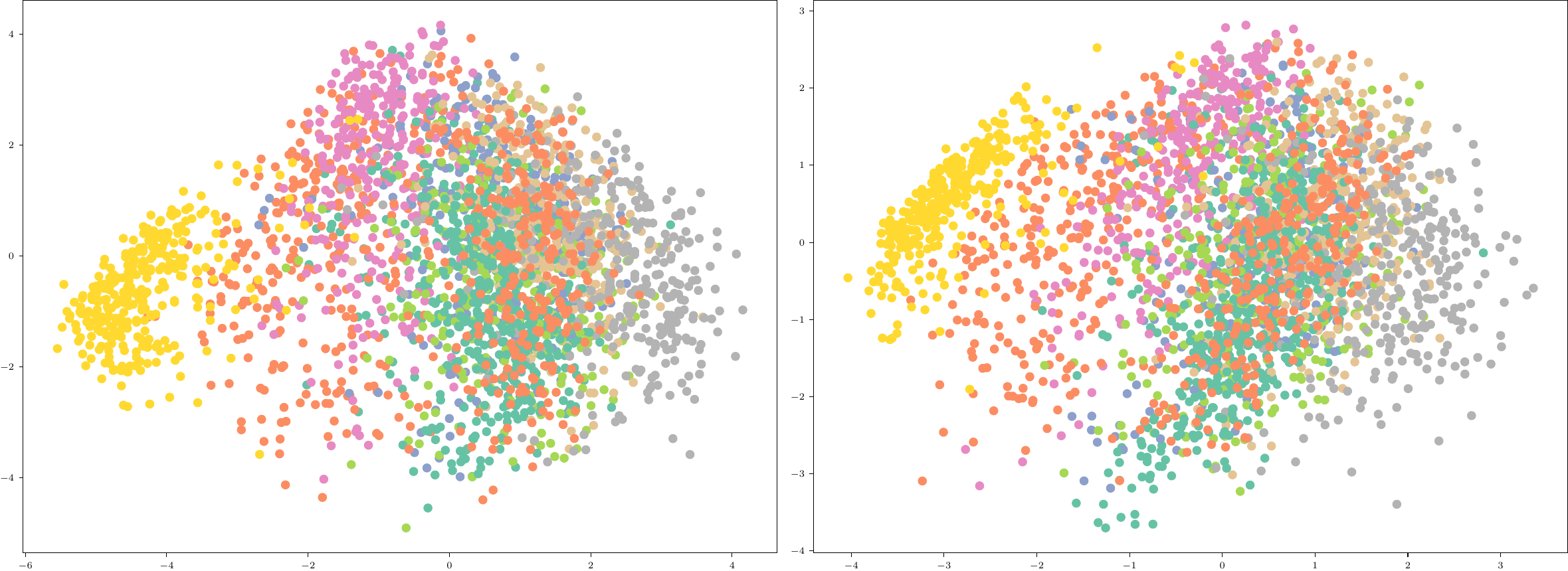}
            \put(-4,11){\rotatebox{90}{\mnist{}}}
            \put(15,38){\texttt{Original}}
            \put(68.5,38){\texttt{TOAST}} 
        \end{overpic}
    \end{minipage}
    \hspace{1em}
    \begin{minipage}[t]{.4\textwidth}
    \centering
        \begin{overpic}[width=\textwidth]{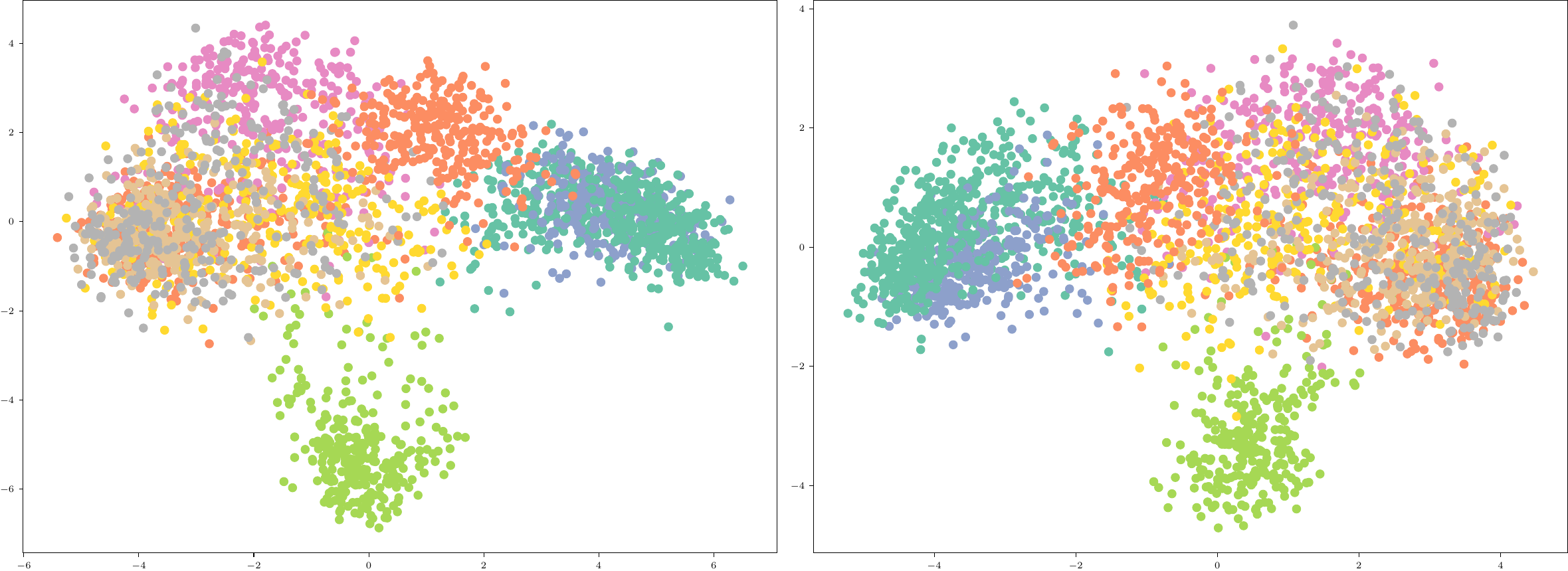}
            \put(-4,8){\rotatebox{90}{\fmnist{}}}
            \put(15,38){\texttt{Original}}
            \put(68.5,38){\texttt{TOAST}} 
        \end{overpic}
        \vspace{.1em}
    \end{minipage}
    \begin{minipage}[t]{.4\textwidth}
        \begin{overpic}[width=\textwidth]{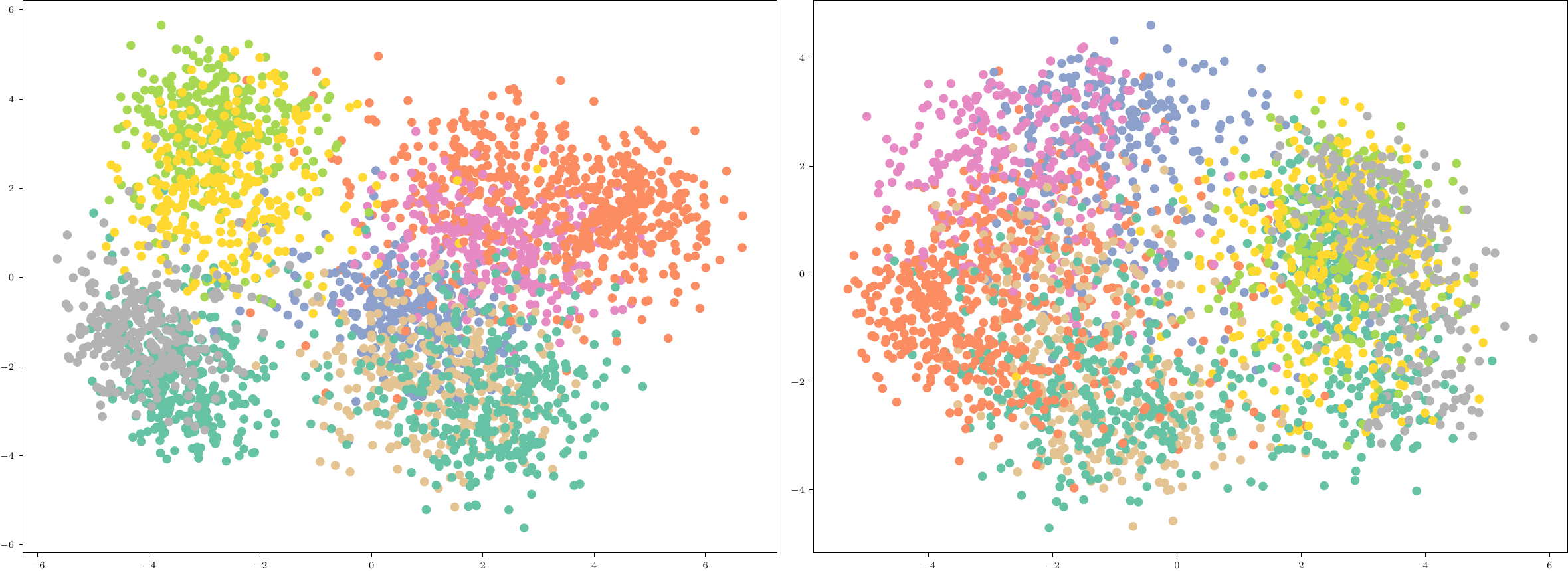}
            \put(-5,3.5){\rotatebox{90}{\cifart{}}}
        \end{overpic}        
    \end{minipage}
    \hspace{1em}
    \begin{minipage}[t]{.4\textwidth}
        \begin{overpic}[width=\textwidth]{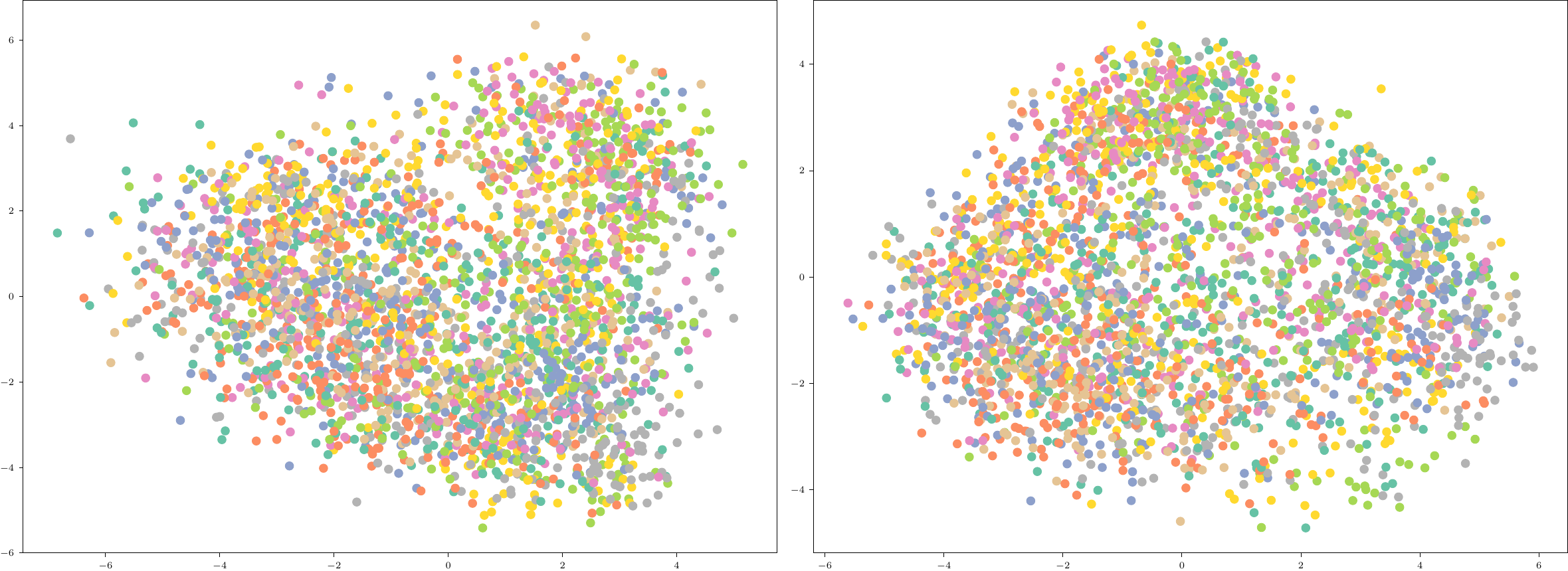}
            \put(-4,6){\rotatebox{90}{\cifarh{}}}
        \end{overpic}        
    \end{minipage}
    \caption{\textbf{Last Block Approximation.} \gls{pca} visualization of the last layer representations for both the original model and the model with its last block approximated from the previous one. Representations are obtained using the \vits{} model across four datasets.}
    \label{fig:app-pca-approx-vit-11}
\end{figure}

This section extends the analysis conducted in \Cref{sec:latent-analysis}, to analyze block-wise internal similarities, to additional models of different dimensionality: \vitt{}, \vits{}, \vitb{}, and \dinos{}. Additionally, we provide visualization using \gls{pca} for \dinos{}, \deits{}, \vits{}, with different datasets and approximating both early and late blocks (see \Cref{fig:app-pca-approx-vit-1,fig:app-pca-approx-vit-11,fig:app-pca-approx-dino-1,fig:app-pca-approx-dino-11,fig:app-pca-approx-deit-11}).

\begin{figure}[h]
    \centering  
    \vspace{1.5em}
    \begin{minipage}[t]{.18\textwidth}  
        \centering
        \begin{overpic}[width=\textwidth]{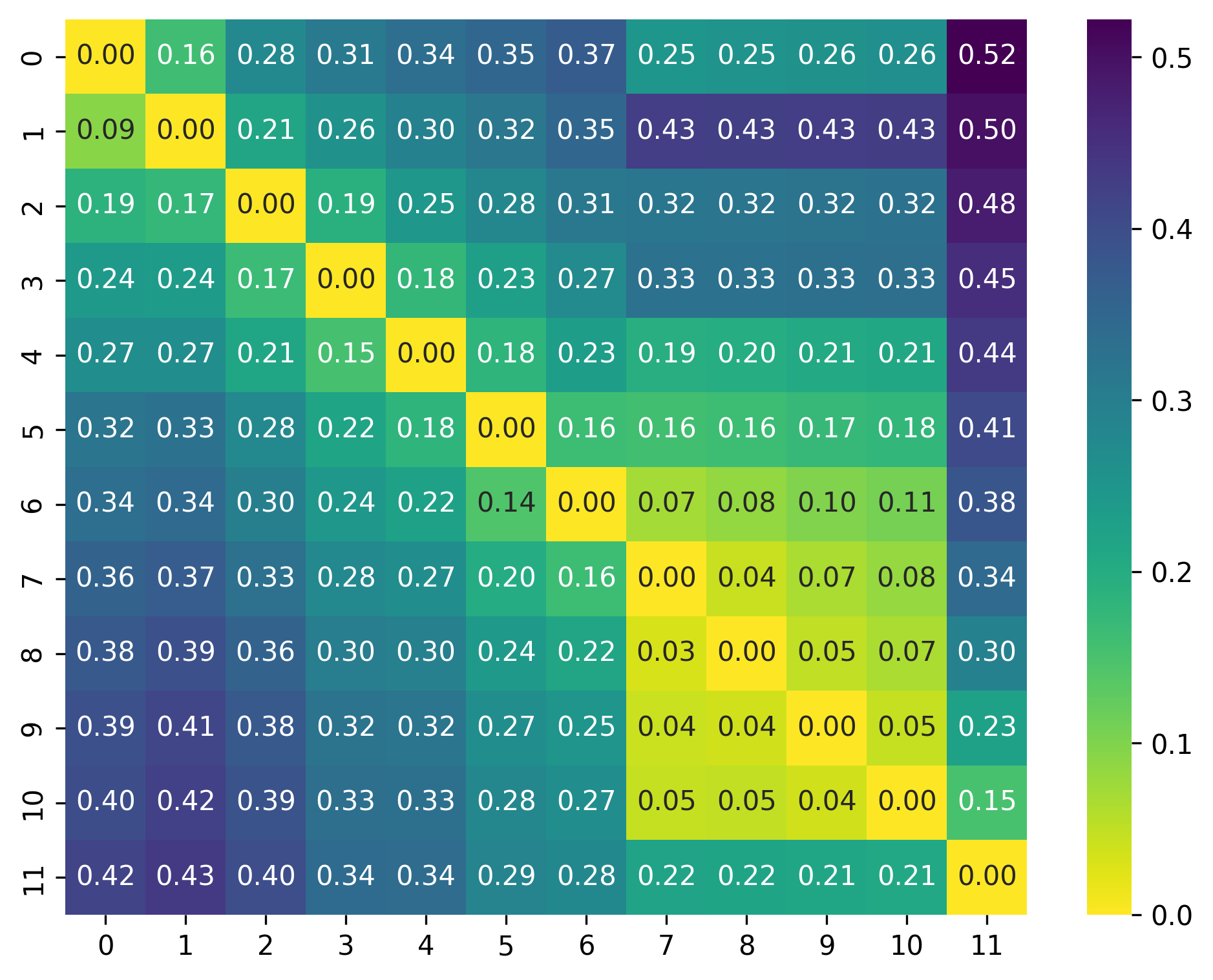}
            \put(-12,26){\rotatebox{90}{\vitt{}}} 
            \put(30,84){\mnist{}} 
        \end{overpic}
    \end{minipage}
    \begin{minipage}[t]{.18\textwidth}
        \centering
        \begin{overpic}[width=\textwidth]{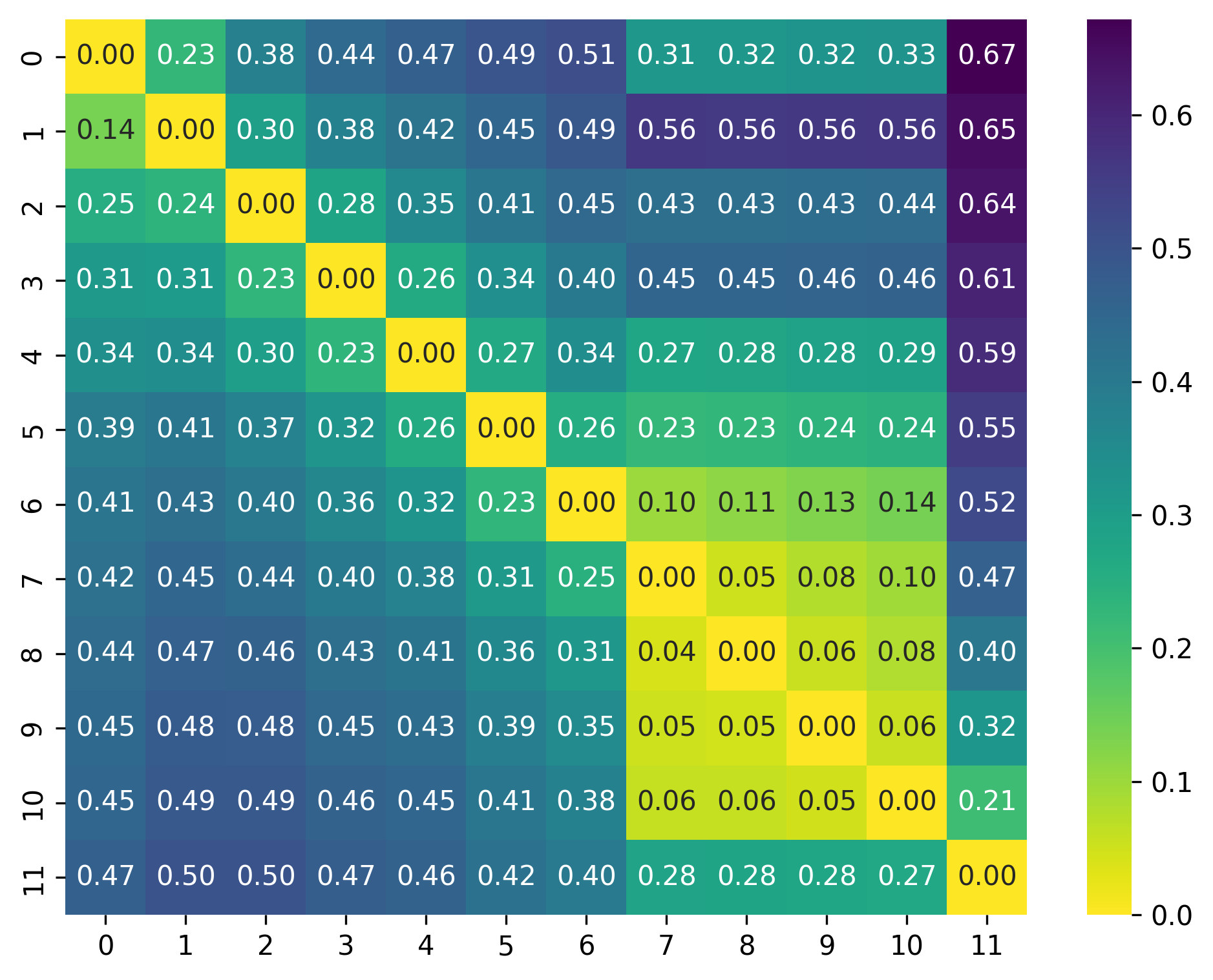}
            \put(24,84){\fmnist{}} 
        \end{overpic}
    \end{minipage}
    \begin{minipage}[t]{.18\textwidth}
        \centering
        \begin{overpic}[width=\textwidth]{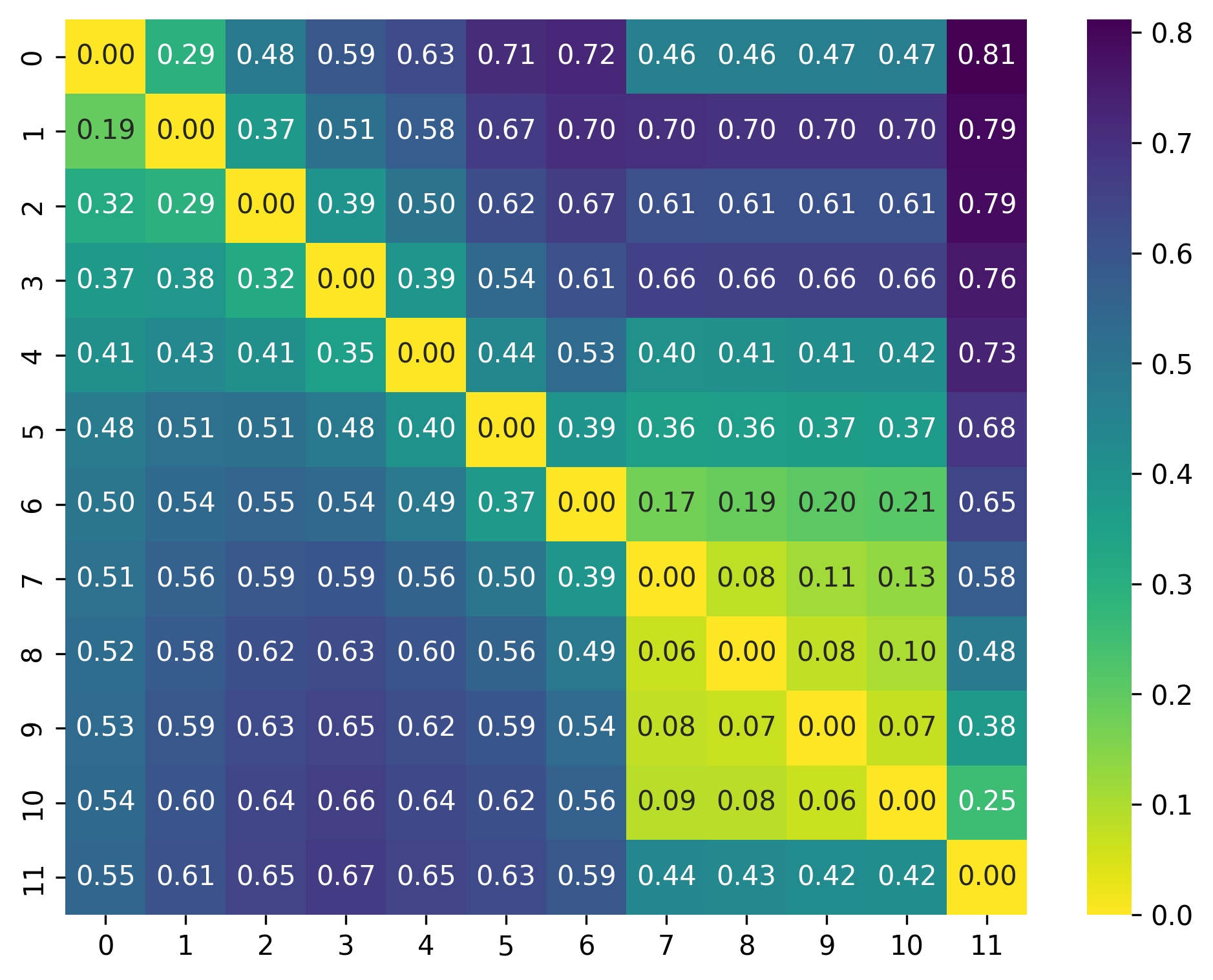}
            \put(21,84){\cifart{}} 
        \end{overpic}
    \end{minipage}
    \begin{minipage}[t]{.18\textwidth}
        \centering
        \begin{overpic}[width=\textwidth]{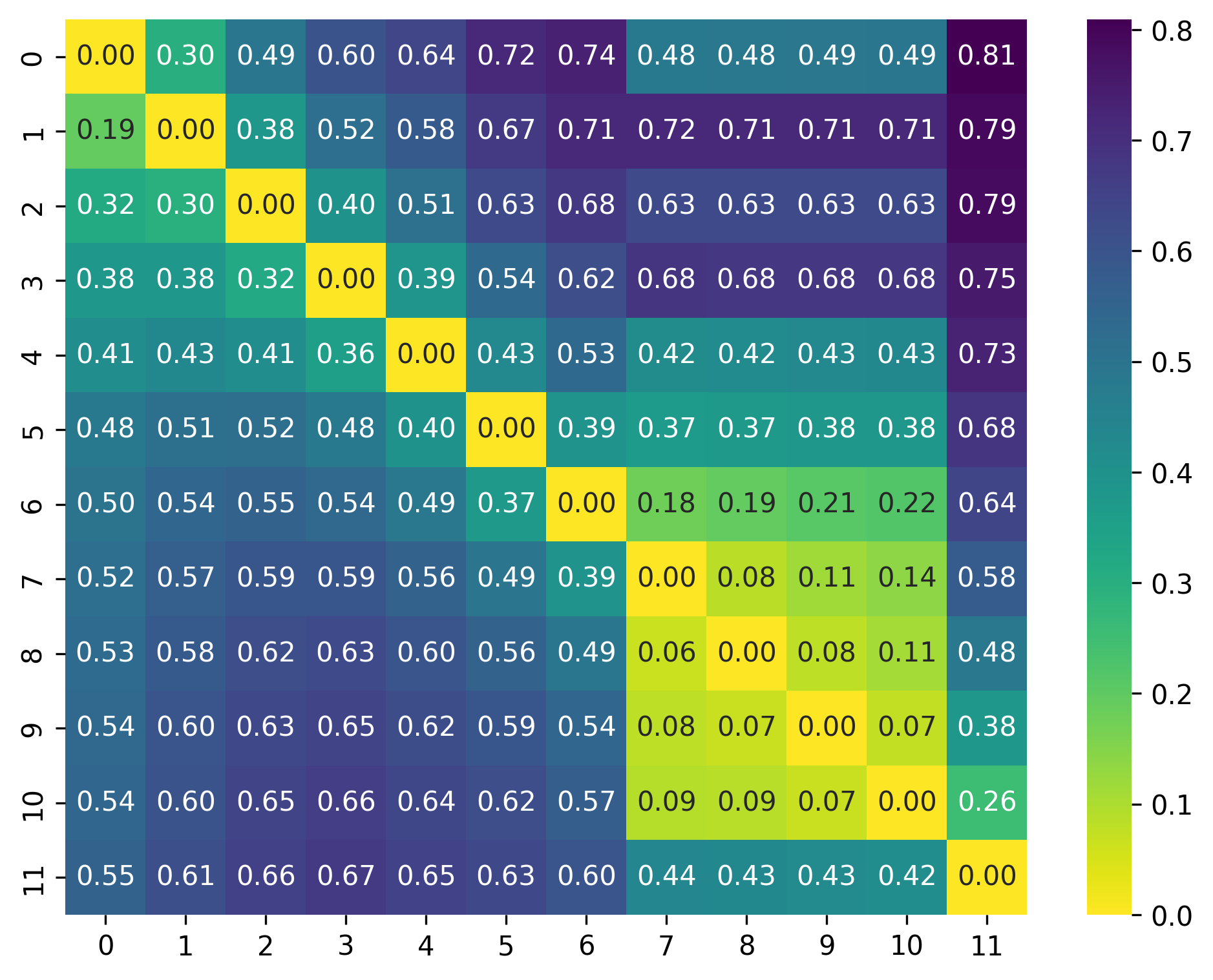}
            \put(17.5,84){\cifarh{}}
        \end{overpic}
    \end{minipage}
    \begin{minipage}[t]{.18\textwidth}
        \centering
        \begin{overpic}[width=\textwidth]{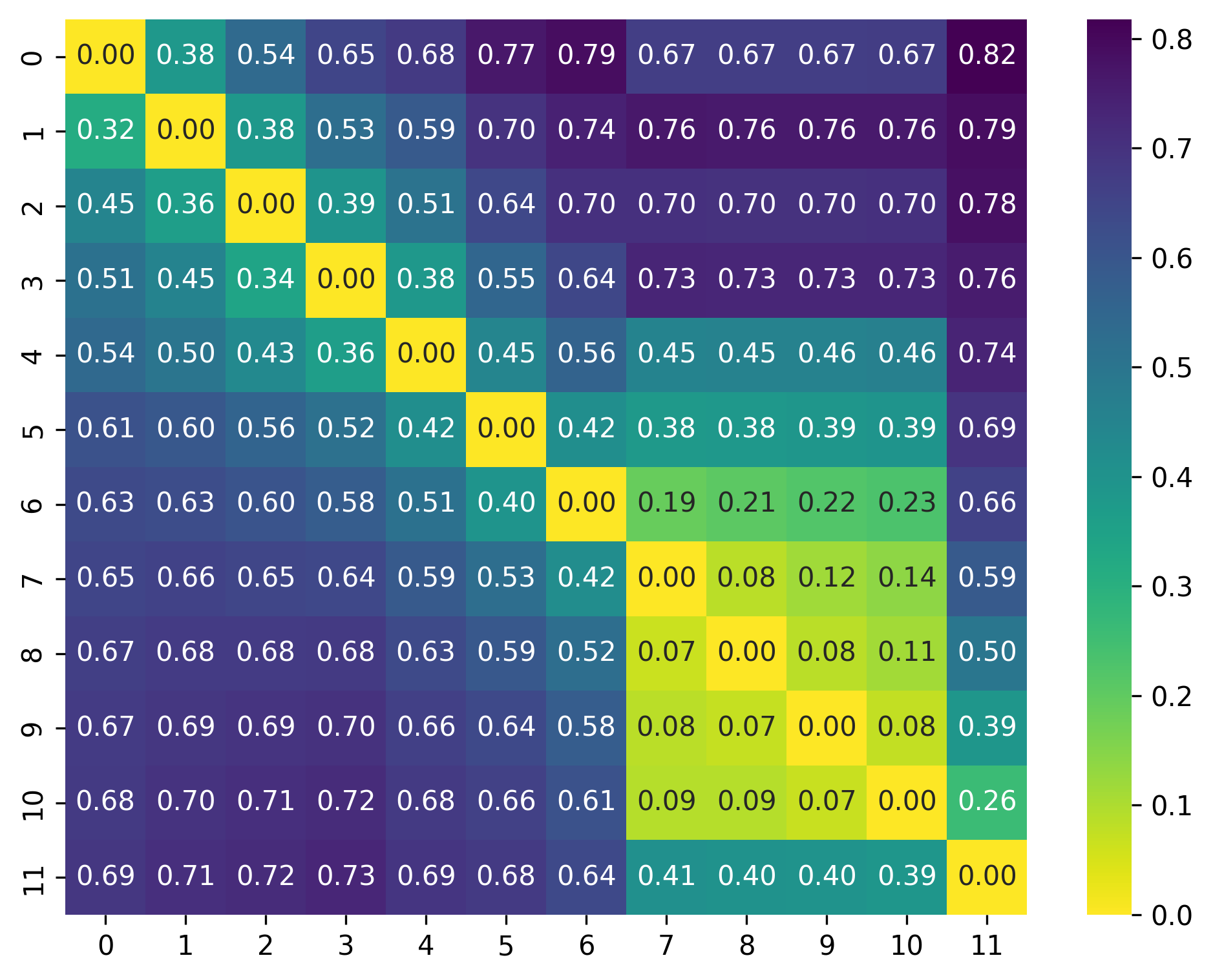}
            \put(14.5,84){\imagenet{}} 
        \end{overpic}
    \end{minipage}
    
    \begin{minipage}[t]{.18\textwidth}  
        \centering
        \begin{overpic}[width=\textwidth]{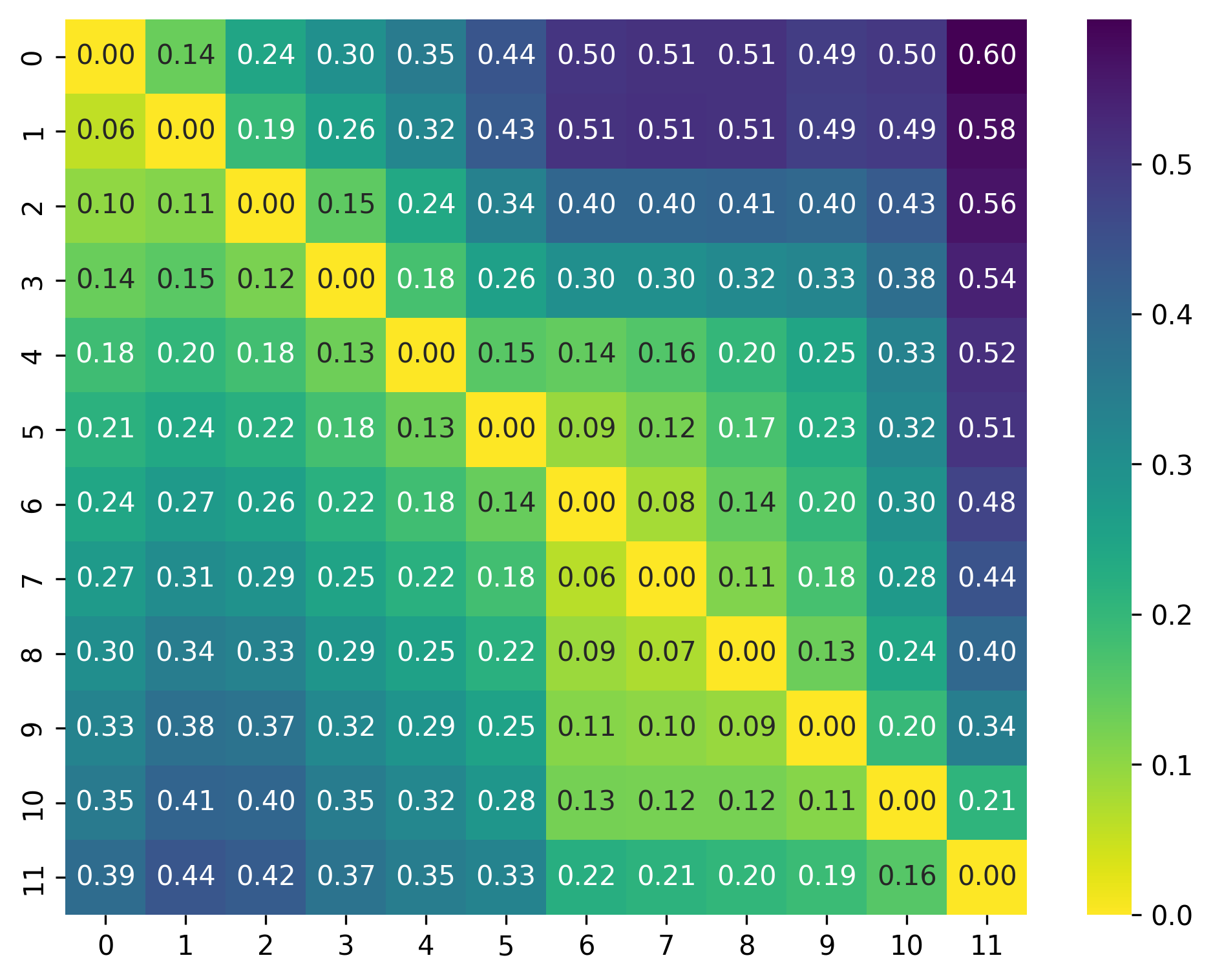}
            \put(-12,26){\rotatebox{90}{\vits{}}} 
        \end{overpic}
    \end{minipage}
    \begin{minipage}[t]{.18\textwidth}
        \centering
        \begin{overpic}[width=\textwidth]{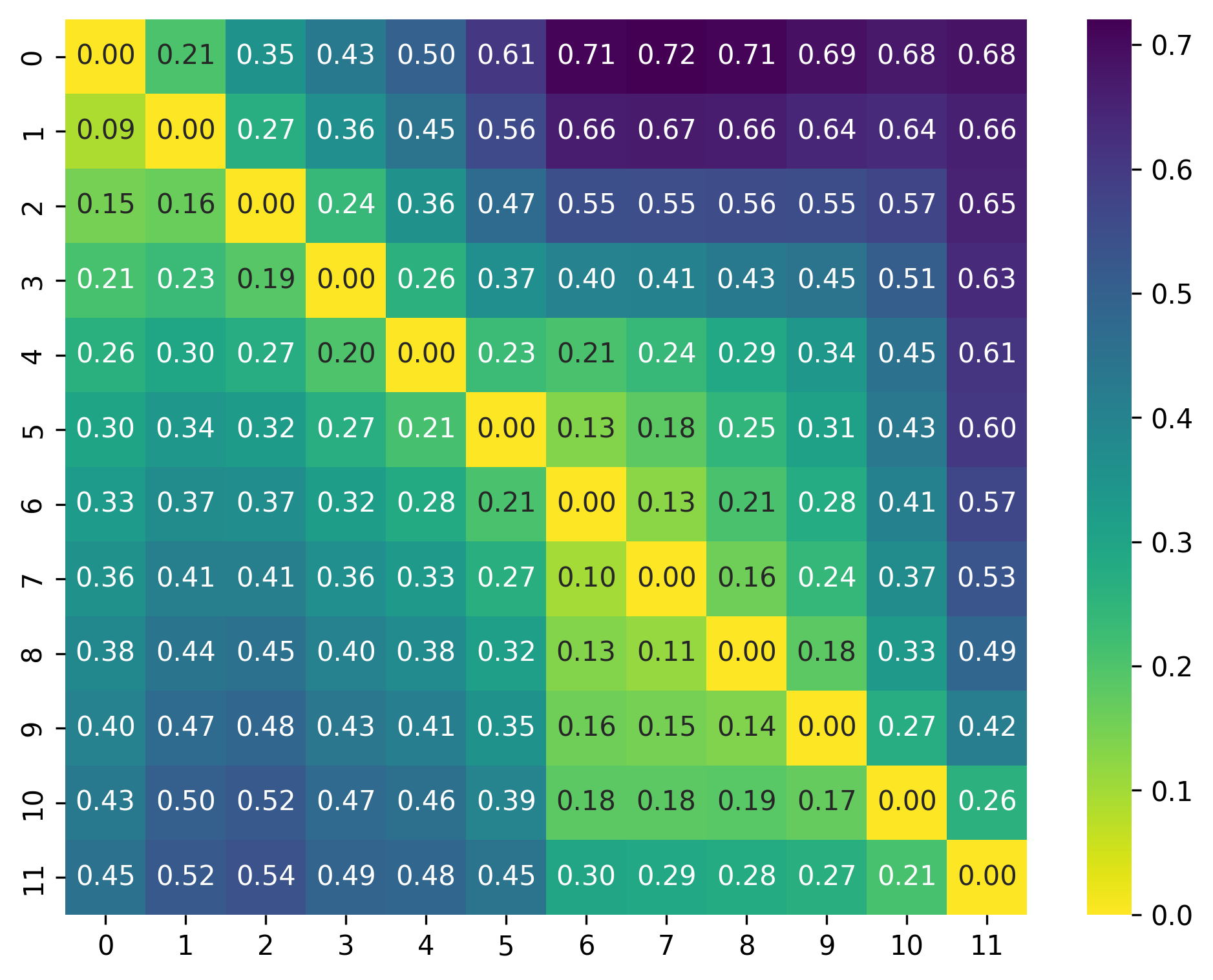}
        \end{overpic}
    \end{minipage}
    \begin{minipage}[t]{.18\textwidth}
        \centering
        \begin{overpic}[width=\textwidth]{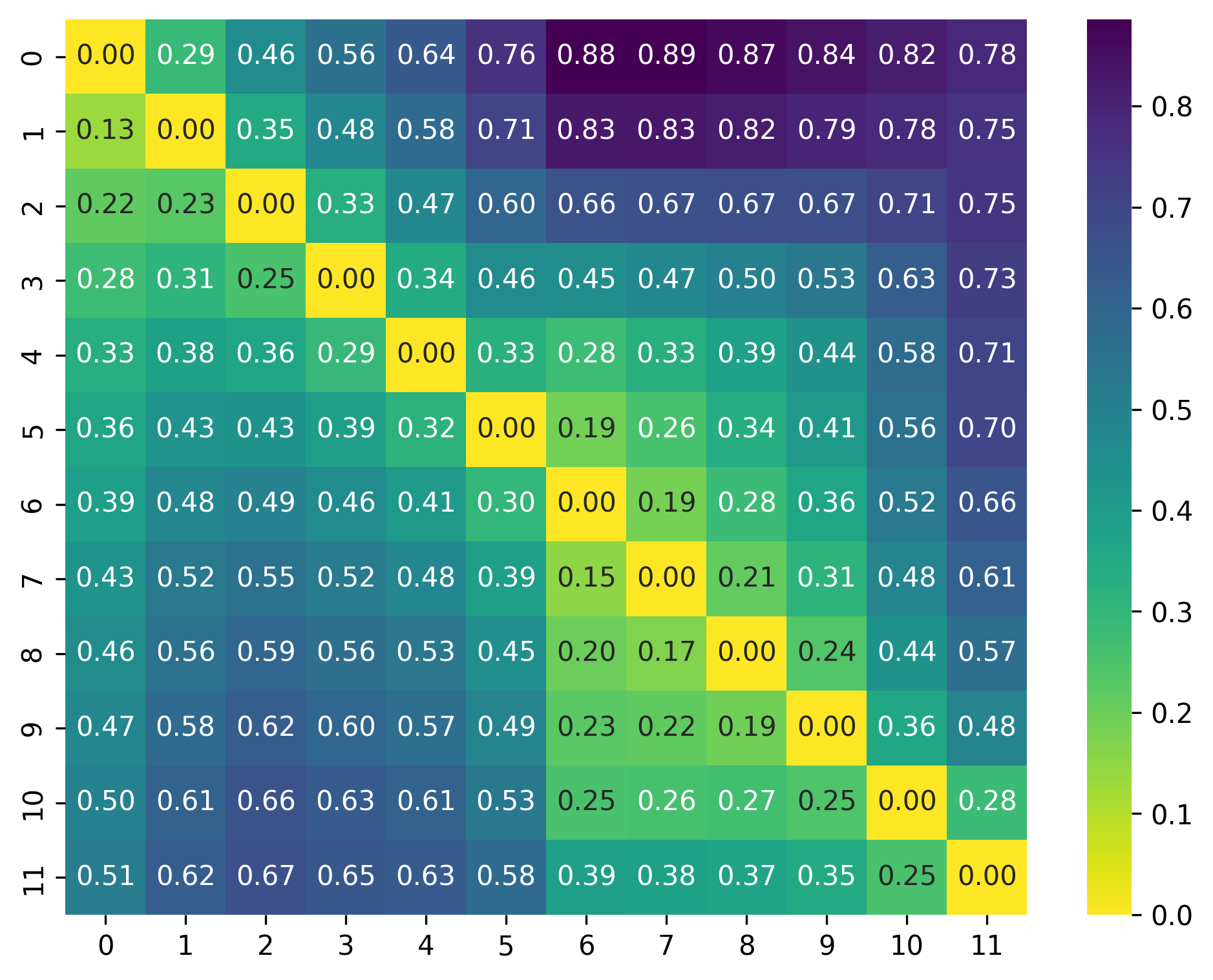}
        \end{overpic}
    \end{minipage}
    \begin{minipage}[t]{.18\textwidth}
        \centering
        \begin{overpic}[width=\textwidth]{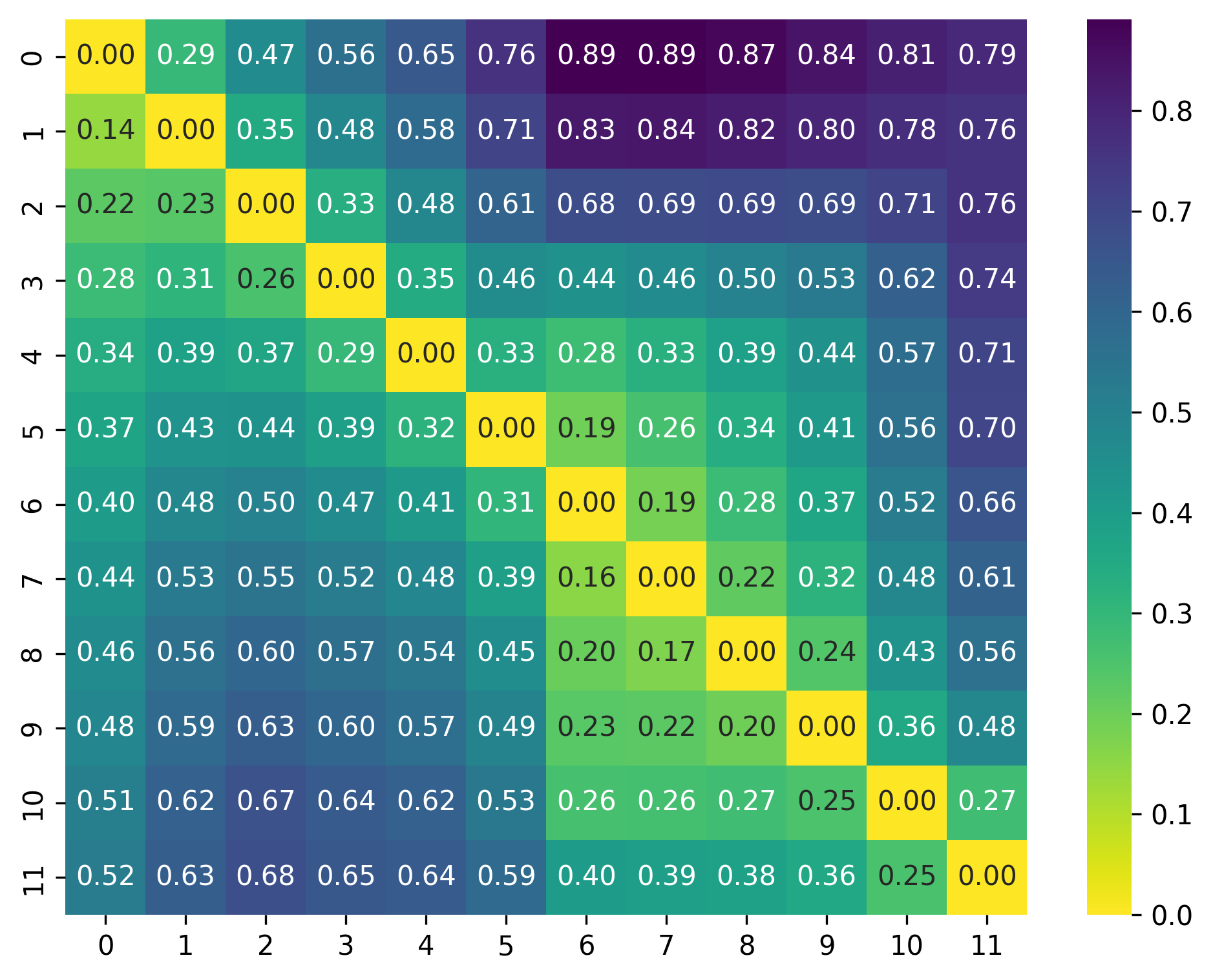}
        \end{overpic}
    \end{minipage}
    \begin{minipage}[t]{.18\textwidth}
        \centering
        \begin{overpic}[width=\textwidth]{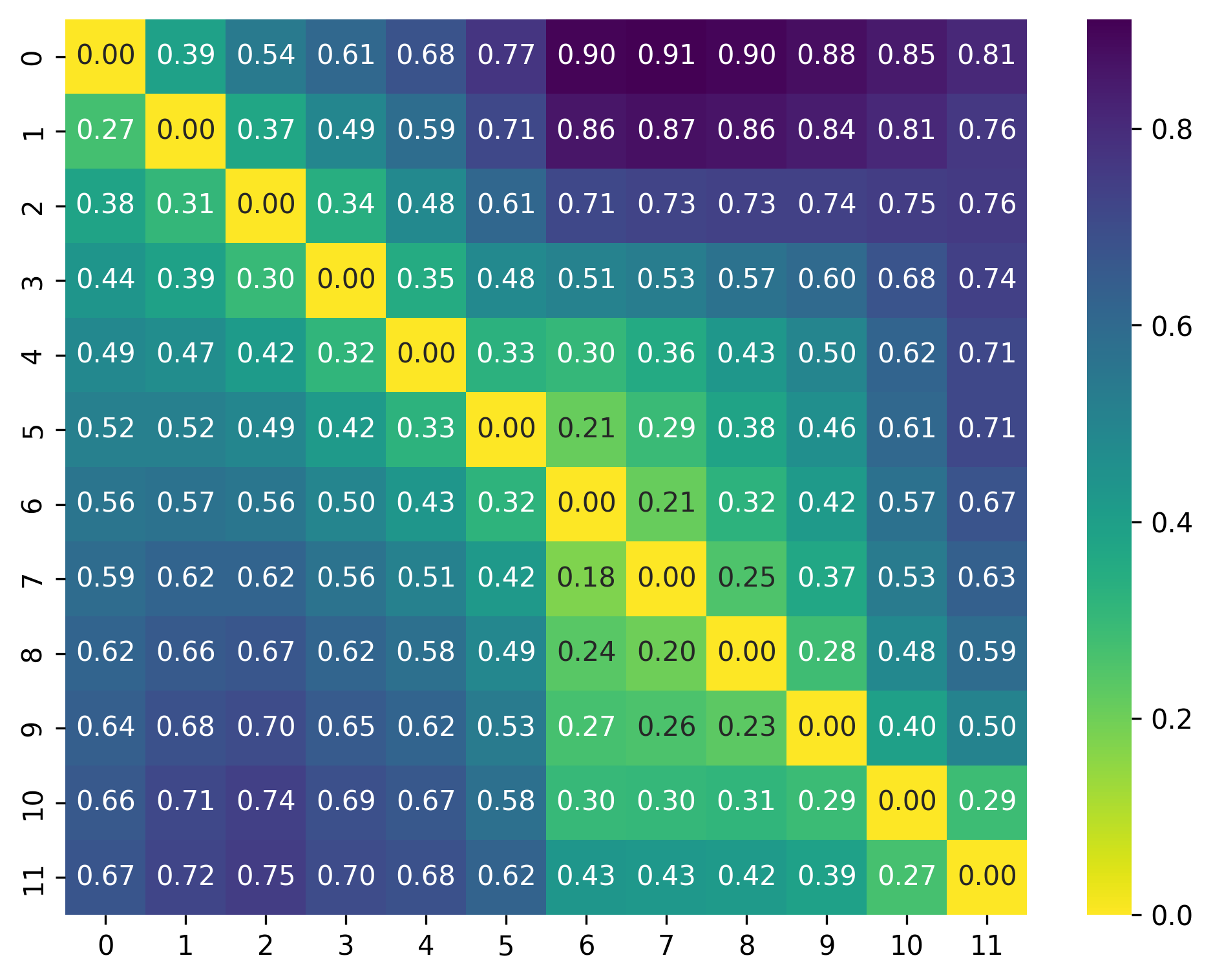}
        \end{overpic}
    \end{minipage}

    \begin{minipage}[t]{.18\textwidth}  
        \centering
        \begin{overpic}[width=\textwidth]{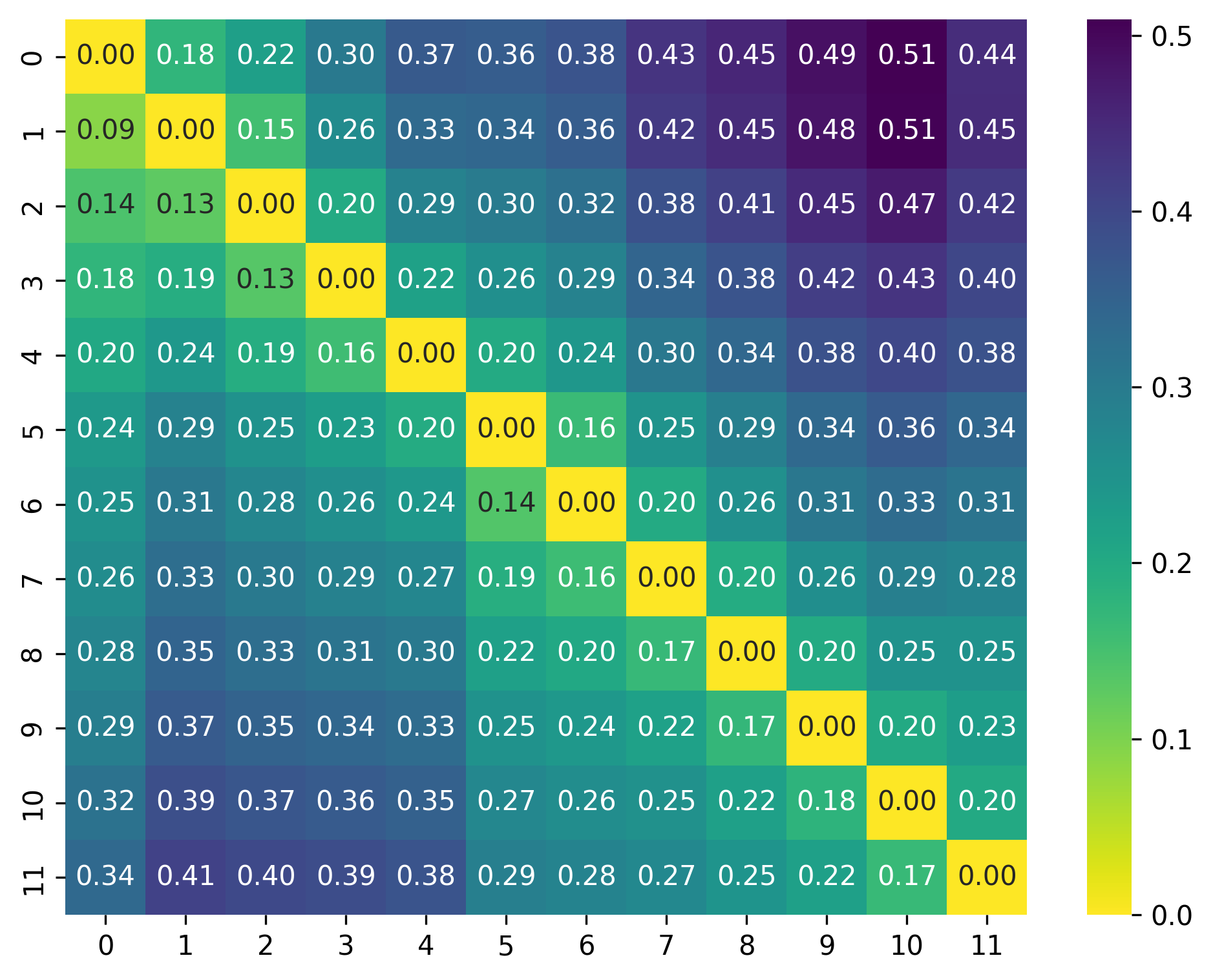}
            \put(-12,24){\rotatebox{90}{\dinos{}}} 
        \end{overpic}
    \end{minipage}
    \begin{minipage}[t]{.18\textwidth}
        \centering
        \includegraphics[width=\textwidth]{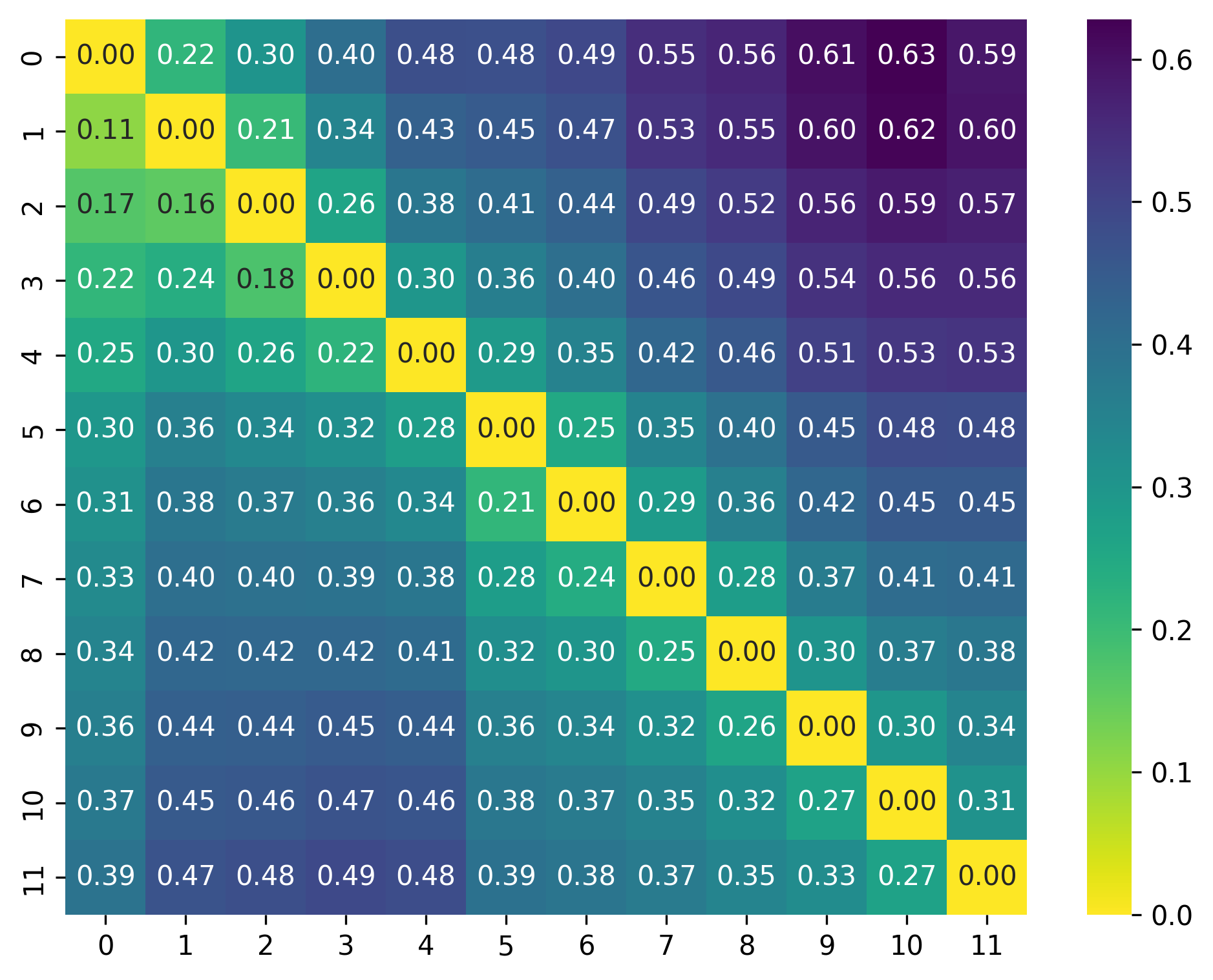}
    \end{minipage}
    \begin{minipage}[t]{.18\textwidth}
        \centering
        \includegraphics[width=\textwidth]{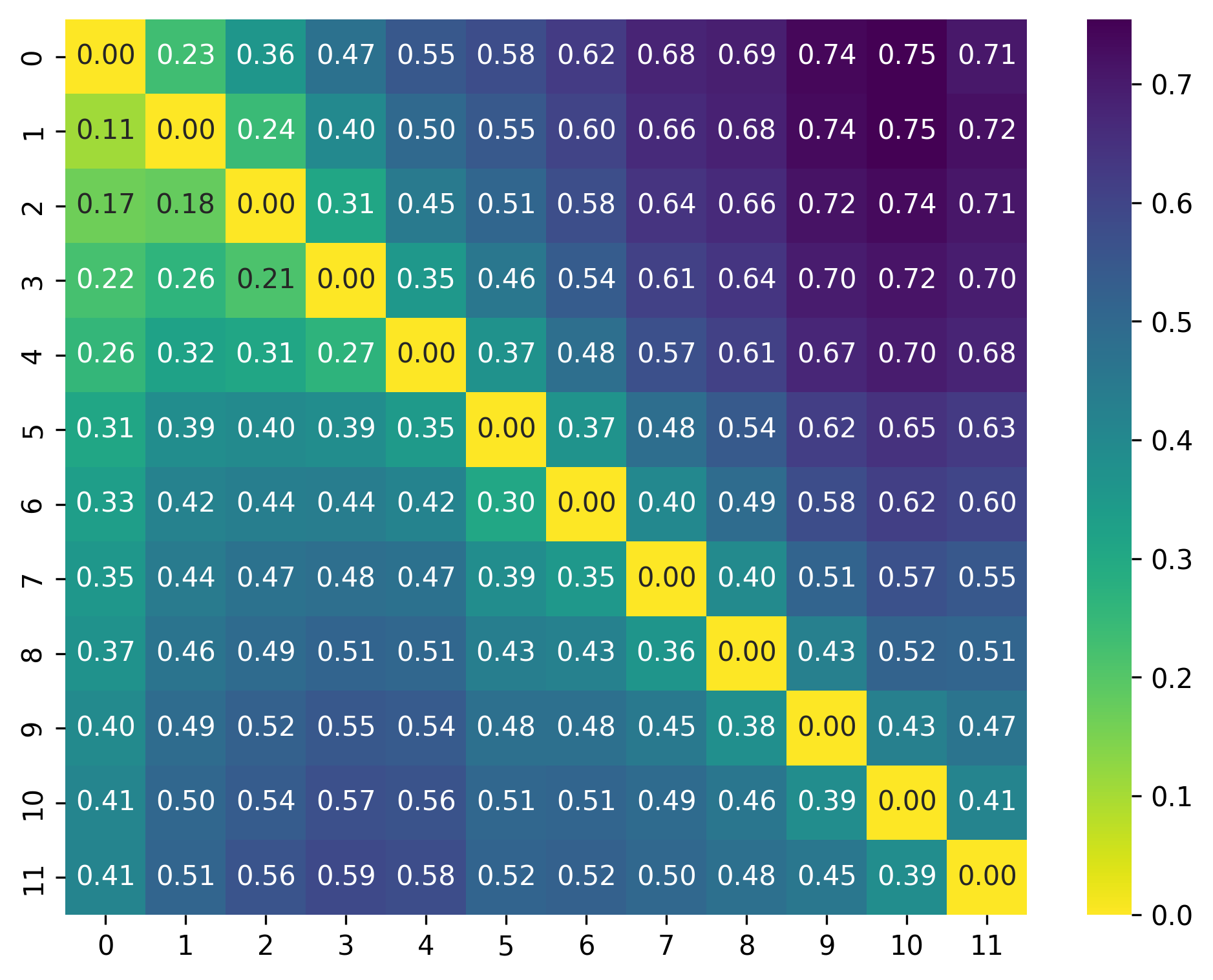}
    \end{minipage}
    \begin{minipage}[t]{.18\textwidth}
        \centering
        \includegraphics[width=\textwidth]{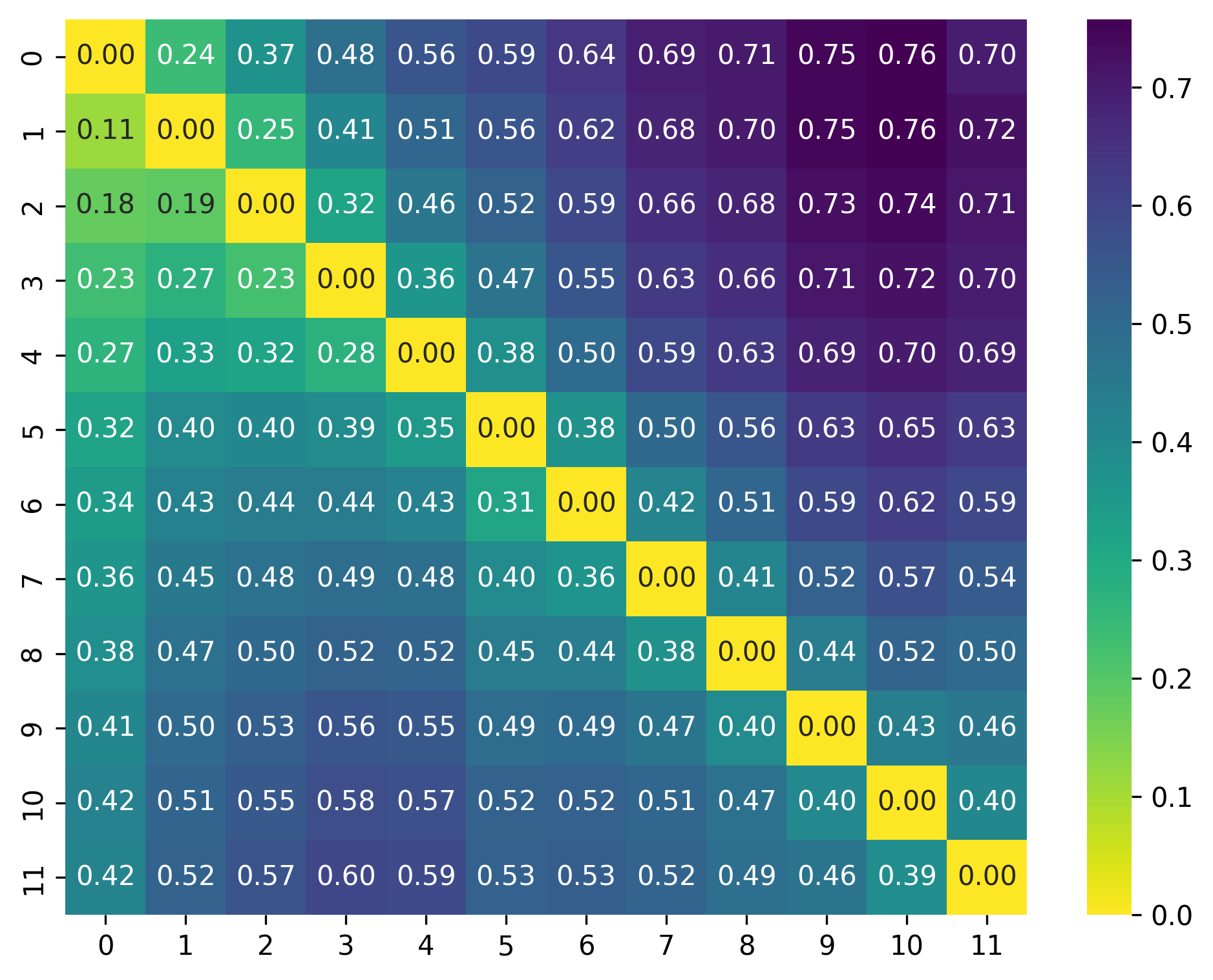}
    \end{minipage}
    \begin{minipage}[t]{.18\textwidth}
        \centering
        \begin{overpic}[width=\textwidth]{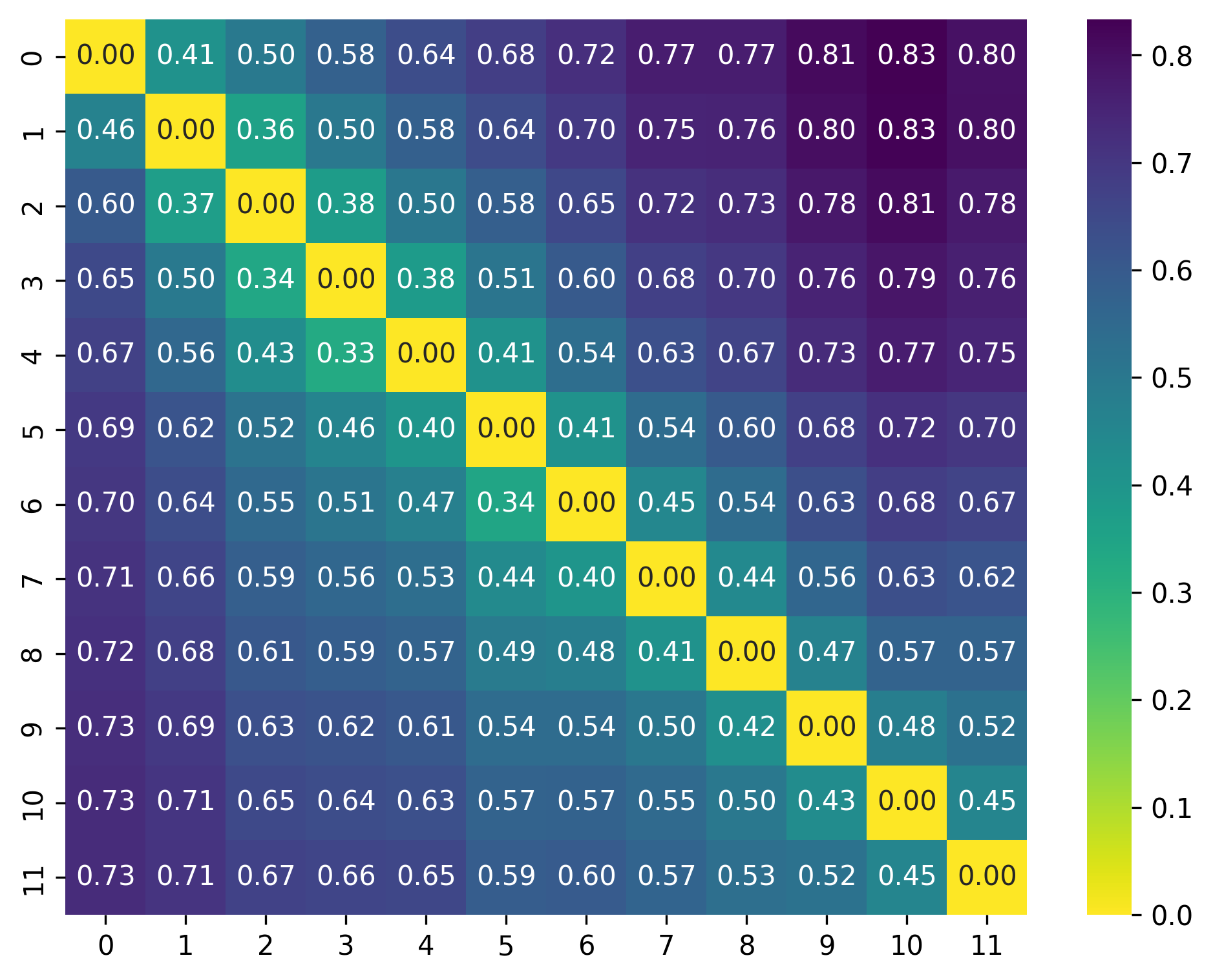}
        \end{overpic}
    \end{minipage}

    \begin{minipage}[t]{.18\textwidth}
        \centering
        \begin{overpic}[width=\textwidth]{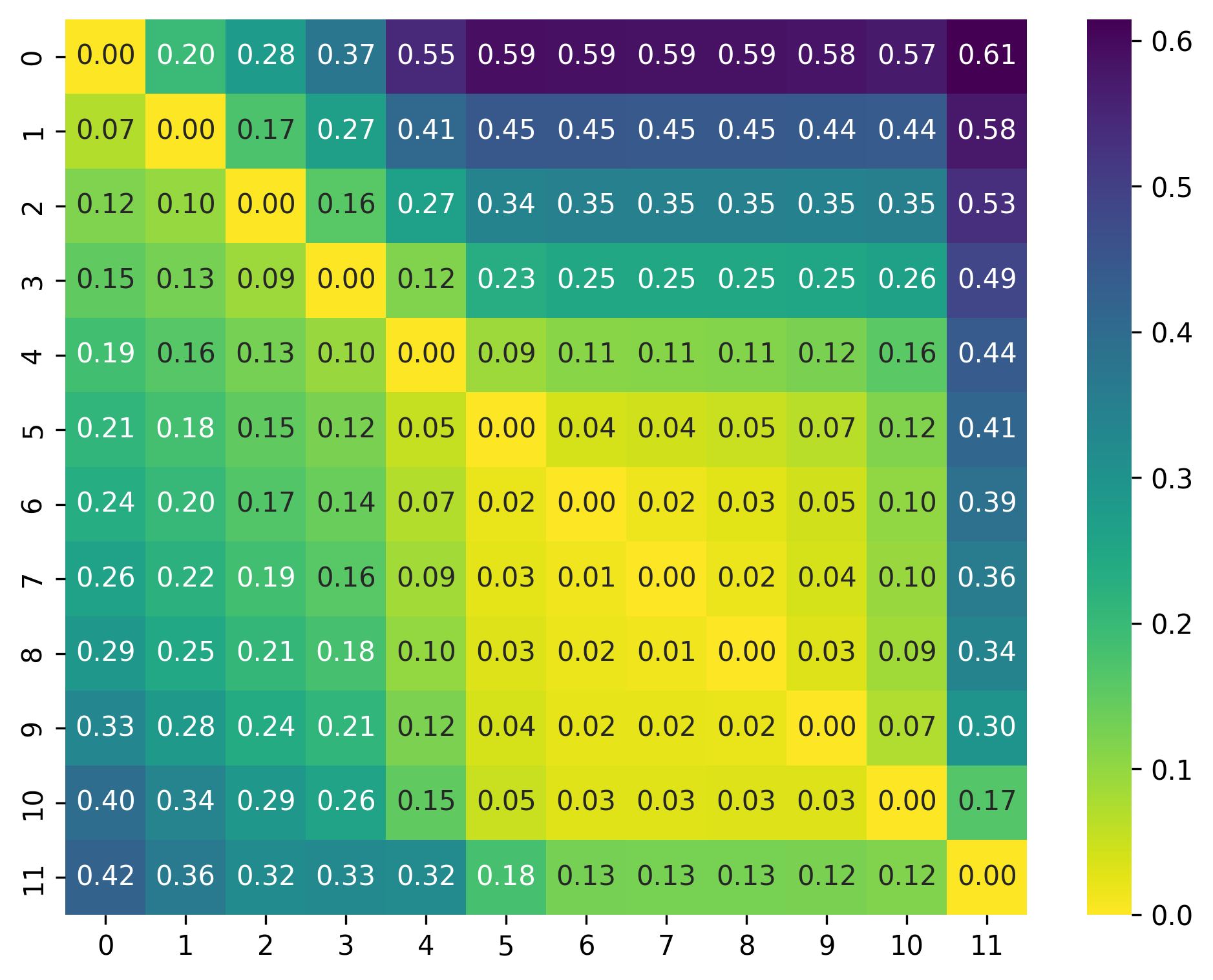}
            \put(-12,26){\rotatebox{90}{\vitb{}}} 
        \end{overpic}
    \end{minipage}
    \begin{minipage}[t]{.18\textwidth}
        \centering
        \includegraphics[width=\textwidth]{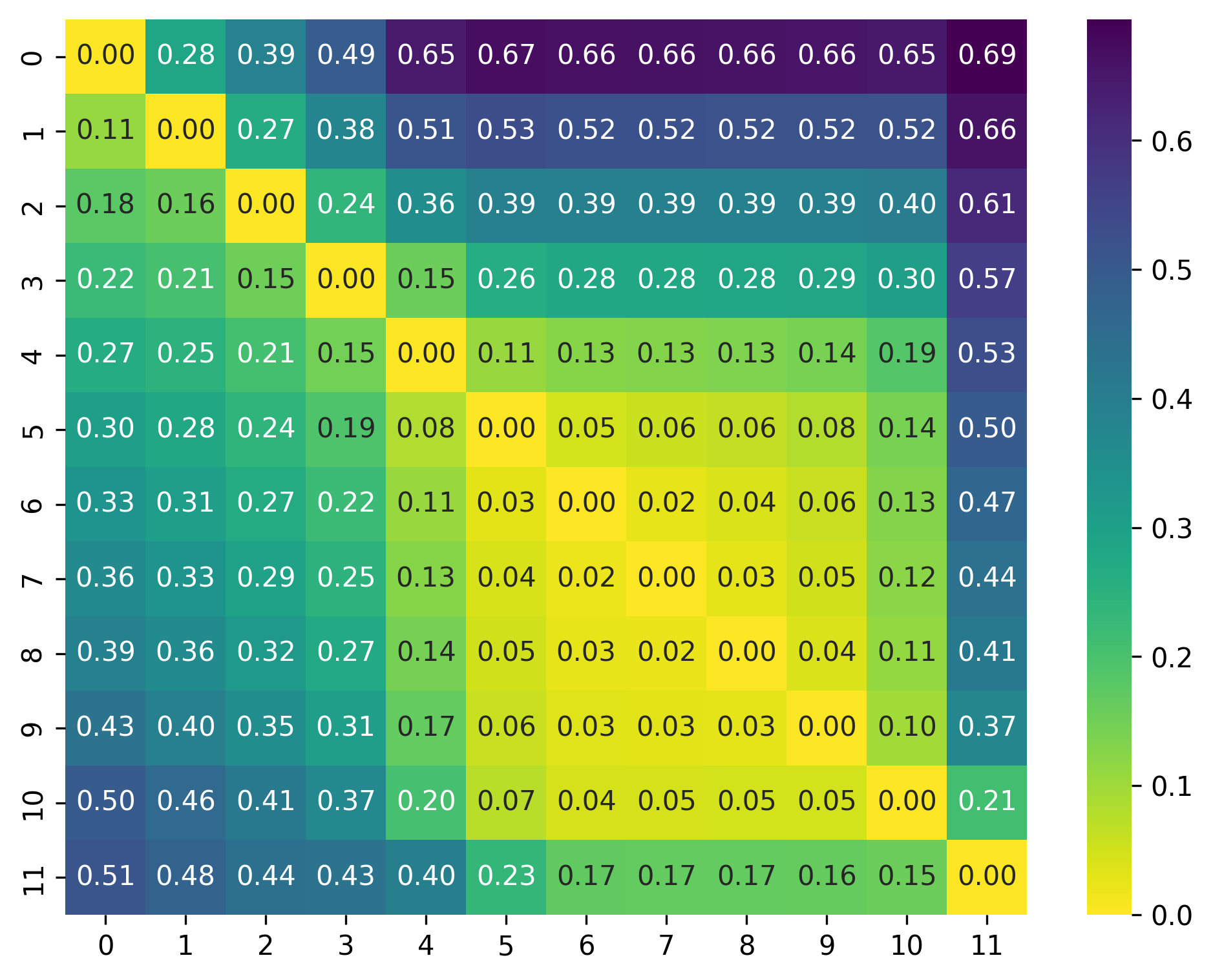}
    \end{minipage}
    \begin{minipage}[t]{.18\textwidth}
        \centering
        \includegraphics[width=\textwidth]{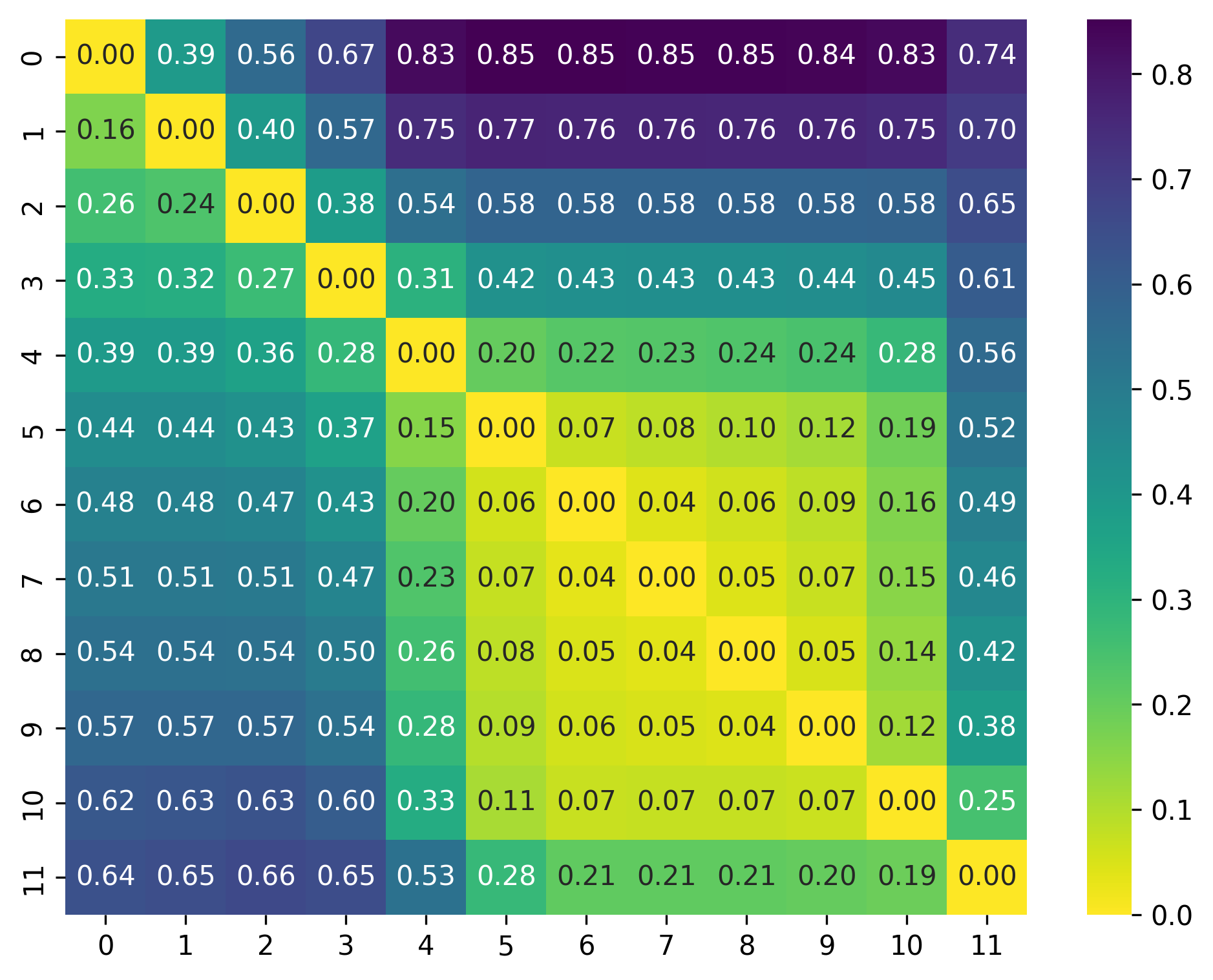}
    \end{minipage}
    \begin{minipage}[t]{.18\textwidth}
        \centering
        \includegraphics[width=\textwidth]{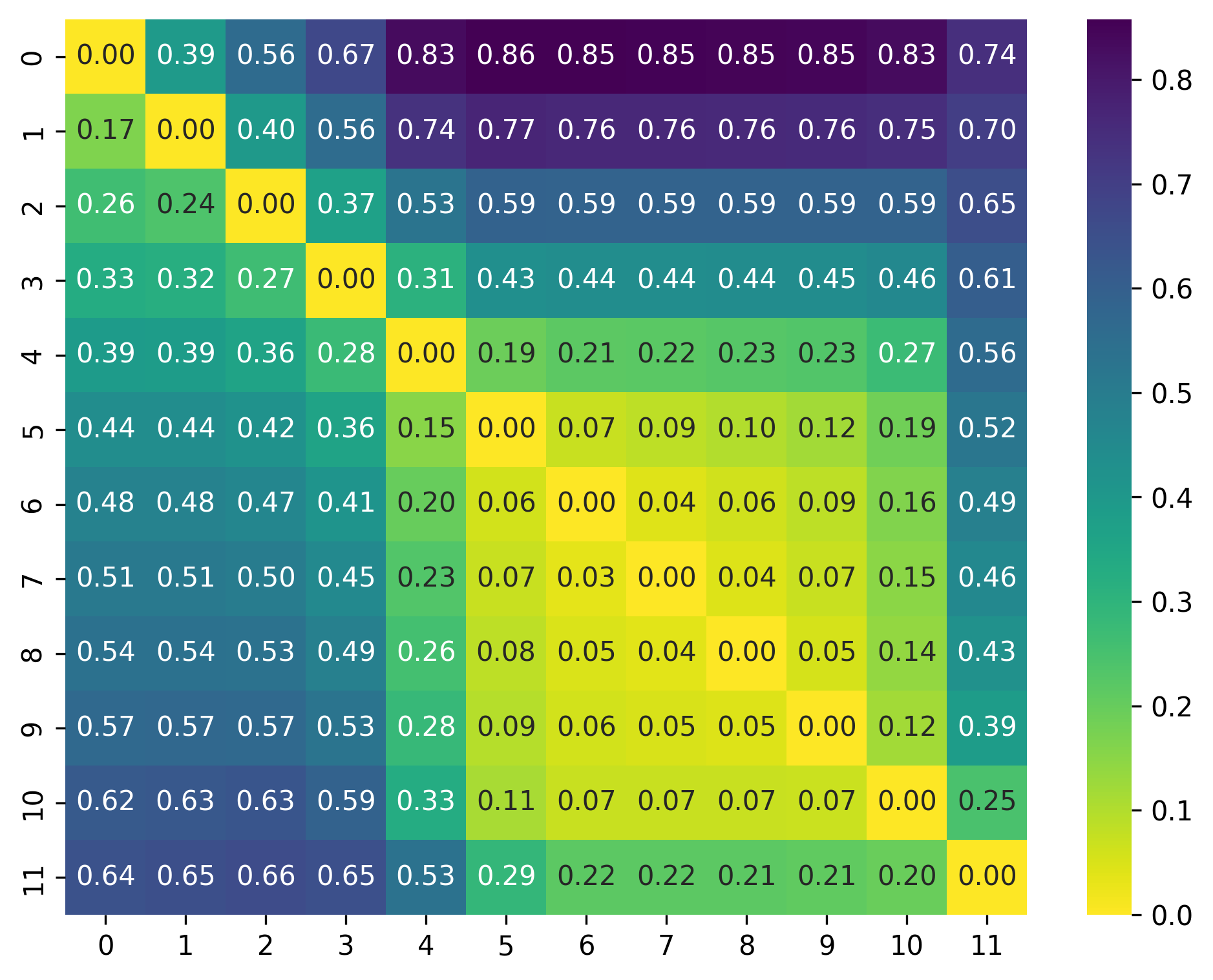}
    \end{minipage}
    \begin{minipage}[t]{.18\textwidth}
        \centering
        \includegraphics[width=\textwidth]{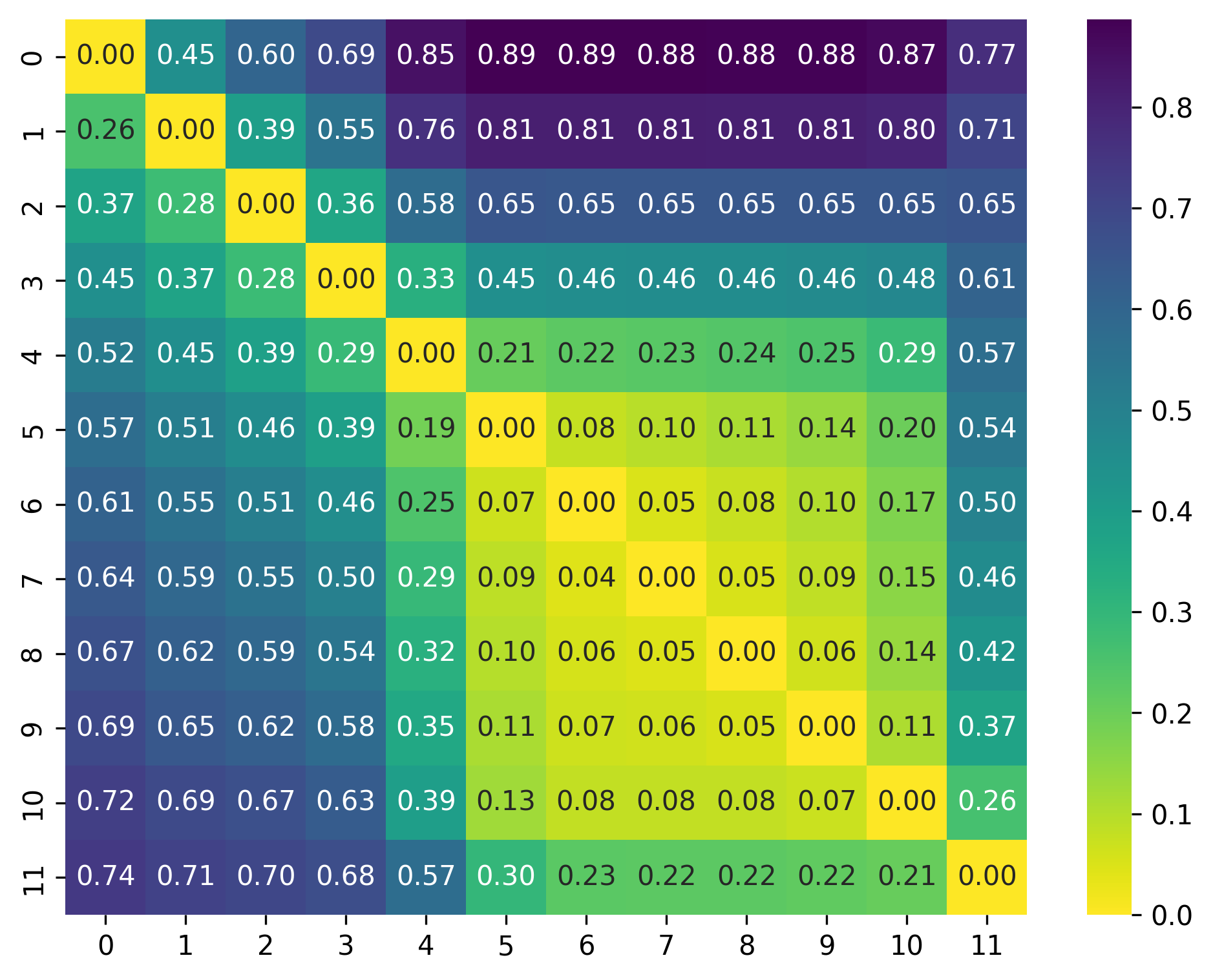}
    \end{minipage}
        \caption{\textbf{Block Similarities.} Block-by-block similarities in \vitt{}, \vits{}, \dinos{}, and \vitb{} models across five datasets: \mnist{}, \fmnist{}, \cifart{}, \cifarh{}, and \imagenet{}. Each matrix quantifies the linear error between latent representations of different blocks, showing potential blocks for approximation. The matrices reveal that the similarity between blocks is predominantly influenced by the model rather than the specific dataset.}
    \label{fig:app-latent-analysis-cka-cls}
\end{figure}

\clearpage
\subsubsection{Layer-wise Approximation Sensitivity} \label{sec:app-layer-sensitivity}

\begin{figure}[h]
    \centering
    \begin{overpic}[width=.8\textwidth]{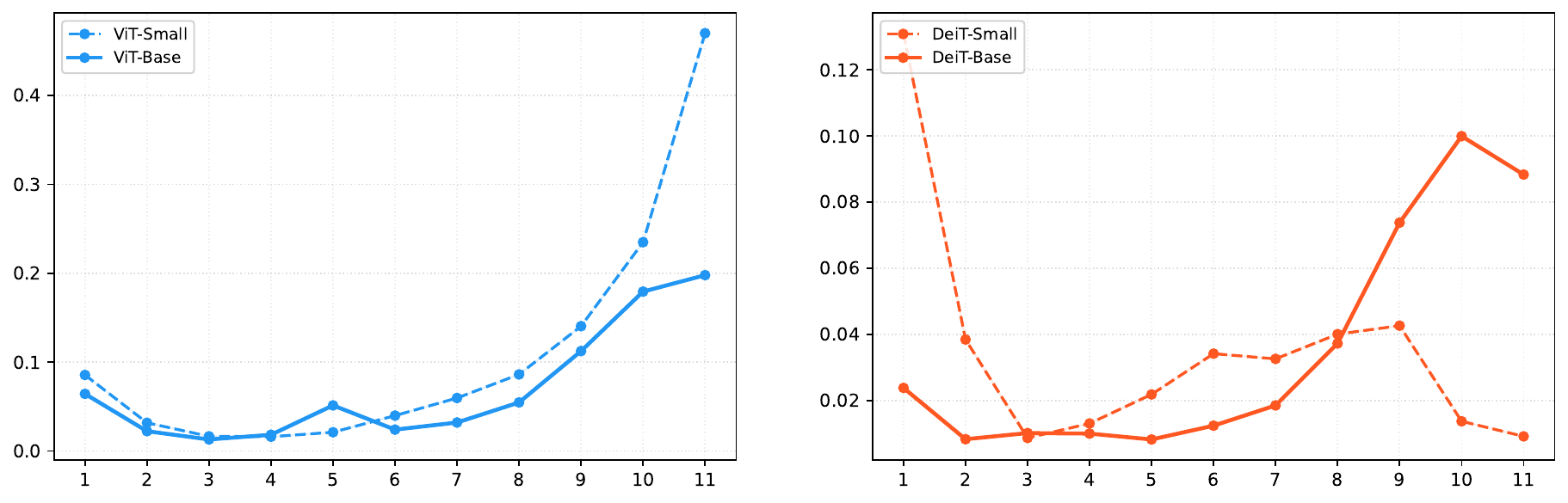}
        \put(-3,7){\rotatebox{90}{Accuracy drop ($\uparrow$)}}
    \end{overpic}
    \caption{\textbf{Layer-wise Skip Sensitivity (inference-only).}
        Accuracy drop ($\uparrow$ = worse) when replacing each individual block with a linear mapping, using the original pretrained classification head (no finetuning), for \vits{}, \vitb{}, \deits{}, and \deitb{} on \imagenet{}.
        Solid lines = base models; dashed lines = small models.
        \dino{} is excluded as it has no pretrained \imagenet{} classifier head.}
    \label{fig:layer-skip-delta-inference}
\end{figure}

This section extends the layer-wise skip analysis summarized in \Cref{sec:latent-analysis}. 
For each model we replace a single transformer block with a learned linear mapping (for each tuple $(j{-}1, j)$) and measure the accuracy drop relative to the unmodified baseline.
Linear translators are fitted on 500 samples and results are averaged over 3 seeds.
We evaluate three protocols: (1)~linear probe on \cifarhf{}, (2)~linear probe on \imagenet{}, and (3)~inference-only with the original pretrained classification head on \imagenet{} (applicable to ViT and \deit{} only, as \dino{} has no pretrained \imagenet{} head).
\Cref{fig:layer-skip-delta} (main paper) shows protocols~(1) and~(2); \Cref{fig:layer-skip-delta-inference} shows protocol~(3).
Full numerical results are reported in \Cref{tab:layer-skip-lp,tab:layer-skip-inference}.

\textbf{\vits{}.}
On \cifarhf{}, the profile is U-shaped: both layer~1 ($+5.11\%$) and layer~11 ($+6.87\%$) are hard to approximate, with the minimum at layer~5 ($+1.79\%$).
On \imagenet{} (linear probe), layer~1 is hard ($+8.26\%$), layers~3--5 are easiest ($+1.83$ to $+2.30\%$), then losses grow monotonically from layer~6 onward, reaching $+14.21\%$ at layer~11.
With a frozen head the same shape is dramatically steeper in the second half (layers~9--11: $+14.04\%$, $+23.49\%$, $+47.00\%$), showing that the final blocks produce highly task-specific representations the pretrained head critically depends on.

\textbf{\vitb{}.}
Layer~1 is hardest on \cifarhf{} ($+6.81\%$); the profile drops quickly, with a local spike at layer~5 ($+3.69\%$) and a minimum at layer~10 ($+0.54\%$).
On \imagenet{}, layer~1 is again hard ($+6.68\%$), easiest at layer~3 ($+1.46\%$), with losses broadly increasing from layer~8 onward ($+4.54\%$ $\to$ $+9.17\%$).
Frozen-head results follow the same ordering with steeper late-layer increases (layers~9--11: $+11.24\%$, $+17.93\%$, $+19.78\%$).

\textbf{\deits{}.}
The clearest pattern across all models.
Layer~1 is by far the hardest in every protocol (\cifarhf{}: $+9.11\%$; \imagenet{} linear: $+11.47\%$; frozen head: $+13.11\%$).
From layer~3 onward losses decrease sharply; on \cifarhf{}, layers~10--11 are beneficial ($-0.55\%$, $-0.04\%$), confirming that the final blocks of \deits{} are genuinely redundant.
On \imagenet{}, late layers remain among the easiest ($+1.36$ to $+1.54\%$ linear probe; $+0.91$ to $+1.37\%$ frozen head) though no longer beneficial.

\textbf{\deitb{}.}
Unusually, the hardest layers on \cifarhf{} are in the middle (peak at layer~4: $+2.55\%$), while early and late layers are easy (drops $\leq 1.02\%$).
The maximum drop is the smallest of all models ($2.55\%$).
On \imagenet{}, the pattern flips: layers~8--11 become the hardest region ($+3.69$ to $+6.45\%$ linear probe; $+7.38$ to $+9.99\%$ frozen head).
This dataset-dependent reversal is unique to \deitb{} and likely reflects that its later blocks encode \imagenet{}-specific representations not captured by \cifarhf{} finetuning.

\textbf{\dinos{}.}
On \cifarhf{}, layer~1 is easiest ($+1.87\%$) and sensitivity generally rises toward late layers, peaking at layer~11 ($+9.10\%$).
On \imagenet{}, the minimum shifts to layers~3--4 ($+1.86\%$), and the late-layer jump is steeper (layer~11: $+17.02\%$).

\textbf{\dinob{}.}
Layers~1--7 are mostly flat and easy on \cifarhf{} ($+1.05$ to $+2.52\%$), except for an isolated spike at layer~8 ($+9.00\%$).
On \imagenet{}, layer~8 remains catastrophic ($+17.62\%$) and, unlike \cifarhf{}, layers~9--11 do not recover ($+4.60\%$, $+7.56\%$, $+10.49\%$), suggesting the disruption propagates into downstream representations at \imagenet{} scale.

To summarize the key findings, layer-wise sensitivity profiles appear to be an intrinsic property of the architectures themselves: the easiest and hardest layers to linearize remain identical across all three evaluation protocols, with only the magnitude of the accuracy drops changing (drops are typically larger with a frozen head, as linear finetuning can partially compensate). However, block replaceability is highly architecture-dependent. For instance, \deits{} exhibits genuinely redundant late layers—approximating layers 10 and 11 actually improves accuracy on \cifarhf{} and incurs minimal penalty on \imagenet{}—whereas \dino{} models are generally easiest to skip in early-to-mid layers but highly sensitive at the end. \deitb{} stands out as the only model with a dataset-dependent reversal, being most sensitive in its middle layers on \cifarhf{} but in its late layers on \imagenet{}. We also note a striking, isolated anomaly in \dinob{}, where layer~8 incurs a massive cost (+9.00\% on \cifarhf{}, +17.62\% on \imagenet{}) despite all other early layers costing $\leq$2.52\%. Finally, as a general rule, Base models are considerably more robust to layer replacements than Small models; excluding the \dinob{} layer-8 outlier, the worst-case accuracy drop for Base models is just 6.45\% (\deitb{}, \imagenet{} linear), compared to much steeper maximum drops like 17.02\% (\dinos{}, \imagenet{}) and 47.00\% (\vits{}, frozen head). These findings align with the analysis conducted in \Cref{sec:latent-analysis}.

\begin{table}[h]
    \centering
    \small
    \caption{\textbf{Layer-wise skip sensitivity (linear probe).}
    Accuracy drop (\%) when replacing each individual block with a linear mapping, evaluated with a retrained linear classifier on \cifarhf{} and \imagenet{}.
    Results are averaged over 3 seeds.}
    \label{tab:layer-skip-lp}
    \begin{tabular}{lcccccc}
        \toprule
        & \multicolumn{3}{c}{\cifarhf{}} & \multicolumn{3}{c}{\imagenet{}} \\
        \cmidrule(lr){2-4} \cmidrule(lr){5-7}
        Model & Baseline & Easiest (layer) & Hardest (layer) & Baseline & Easiest (layer) & Hardest (layer) \\
        \midrule
        \vits{}  & 81.50 & +1.79 (5)  & +6.87 (11)  & 73.24 & +1.83 (4)  & +14.21 (11) \\
        \vitb{}  & 83.24 & +0.54 (10) & +6.81 (1)   & 76.99 & +1.46 (3)  & +9.17 (11)  \\
        \deits{} & 71.13 & $-$0.55 (10) & +9.11 (1) & 73.93 & +0.66 (3)  & +11.47 (1)  \\
        \deitb{} & 75.54 & +0.36 (10) & +2.55 (4)   & 77.70 & +0.54 (5)  & +6.45 (10)  \\
        \dinos{} & 82.04 & +1.87 (1)  & +9.10 (11)  & 67.45 & +1.86 (4)  & +17.02 (11) \\
        \dinob{} & 87.27 & +1.05 (1)  & +9.00 (8)   & 74.25 & +0.73 (4)  & +17.62 (8)  \\
        \bottomrule
    \end{tabular}
\end{table}

\begin{table}[h]
    \centering
    \small
    \caption{\textbf{Layer-wise skip sensitivity (inference-only).}
    Accuracy drop (\%) when replacing each individual block with a linear mapping, evaluated with the original pretrained classification head on \imagenet{} (no finetuning).
    \dino{} models are excluded as they have no pretrained \imagenet{} classifier head.}
    \label{tab:layer-skip-inference}
    \begin{tabular}{lcccc}
        \toprule
        Model & Baseline (\%) & Easiest (layer) & Hardest (layer) \\
        \midrule
        \vits{}  & 79.86 & +1.61 (4)  & +47.00 (11) \\
        \vitb{}  & 80.31 & +1.29 (3)  & +19.78 (11) \\
        \deits{} & 77.13 & +0.86 (3)  & +13.11 (1)  \\
        \deitb{} & 80.20 & +0.82 (5)  & +9.99 (10)  \\
        \bottomrule
    \end{tabular}
\end{table}

\subsection{Image Classification Performance} \label{sec:app-image-classification}

This section presents additional experiments that complement and extend those detailed in \Cref{sec:classification}. Datasets and models are the ones detailed in \Cref{tab:dataset-info,tab:pretrained-info}.


\begin{table}[h]
    \caption{\textbf{\vits{} Image Classification Performance Across Seeds.} Classification accuracy scores for \vits{} using multiple datasets, and 3 seeds. The ``Approx.'' column specifies the blocks used for approximation, where the first value represents the block whose output is used to approximate the second block's output, while the ``Params.'' column shows the number of parameters removed by the approximation compared to the original model.}
    \label{table:app-vit-s-classification}
    \centering
    \resizebox{.95\textwidth}{!}{%
    \begin{tabular}{cccccccc}
        \toprule
        Approx. & Params. & \mnist{} & \fmnist{} & \cifart{} & \cifarhc{} & \cifarhf{} & \imagenet{} \\
        \midrule
        1 $\rightarrow$ 5                  & \texttt{-6.75M} & $92.28 \pm 0.81$ & $86.90 \pm 0.72$ & $85.07 \pm 0.55$ & $68.01 \pm 0.31$ & $59.21 \pm 0.12$ & $44.04 \pm 0.42$ \\
        \cmidrule(l){1-8}
        2 $\rightarrow$ 5                  & \texttt{-5.12M} & $94.76 \pm 0.20$ & $88.57 \pm 0.31$ & $91.01 \pm 0.37$ & $77.77 \pm 0.22$ & $69.75 \pm 0.36$ & $60.38 \pm 0.12$ \\
        7 $\rightarrow$ 10                 & \texttt{-5.12M} & $94.58 \pm 0.28$ & $88.44 \pm 0.35$ & $87.36 \pm 0.17$ & $72.58 \pm 0.69$ & $62.03 \pm 0.56$ & $35.80 \pm 0.11$ \\
        \cmidrule(l){1-8}
        1 $\rightarrow$ 3                  & \texttt{-3.50M} & $94.60 \pm 0.78$ & $88.36 \pm 0.44$ & $91.97 \pm 0.16$ & $79.36 \pm 0.54$ & $72.41 \pm 0.08$ & $64.99 \pm 0.29$ \\
        2 $\rightarrow$ 4                  & \texttt{-3.50M} & $95.08 \pm 0.18$ & $88.83 \pm 0.21$ & $92.86 \pm 0.11$ & $81.45 \pm 0.44$ & $74.43 \pm 0.27$ & $67.52 \pm 0.16$ \\
        3 $\rightarrow$ 5                  & \texttt{-3.50M} & $94.75 \pm 0.57$ & $88.81 \pm 0.19$ & $94.09 \pm 0.06$ & $83.16 \pm 0.34$ & $76.17 \pm 0.45$ & $67.27 \pm 0.45$ \\
        1 $\rightarrow$ 2, 3 $\rightarrow$ 4 & \texttt{-3.50M} & $94.68 \pm 0.69$ & $88.30 \pm 0.25$ & $91.91 \pm 0.25$ & $79.72 \pm 0.16$ & $72.17 \pm 0.15$ & $65.38 \pm 0.03$ \\
        1 $\rightarrow$ 2, 4 $\rightarrow$ 5 & \texttt{-3.50M} & $94.58 \pm 0.77$ & $88.95 \pm 0.07$ & $92.29 \pm 0.28$ & $80.14 \pm 0.10$ & $72.45 \pm 0.35$ & $64.42 \pm 0.24$ \\
        \cmidrule(l){1-8}
        0 $\rightarrow$ 1                  & \texttt{-1.63M} & $\mathbf{95.69} \pm 0.29$ & $88.81 \pm 0.19$ & $93.68 \pm 0.22$ & $83.55 \pm 0.23$ & $76.49 \pm 0.29$ & $65.11 \pm 0.27$ \\
        1 $\rightarrow$ 2                  & \texttt{-1.63M} & $95.40 \pm 0.57$ & $88.53 \pm 0.63$ & $93.90 \pm 0.11$ & $83.98 \pm 0.22$ & $76.99 \pm 0.26$ & $70.32 \pm 0.38$ \\
        2 $\rightarrow$ 3                  & \texttt{-1.63M} & $95.43 \pm 0.45$ & $88.93 \pm 0.62$ & $94.90 \pm 0.26$ & $85.72 \pm 0.48$ & $78.96 \pm 0.05$ & $71.26 \pm 0.03$ \\
        3 $\rightarrow$ 4                  & \texttt{-1.63M} & $95.43 \pm 0.39$ & $88.77 \pm 0.36$ & $95.05 \pm 0.17$ & $85.99 \pm 0.37$ & $79.49 \pm 0.32$ & $\mathbf{71.40} \pm 0.22$ \\
        4 $\rightarrow$ 5                  & \texttt{-1.63M} & $95.39 \pm 0.35$ & $89.18 \pm 0.51$ & $\mathbf{95.41} \pm 0.12$ & $86.27 \pm 0.27$ & $\mathbf{79.61} \pm 0.14$ & $70.98 \pm 0.16$ \\
        5 $\rightarrow$ 6                  & \texttt{-1.63M} & $95.14 \pm 0.56$ & $89.30 \pm 0.54$ & $94.89 \pm 0.27$ & $\mathbf{86.49} \pm 0.33$ & $79.29 \pm 0.19$ & $69.25 \pm 0.09$ \\
        6 $\rightarrow$ 7                  & \texttt{-1.63M} & $95.11 \pm 0.42$ & $88.94 \pm 0.66$ & $94.81 \pm 0.26$ & $85.33 \pm 0.30$ & $78.06 \pm 0.17$ & $67.41 \pm 0.08$ \\
        7 $\rightarrow$ 8                  & \texttt{-1.63M} & $95.64 \pm 0.46$ & $89.41 \pm 0.45$ & $94.50 \pm 0.34$ & $85.30 \pm 0.50$ & $78.03 \pm 0.12$ & $66.22 \pm 0.10$ \\
        8 $\rightarrow$ 9                  & \texttt{-1.63M} & $95.36 \pm 0.47$ & $\mathbf{89.64} \pm 0.37$ & $94.36 \pm 0.14$ & $84.66 \pm 0.25$ & $77.88 \pm 0.20$ & $64.03 \pm 0.29$ \\
        9 $\rightarrow$ 10                 & \texttt{-1.63M} & $95.52 \pm 0.41$ & $89.57 \pm 0.10$ & $94.58 \pm 0.27$ & $81.76 \pm 0.34$ & $76.45 \pm 0.22$ & $61.82 \pm 0.24$ \\
        10 $\rightarrow$ 11                & \texttt{-1.63M} & $94.83 \pm 0.20$ & $89.11 \pm 0.43$ & $94.08 \pm 0.27$ & $82.13 \pm 0.70$ & $77.45 \pm 0.29$ & $63.92 \pm 0.25$ \\
        \cmidrule(l){1-8}
        original                          & \texttt{22.06M} & $\underline{95.59} \pm 0.42$ & $\underline{89.04} \pm 0.85$ & $\underline{95.68} \pm 0.24$ & $\underline{87.61} \pm 0.39$ & $\underline{81.50} \pm 0.39$ & $\underline{73.24} \pm 0.13$ \\
        \bottomrule
    \end{tabular}
    }
\end{table}


\begin{table}[h]
    \caption{\textbf{\dinos{} Image Classification Performance Across Seeds.} Classification accuracy scores for \dinos{} using multiple datasets, and 3 seeds. The ``Approx.'' column specifies the blocks used for approximation, where the first value represents the block whose output is used to approximate the second block's output, while the ``Params.'' column shows the number of parameters removed by the approximation compared to the original model.}
    \label{table:app-dino-s-classification}
    \centering
    \resizebox{.95\textwidth}{!}{%
    \begin{tabular}{cccccccc}
        \toprule
        Approx. & Params. & \mnist{} & \fmnist{} & \cifart{} & \cifarhc{} & \cifarhf{} & \imagenet{} \\
        \midrule
        1 $\rightarrow$ 5                  & \texttt{-6.75M} & $96.25 \pm 0.30$ & $86.50 \pm 1.42$ & $80.11 \pm 0.95$ & $59.15 \pm 0.45$ & $51.24 \pm 0.51$ & $18.70 \pm 0.09$ \\
        \cmidrule(l){1-8}
        2 $\rightarrow$ 5                  & \texttt{-5.12M} & $95.86 \pm 0.52$ & $87.99 \pm 0.30$ & $85.28 \pm 0.99$ & $67.50 \pm 1.02$ & $59.57 \pm 0.45$ & $40.63 \pm 0.59$ \\
        7 $\rightarrow$ 10                 & \texttt{-5.12M} & $96.05 \pm 1.44$ & $88.28 \pm 1.25$ & $91.00 \pm 0.82$ & $78.47 \pm 0.61$ & $70.56 \pm 0.25$ & $45.66 \pm 0.69$ \\
        \cmidrule(l){1-8}
        1 $\rightarrow$ 3                  & \texttt{-3.50M} & $96.61 \pm 0.34$ & $88.48 \pm 0.61$ & $91.73 \pm 0.36$ & $78.62 \pm 0.87$ & $72.33 \pm 0.37$ & $56.85 \pm 0.21$ \\
        2 $\rightarrow$ 4                  & \texttt{-3.50M} & $96.79 \pm 0.58$ & $88.34 \pm 0.33$ & $91.31 \pm 0.16$ & $76.41 \pm 0.44$ & $69.71 \pm 0.31$ & $60.16 \pm 0.41$ \\
        3 $\rightarrow$ 5                  & \texttt{-3.50M} & $96.76 \pm 1.02$ & $88.65 \pm 0.92$ & $91.00 \pm 0.49$ & $75.51 \pm 0.45$ & $69.31 \pm 0.05$ & $57.47 \pm 0.11$ \\
        1 $\rightarrow$ 2, 3 $\rightarrow$ 4 & \texttt{-3.50M} & $96.71 \pm 0.62$ & $88.69 \pm 0.46$ & $92.57 \pm 0.54$ & $79.16 \pm 1.02$ & $72.88 \pm 0.57$ & $59.79 \pm 0.19$ \\
        1 $\rightarrow$ 2, 4 $\rightarrow$ 5 & \texttt{-3.50M} & $96.81 \pm 0.31$ & $88.67 \pm 1.23$ & $93.50 \pm 0.26$ & $79.35 \pm 1.00$ & $73.55 \pm 0.38$ & $58.62 \pm 0.25$ \\
        \cmidrule(l){1-8}
        0 $\rightarrow$ 1                  & \texttt{-1.63M} & $96.71 \pm 0.79$ & $88.97 \pm 1.12$ & $\mathbf{95.67} \pm 0.12$ & $\mathbf{85.89} \pm 0.56$ & $\mathbf{80.15} \pm 0.35$ & $61.25 \pm 0.22$ \\
        1 $\rightarrow$ 2                  & \texttt{-1.63M} & $96.69 \pm 0.90$ & $88.26 \pm 1.10$ & $95.38 \pm 0.09$ & $84.86 \pm 0.84$ & $79.38 \pm 0.23$ & $64.86 \pm 0.36$ \\
        2 $\rightarrow$ 3                  & \texttt{-1.63M} & $96.42 \pm 0.36$ & $88.31 \pm 1.20$ & $94.71 \pm 0.33$ & $84.15 \pm 0.94$ & $77.74 \pm 0.85$ & $65.16 \pm 0.69$ \\
        3 $\rightarrow$ 4                  & \texttt{-1.63M} & $96.82 \pm 0.68$ & $88.77 \pm 0.78$ & $94.87 \pm 0.30$ & $83.96 \pm 0.62$ & $77.71 \pm 0.08$ & $\mathbf{65.35} \pm 0.56$ \\
        4 $\rightarrow$ 5                  & \texttt{-1.63M} & $96.82 \pm 0.60$ & $89.15 \pm 0.72$ & $94.63 \pm 0.26$ & $83.04 \pm 0.62$ & $77.13 \pm 0.17$ & $64.28 \pm 0.24$ \\
        5 $\rightarrow$ 6                  & \texttt{-1.63M} & $96.81 \pm 0.85$ & $88.75 \pm 0.86$ & $95.33 \pm 0.19$ & $84.83 \pm 0.04$ & $79.37 \pm 0.25$ & $64.88 \pm 0.43$ \\
        6 $\rightarrow$ 7                  & \texttt{-1.63M} & $\mathbf{96.99} \pm 0.88$ & $\mathbf{89.42} \pm 0.68$ & $95.21 \pm 0.10$ & $83.82 \pm 0.53$ & $78.54 \pm 0.64$ & $63.61 \pm 0.62$ \\
        7 $\rightarrow$ 8                  & \texttt{-1.63M} & $96.76 \pm 0.38$ & $89.05 \pm 1.29$ & $95.37 \pm 0.14$ & $84.57 \pm 0.42$ & $78.95 \pm 0.37$ & $61.59 \pm 0.31$ \\
        8 $\rightarrow$ 9                  & \texttt{-1.63M} & $96.62 \pm 0.85$ & $88.45 \pm 1.21$ & $95.21 \pm 0.36$ & $84.98 \pm 0.22$ & $79.35 \pm 0.22$ & $61.73 \pm 0.43$ \\
        9 $\rightarrow$ 10                 & \texttt{-1.63M} & $96.66 \pm 0.33$ & $88.53 \pm 0.71$ & $94.55 \pm 0.25$ & $83.97 \pm 1.25$ & $77.06 \pm 0.36$ & $58.56 \pm 0.25$ \\
        10 $\rightarrow$ 11                & \texttt{-1.63M} & $94.61 \pm 0.66$ & $86.96 \pm 1.18$ & $92.11 \pm 0.32$ & $79.85 \pm 0.26$ & $73.01 \pm 0.51$ & $50.76 \pm 0.33$ \\
        \cmidrule(l){1-8}
        original                          & \texttt{22.06M} & $\underline{96.57} \pm 0.64$ & $\underline{88.07} \pm 1.40$ & $\underline{96.24} \pm 0.08$ & $\underline{87.53} \pm 0.45$ & $\underline{82.04} \pm 0.42$ & $\underline{67.45} \pm 0.45$ \\
        \bottomrule
    \end{tabular}
    }
\end{table}


\begin{table}
    \caption{\textbf{\vitt{} Image Classification Performance.} Classification accuracy scores for \vitt{} using multiple datasets, and 3 seeds. The ``Approx.'' column specifies the blocks used for approximation, where the first value represents the block whose output is used to approximate the second block's output, while the ``Params.'' column shows the number of parameters removed by the approximation compared to the original model.\looseness=-1}
    \label{table:app-vit-t-classification}
    \centering
    \resizebox{.95\textwidth}{!}{%
    \begin{tabular}{cccccccc}
        \toprule
        Approx. & Params. & \mnist{} & \fmnist{} & \cifart{} & \cifarhc{} & \cifarhf{} & \imagenet{} \\
        \midrule
        1 $\rightarrow$ 5                  & \texttt{-1.78M} & $87.66 \pm 0.57$ & $85.10 \pm 0.42$ & $73.68 \pm 0.46$ & $53.46 \pm 0.29$ & $44.61 \pm 0.42$ & $22.21 \pm 0.39$ \\
        \cmidrule(l){1-8}
        2 $\rightarrow$ 5                  & \texttt{-1.34M} & $90.59 \pm 0.79$ & $85.84 \pm 0.18$ & $82.41 \pm 0.11$ & $62.87 \pm 0.21$ & $54.68 \pm 0.21$ & $35.14 \pm 0.38$ \\
        7 $\rightarrow$ 10                 & \texttt{-1.34M} & $92.41 \pm 0.47$ & $86.50 \pm 0.19$ & $82.48 \pm 0.85$ & $69.26 \pm 0.65$ & $61.15 \pm 0.28$ & $39.03 \pm 0.13$ \\
        \cmidrule(l){1-8}
        1 $\rightarrow$ 3                  & \texttt{-0.89M} & $90.55 \pm 1.04$ & $85.91 \pm 0.22$ & $80.48 \pm 0.29$ & $63.43 \pm 0.25$ & $54.57 \pm 0.32$ & $43.68 \pm 0.26$ \\
        2 $\rightarrow$ 4                  & \texttt{-0.89M} & $92.81 \pm 0.56$ & $86.58 \pm 0.05$ & $86.85 \pm 0.17$ & $70.49 \pm 0.30$ & $63.53 \pm 0.23$ & $49.94 \pm 0.27$ \\
        3 $\rightarrow$ 5                  & \texttt{-0.89M} & $91.84 \pm 0.69$ & $86.80 \pm 0.04$ & $88.00 \pm 0.04$ & $72.67 \pm 0.30$ & $65.66 \pm 0.14$ & $48.48 \pm 0.37$ \\
        1 $\rightarrow$ 2, 3 $\rightarrow$ 4 & \texttt{-0.89M} & $91.94 \pm 0.78$ & $86.71 \pm 0.20$ & $83.43 \pm 0.41$ & $66.92 \pm 0.42$ & $60.07 \pm 0.48$ & $45.14 \pm 0.15$ \\
        1 $\rightarrow$ 2, 4 $\rightarrow$ 5 & \texttt{-0.89M} & $90.86 \pm 0.66$ & $86.57 \pm 0.24$ & $84.61 \pm 0.14$ & $68.07 \pm 0.55$ & $60.11 \pm 0.61$ & $44.84 \pm 0.26$ \\
        \cmidrule(l){1-8}
        0 $\rightarrow$ 1                  & \texttt{-0.45M} & $91.74 \pm 0.48$ & $86.22 \pm 0.23$ & $83.32 \pm 0.22$ & $68.58 \pm 0.41$ & $61.05 \pm 0.36$ & $44.12 \pm 0.20$ \\
        1 $\rightarrow$ 2                  & \texttt{-0.45M} & $91.65 \pm 0.61$ & $86.26 \pm 0.24$ & $85.84 \pm 0.08$ & $71.12 \pm 0.06$ & $63.85 \pm 0.37$ & $54.34 \pm 0.44$ \\
        2 $\rightarrow$ 3                  & \texttt{-0.45M} & $92.89 \pm 0.18$ & $86.49 \pm 0.06$ & $88.89 \pm 0.08$ & $74.90 \pm 0.25$ & $68.03 \pm 0.37$ & $57.83 \pm 0.07$ \\
        3 $\rightarrow$ 4                  & \texttt{-0.45M} & $93.10 \pm 0.43$ & $\mathbf{87.34} \pm 0.03$ & $89.73 \pm 0.37$ & $76.45 \pm 0.17$ & $70.04 \pm 0.35$ & $57.55 \pm 0.14$ \\
        4 $\rightarrow$ 5                  & \texttt{-0.45M} & $92.43 \pm 0.20$ & $87.22 \pm 0.10$ & $90.11 \pm 0.32$ & $76.40 \pm 0.42$ & $69.97 \pm 0.37$ & $55.91 \pm 0.10$ \\
        5 $\rightarrow$ 6                  & \texttt{-0.45M} & $\mathbf{93.57} \pm 0.11$ & $86.80 \pm 0.13$ & $90.17 \pm 0.27$ & $76.47 \pm 0.35$ & $70.69 \pm 0.49$ & $55.43 \pm 0.38$ \\
        6 $\rightarrow$ 7                  & \texttt{-0.45M} & $92.13 \pm 0.37$ & $86.77 \pm 0.02$ & $87.73 \pm 0.22$ & $72.35 \pm 0.31$ & $66.73 \pm 0.45$ & $47.39 \pm 0.45$ \\
        7 $\rightarrow$ 8                  & \texttt{-0.45M} & $93.20 \pm 0.06$ & $86.90 \pm 0.30$ & $88.58 \pm 0.26$ & $75.80 \pm 0.29$ & $69.28 \pm 0.41$ & $53.48 \pm 0.24$ \\
        8 $\rightarrow$ 9                  & \texttt{-0.45M} & $92.76 \pm 0.11$ & $87.18 \pm 0.17$ & $89.57 \pm 0.33$ & $76.43 \pm 0.50$ & $71.07 \pm 0.33$ & $56.07 \pm 0.77$ \\
        9 $\rightarrow$ 10                 & \texttt{-0.45M} & $92.39 \pm 0.10$ & $86.74 \pm 0.18$ & $89.86 \pm 0.31$ & $77.34 \pm 0.04$ & $71.70 \pm 0.37$ & $57.45 \pm 0.29$ \\
        10 $\rightarrow$ 11                & \texttt{-0.45M} & $90.92 \pm 0.48$ & $86.89 \pm 0.12$ & $\mathbf{90.98} \pm 0.21$ & $\mathbf{78.85} \pm 0.38$ & $\mathbf{72.29} \pm 0.42$ & $\mathbf{58.94} \pm 0.22$ \\
        \cmidrule(l){1-8}
        original                          & \texttt{5.72M} & $\underline{93.22} \pm 0.18$ & $\underline{86.99} \pm 0.29$ & $\underline{91.29} \pm 0.06$ & $\underline{79.27} \pm 0.23$ & $\underline{73.45} \pm 0.38$ & $\underline{63.02} \pm 0.22$ \\
        \bottomrule
    \end{tabular}
    }
\end{table}

\begin{table}[h]
        \caption{\textbf{\vitb{} Image Classification Performance.} Classification accuracy scores for \vitb{} using multiple datasets, and 3 seeds. The ``Approx.'' column specifies the blocks used for approximation, where the first value represents the block whose output is used to approximate the second block's output, while the ``Params.'' column shows the number of parameters removed by the approximation compared to the original model.\looseness=-1}
        \label{table:app-vitb-classification-results}
        \centering
        \resizebox{.8\textwidth}{!}{%
                \begin{tabular}{ccccccc}
                        \toprule
                                           &                 & \multicolumn{5}{c}{Accuracy $\uparrow$ }                                                                                                                             \\
                        \cmidrule(l){3-7}
                        Approx.            &  Params.           & \mnist{}                                 & \fmnist{}                    & \cifart{}                    & \cifarhc{}                   & \cifarhf{}                   \\
                        \midrule
                        1 $\rightarrow$ 5  & \texttt{-25.99M} & $87.06 \pm 0.53$                         & $84.33 \pm 0.61$             & $73.54 \pm 0.57$             & $51.67 \pm 1.10$             & $38.98 \pm 0.72$             \\
                        \cmidrule(l){1-7}
                        2 $\rightarrow$ 5  & \texttt{-19.49M} & $94.20 \pm 0.21$                         & $87.80 \pm 0.24$             & $87.10 \pm 0.83$             & $71.68 \pm 0.50$             & $61.19 \pm 0.37$             \\
                        \cmidrule(l){1-7}
                        1 $\rightarrow$ 3  & \texttt{-13M} & $96.51 \pm 0.42$                         & $88.72 \pm 0.41$             & $93.71 \pm 0.13$             & $83.05 \pm 0.23$             & $74.74 \pm 0.29$             \\
                        3 $\rightarrow$ 5  & \texttt{-13M} & $95.59 \pm 0.09$                         & $88.28 \pm 0.20$             & $93.11 \pm 0.06$             & $83.50 \pm 0.17$             & $74.35 \pm 0.47$             \\
                        2 $\rightarrow$ 4  & \texttt{-13M} & $96.21 \pm 0.33$                         & $89.21 \pm 0.64$             & $94.59 \pm 0.32$             & $85.13 \pm 0.24$             & $76.82 \pm 0.41$             \\
                        8 $\rightarrow$ 10 & \texttt{-13M} & $96.54 \pm 0.21$                         & $\textbf{89.72} \pm 0.52$    & $95.05 \pm 0.26$             & $85.78 \pm 0.37$             & $79.62 \pm 0.14$             \\
                        9 $\rightarrow$ 11 & \texttt{-13M} & $95.59 \pm 0.52$                         & $89.49 \pm 0.26$             & $93.22 \pm 0.56$             & $82.23 \pm 0.44$             & $76.33 \pm 0.10$             \\
                        \cmidrule(l){1-7}
                        3 $\rightarrow$ 4  & \texttt{-6.5M} & $96.86 \pm 0.35$                         & $89.69 \pm 1.09$             & $\textbf{96.18} \pm 0.09$    & $\textbf{89.18} \pm 0.06$    & $\textbf{82.50} \pm 0.17$    \\
                        4 $\rightarrow$ 5  & \texttt{-6.5M} & $96.55 \pm 0.23$                         & $89.13 \pm 0.50$             & $95.39 \pm 0.23$             & $87.43 \pm 0.15$             & $80.30 \pm 0.16$             \\
                        0 $\rightarrow$ 1  & \texttt{-6.5M} & $96.75 \pm 0.29$                         & $88.97 \pm 0.26$             & $93.74 \pm 0.15$             & $84.49 \pm 0.20$             & $76.54 \pm 0.29$             \\
                        1 $\rightarrow$ 2  & \texttt{-6.5M} & $96.88 \pm 0.01$                         & $89.29 \pm 0.24$             & $95.63 \pm 0.11$             & $87.46 \pm 0.20$             & $80.64 \pm 0.23$             \\
                        2 $\rightarrow$ 3  & \texttt{-6.5M} & $\textbf{96.91} \pm 0.17$                & $89.69 \pm 0.61$             & $96.00 \pm 0.18$             & $88.38 \pm 0.13$             & $81.59 \pm 0.35$             \\
                        \cmidrule(l){1-7}
                        original           & \texttt{86.39M} & $\underline{95.61} \pm 0.22$             & $\underline{89.64} \pm 0.57$ & $\underline{96.25} \pm 0.17$ & $\underline{89.52} \pm 0.23$ & $\underline{83.41} \pm 0.20$ \\
                        \bottomrule
                \end{tabular}
        }
\end{table}

\clearpage

\subsubsection{E2E Finetuning} \label{sec:app-e2e-finetuning}

In this section we perform \gls{e2e} finetuning experiments on \cifarh{} for \deits{}, \dinob{}, and \vitl{}, demonstrating that \gls{toast} provides a strong initialization for finetuning: even under aggressive compression (e.g., removing 26M parameters from \dinob{}), \gls{e2e} finetuning recovers accuracy to within 3.56\% of the original.\looseness=-1

\begin{table}[h]
    \centering
    \caption{\textbf{E2E vs Head-only Finetuning Classification Performance on \cifarhf{}.} Image classification accuracy when end-to-end finetuning the network vs only finetuning the head for \deits{}, \dinob{}, and \vitl{} using \cifarhf{}. The ``Approx.'' column indicates the block pairs where the first block approximates the second.\looseness=-1}
    \label{table:classification-e2e-results-cifar100-complete}
    \setlength{\tabcolsep}{2pt}
    \resizebox{.8\textwidth}{!}{
        \begin{tabular}{cccccccc}
            \toprule
             &                                           &                                                         &                                           & \multicolumn{2}{c}{Head-only finetuning (Acc \%)} & \multicolumn{2}{c}{End-to-end finetuning (Acc \%)}                                                                                                         \\
            \cmidrule(lr){5-6}
            \cmidrule(lr){7-8}

             &                                           & Approx.                                                 & Params.                                   & Identity                                          & Linear                                             & Identity                                          & Linear                                            \\
            \midrule
             & \multirow{5}{*}{\rotatebox{90}{\deits{}}}
             & 3 $\rightarrow$ 4, 9 $\rightarrow$ 11     & \texttt{-4.88M}                                         & $68.48 \pm 0.34 \scriptstyle{(-3.44\%)}$  & $70.64 \pm 0.37 \scriptstyle{(-0.39\%)}$          & $86.03 \pm 0.03 \scriptstyle{(-1.05\%)}$           & $85.99 \pm 0.00 \scriptstyle{(-1.09\%)}$                                                              \\
             &                                           & 9 $\rightarrow$ 11                                      & \texttt{-3.25M}                           & $72.28 \pm 0.36 \mathbf{\scriptstyle{(+1.92\%)}}$ & $72.04 \pm 0.42 \mathbf{\scriptstyle{(+1.57\%)}}$  & $86.81 \pm 0.01 \scriptstyle{(-0.15\%)}$          & $86.86 \pm 0.01 \scriptstyle{(-0.09\%)}$          \\
             &                                           & 8 $\rightarrow$ 9                                       & \texttt{-1.62M}                           & $71.34 \pm 0.10 \scriptstyle{(+0.60\%)}$          & $70.80 \pm 0.12 \scriptstyle{(-0.17\%)}$           & $87.13 \pm 0.02 \mathbf{\scriptstyle{(+0.22\%)}}$ & $86.95 \pm 0.01 \mathbf{\scriptstyle{(+0.01\%)}}$ \\
             &                                           & 9 $\rightarrow$ 10                                      & \texttt{-1.62M}                           & $71.66 \pm 0.39 \scriptstyle{(+1.04\%)}$          & $71.49 \pm 0.20 \scriptstyle{(+0.80\%)}$           & $87.01 \pm 0.02 \scriptstyle{(+0.08\%)}$          & $86.85 \pm 0.01 \scriptstyle{(-0.10\%)}$          \\
            \cmidrule(l){3-8}
             &                                           & original                                                & \texttt{21.81M}                           & $70.92 \pm 0.18$                                  & $70.92 \pm 0.18$                                   & $86.94 \pm 0.03$                                  & $86.94 \pm 0.03$                                  \\
            \cmidrule(l){2-8}

             & \multirow{6}{*}{\rotatebox{90}{\dinob{}}}
             & 0 $\rightarrow$ 4                         & \texttt{-26M}                                        & $18.29 \pm 0.86 \scriptstyle{(-79.09\%)}$ & $62.25 \pm 0.54 \scriptstyle{(-28.83\%)}$         & $56.15 \pm 0.01 \scriptstyle{(-39.52\%)}$          & $89.54 \pm 0.00 \scriptstyle{(-3.56\%)}$                                                              \\
             &                                           & 0 $\rightarrow$ 1, 2 $\rightarrow$ 3, 4 $\rightarrow$ 5 & \texttt{-19.5M}                          & $29.05 \pm 0.31 \scriptstyle{(-66.79\%)}$         & $79.06 \pm 0.27 \scriptstyle{(-9.60\%)}$           & $70.06 \pm 0.00 \scriptstyle{(-24.54\%)}$         & $91.34 \pm 0.00 \scriptstyle{(-1.63\%)}$          \\
             &                                           & 0 $\rightarrow$ 1, 2 $\rightarrow$ 3                    & \texttt{-13M}                          & $33.25 \pm 0.18 \scriptstyle{(-61.99\%)}$         & $84.18 \pm 0.18 \scriptstyle{(-3.76\%)}$           & $78.69 \pm 0.22 \scriptstyle{(-15.25\%)}$         & $92.34 \pm 0.00 \scriptstyle{(-0.55\%)}$          \\
             &                                           & 0 $\rightarrow$ 1                                       & \texttt{-6.5M}                           & $78.83 \pm 0.22 \mathbf{\scriptstyle{(-9.87\%)}}$ & $86.64 \pm 0.37 \mathbf{\scriptstyle{(-0.94\%)}}$  & $92.39 \pm 0.00 \mathbf{\scriptstyle{(-0.50\%)}}$ & $92.78 \pm 0.00 \mathbf{\scriptstyle{(-0.08\%)}}$ \\
             &                                           & 2 $\rightarrow$ 3                                       & \texttt{-6.5M}                           & $47.51 \pm 0.52 \scriptstyle{(-45.68\%)}$         & $86.06 \pm 0.20 \scriptstyle{(-1.60\%)}$           & $91.56 \pm 0.00 \scriptstyle{(-1.39\%)}$          & $92.75 \pm 0.00 \scriptstyle{(-0.11\%)}$          \\
            \cmidrule(l){3-8}
             &                                           & original                                                & \texttt{86.58M}                           & $87.46 \pm 0.04$                                  & $87.46 \pm 0.04$                                   & $92.85 \pm 0.00$                                  & $92.85 \pm 0.00$                                  \\
            \cmidrule(l){2-8}

             & \multirow{11}{*}{\rotatebox{90}{\vitl{}}}
             & 2 $\rightarrow$ 4, 18 $\rightarrow$ 23    & \texttt{-80.83M}                                        & $74.41 \pm 0.44 \scriptstyle{(-13.79\%)}$ & $84.02 \pm 0.39 \scriptstyle{(-2.66\%)}$          & $89.19 \pm 0.01 \scriptstyle{(-1.94\%)}$           & $88.73 \pm 0.00 \scriptstyle{(-2.44\%)}$                                                              \\
             &                                           & 17 $\rightarrow$ 23                                     & \texttt{-69.28M}                          & $85.32 \pm 0.45 \scriptstyle{(-1.16\%)}$          & $84.55 \pm 0.44 \scriptstyle{(-2.05\%)}$           & $90.73 \pm 0.00 \scriptstyle{(-0.24\%)}$          & $90.36 \pm 0.00 \scriptstyle{(-0.65\%)}$          \\
             &                                           & 3 $\rightarrow$ 4, 19 $\rightarrow$ 23                  & \texttt{-57.74M}                          & $84.23 \pm 0.08 \scriptstyle{(-2.43\%)}$          & $85.81 \pm 0.39 \scriptstyle{(-0.59\%)}$           & $90.34 \pm 0.00 \scriptstyle{(-0.67\%)}$          & $90.12 \pm 0.00 \scriptstyle{(-0.91\%)}$          \\
             &                                           & 3 $\rightarrow$ 4, 20 $\rightarrow$ 23                  & \texttt{-46.19M}                          & $84.68 \pm 0.18 \scriptstyle{(-1.90\%)}$          & $86.30 \pm 0.11 \scriptstyle{(-0.03\%)}$           & $90.44 \pm 0.00 \scriptstyle{(-0.56\%)}$          & $90.03 \pm 0.00 \scriptstyle{(-1.01\%)}$          \\
             &                                           & 20 $\rightarrow$ 23                                     & \texttt{-34.64M}                          & $86.61 \pm 0.07 \mathbf{\scriptstyle{(+0.33\%)}}$ & $86.55 \pm 0.22 \scriptstyle{(+0.27\%)}$           & $91.04 \pm 0.00 \mathbf{\scriptstyle{(+0.10\%)}}$ & $90.74 \pm 0.00 \scriptstyle{(-0.23\%)}$          \\
             &                                           & 3 $\rightarrow$ 4, 21 $\rightarrow$ 23                  & \texttt{-34.64M}                          & $84.86 \pm 0.28 \scriptstyle{(-1.70\%)}$          & $86.37 \pm 0.28 \scriptstyle{(+0.06\%)}$           & $90.54 \pm 0.00 \scriptstyle{(-0.45\%)}$          & $89.40 \pm 0.01 \scriptstyle{(-1.70\%)}$          \\
             &                                           & 20 $\rightarrow$ 22                                     & \texttt{-23.09M}                          & $86.30 \pm 0.23 \scriptstyle{(-0.03\%)}$          & $86.52 \pm 0.12 \scriptstyle{(+0.24\%)}$           & $90.94 \pm 0.00 \scriptstyle{(-0.01\%)}$          & $90.66 \pm 0.00 \scriptstyle{(-0.32\%)}$          \\
             &                                           & 3 $\rightarrow$ 4, 21 $\rightarrow$ 22                  & \texttt{-23.09M}                          & $84.58 \pm 0.19 \scriptstyle{(-2.02\%)}$          & $86.20 \pm 0.11 \scriptstyle{(-0.14\%)}$           & $90.65 \pm 0.00 \scriptstyle{(-0.33\%)}$          & $86.81 \pm 0.05 \scriptstyle{(-4.56\%)}$          \\
             &                                           & 20 $\rightarrow$ 21                                     & \texttt{-11.55M}                          & $86.44 \pm 0.24 \scriptstyle{(+0.14\%)}$          & $86.39 \pm 0.08 \scriptstyle{(+0.08\%)}$           & $90.91 \pm 0.00 \scriptstyle{(-0.04\%)}$          & $90.88 \pm 0.00 \scriptstyle{(-0.07\%)}$          \\
             &                                           & 21 $\rightarrow$ 22                                     & \texttt{-11.55M}                          & $86.55 \pm 0.01 \scriptstyle{(+0.26\%)}$          & $86.72 \pm 0.24 \mathbf{\scriptstyle{(+0.46\%)}}$  & $90.98 \pm 0.00 \scriptstyle{(+0.04\%)}$          & $90.86 \pm 0.00 \scriptstyle{(-0.10\%)}$          \\
            \cmidrule(l){3-8}
             &                                           & original                                                & \texttt{304.35M}                          & $86.32 \pm 0.08$                                  & $86.32 \pm 0.08$                                   & $90.95 \pm 0.00$                                  & $90.95 \pm 0.00$                                  \\

            \bottomrule
        \end{tabular}
    }
\end{table}

\subsubsection{Zero-shot Image Classification} \label{sec:app-image-zeroshot}

\begin{wraptable}{l}{0.4\textwidth}
\centering
\vspace{-1.5em}
\caption{\textbf{Zero-shot image classification.} Accuracy scores for \clipb{} on \imagenet{}. The ``Approx.'' column indicates the block pairs where the first block approximates the second. The ``$\Delta$'' column indicates the change in accuracy.}
    \label{table:clip-zeroshot}
    \centering
    \resizebox{.35\textwidth}{!}{
        \begin{tabular}{@{}c c c c@{}}
        \toprule
        Params.               & Approx.             & Accuracy $\uparrow$ & $\Delta$            \\
        \midrule
        \multirow{11}{*}{\texttt{-6.49M}}
                              & 0 $\rightarrow$ 1   & 57.93               & -$17.41\%$          \\
                              & 1 $\rightarrow$ 2   & 64.20               & -$8.56\%$           \\
                              & 2 $\rightarrow$ 3   & \textbf{66.35}      & \textbf{-5.51\%}    \\
                              & 3 $\rightarrow$ 4   & 64.65               & -$7.90\%$           \\
                              & 4 $\rightarrow$ 5   & 64.86               & -$7.60\%$           \\
                              & 5 $\rightarrow$ 6   & 58.05               & -$17.32\%$          \\
                              & 6 $\rightarrow$ 7   & 61.56               & -$12.31\%$          \\
                              & 7 $\rightarrow$ 8   & 58.53               & -$16.64\%$          \\
                              & 8 $\rightarrow$ 9   & 52.32               & -$25.50\%$          \\
                              & 9 $\rightarrow$ 10  & 59.21               & -$15.68\%$          \\
                              & 10 $\rightarrow$ 11 & 22.64               & -$67.75\%$          \\
        \midrule
        \texttt{149.07M}      & original            & \underline{70.21}   & –                   \\
        \bottomrule
        \end{tabular}
    }
\end{wraptable}
To further assess the effectiveness of our approach, we evaluate \gls{toast} in a zero-shot image classification setting. This evaluation utilizes the \clipb{} model \citep{clip}, which was pretrained on \laion{} \cite{NEURIPS2022_a1859deb}, with \imagenet{} serving as the downstream evaluation dataset. The analysis is conducted only on the base version, as larger versions (e.g., \clipl{} or \cliph{}) contain too many parameters and are thus beyond the scope of this paper. As in previous experiments, the model remains frozen, and block approximations are computed using a shared linear transformation applied across all tokens, based on a subset of 3,000 training samples. Importantly, we apply these approximations only to the vision encoder, leaving the text encoder unchanged. We follow the standard \imagenet{} prompt templates. The results in \Cref{table:clip-zeroshot} lead to the conclusion that the impact on zero-shot accuracy is highly dependent on the targeted block's position. The choice of which blocks to approximate is therefore crucial. For instance, approximating (2 $\rightarrow$ 3) results in a modest accuracy drop of $5.51\%$, yielding a competitive model with fewer parameters. In contrast, approximating the final block (i.e., 10 $\rightarrow$ 11) causes a catastrophic performance collapse of 67.75\%. This indicates that, for \clipb{}, later layers in the vision encoder appear to capture uniquely critical information for zero-shot generalization that cannot be effectively replicated by earlier ones. To the best of our knowledge, this work is the first to investigate training-free model size reduction in this challenging setting.

\subsubsection{TOAST Applicability to Other Tasks or Domains} \label{app:text-classification}

This section presents additional experiments that complement and extend those detailed in \Cref{sec:add-experiments}. Datasets and models are the ones detailed in \Cref{tab:dataset-info,tab:pretrained-info}.

\begin{table}[h]
    \centering
    \caption{\textbf{\gls{toast} Text Classification Performance on \agnews{}}. Text classification accuracy, \gls{gflops}, and throughput for \mbertb{} using \agnews{}. The ``Approx.'' column specifies the block mapping (output of the first block is used to approximate the output of the second). \gls{mlp} is a trained approximator, while Linear is closed-form and training-free. Results are averaged over three seeds. }
    \label{table:app-text-classification}
    \centering
    \resizebox{.9\textwidth}{!}{%
        \begin{tabular}{ccccccccc}
            \toprule
                                                                        &                     & \multicolumn{3}{c}{Linear} & \multicolumn{3}{c}{\gls{mlp}}                                                                                     \\
            \cmidrule(l){3-5} \cmidrule(l){6-8}
            Approx.                                                     & Params $\downarrow$ & Accuracy \% $\uparrow$      & GFLOPs $\downarrow$           & tokens/s $\uparrow$ & Accuracy \% $\uparrow$ & GFLOPs $\downarrow$ & tokens/s $\uparrow$ \\
            \midrule
            11 $\rightarrow$ 21                                         & \texttt{92.82M}     & $0.81 \pm 0.05$            & 12.7                          & 2264.0           & $0.73 \pm 0.00$       & 12.68               & 2216.50          \\
            4 $\rightarrow$ 8, 11 $\rightarrow$ 14, 18 $\rightarrow$ 21 & \texttt{92.82M}     & $0.82 \pm 0.07$            & 12.7                          & 2220.7           & $0.73 \pm 0.01$       & 12.68               & 2155.16          \\
            \midrule
            4 $\rightarrow$ 7, 18 $\rightarrow$ 21                      & \texttt{109.68M}    & $0.82 \pm 0.07$            & 15.9                          & 1803.9           & $0.71 \pm 0.02$       & 15.85               & 1771.80          \\
            \midrule
            4 $\rightarrow$ 8                                           & \texttt{126.54M}    & $0.86 \pm 0.02$            & 19.0                          & 1636.0           & $0.82 \pm 0.01$       & 19.03               & 1632.65          \\
            \midrule
            11 $\rightarrow$ 14                                         & \texttt{132.16M}    & $0.86 \pm 0.02$            & 20.1                          & 1544.3           & $0.82 \pm 0.01$       & 20.08               & 1540.23          \\
            18 $\rightarrow$ 21                                         & \texttt{132.16M}    & $0.85 \pm 0.02$            & 20.1                          & 1472.8           & $0.82 \pm 0.01$       & 20.08               & 1467.56          \\
            \midrule
            1 $\rightarrow$ 2                                           & \texttt{143.4M}    & $0.84 \pm 0.01$            & 22.2                          & 1386.0           & $0.84 \pm 0.00$       & 22.20               & 1385.20          \\
            2 $\rightarrow$ 3                                           & \texttt{143.4M}    & $0.86 \pm 0.00$            & 22.2                          & 1379.8           & $0.86 \pm 0.00$       & 22.20               & 1388.06          \\
            3 $\rightarrow$ 4                                           & \texttt{143.4M}    & $0.82 \pm 0.01$            & 22.2                          & 1391.6           & $0.83 \pm 0.00$       & 22.20               & 1392.14          \\
            4 $\rightarrow$ 5                                           & \texttt{143.4M}    & $0.88 \pm 0.00$            & 22.2                          & 1380.3           & $0.81 \pm 0.01$       & 22.20               & 1384.42          \\
            5 $\rightarrow$ 6                                           & \texttt{143.4M}    & $0.86 \pm 0.02$            & 22.2                          & 1385.0           & $0.83 \pm 0.00$       & 22.20               & 1392.14          \\
            6 $\rightarrow$ 7                                           & \texttt{143.4M}    & $0.86 \pm 0.02$            & 22.2                          & 1387.8           & $0.85 \pm 0.01$       & 22.20               & 1387.81          \\
            7 $\rightarrow$ 8                                           & \texttt{143.4M}    & $0.87 \pm 0.01$            & 22.2                          & 1384.8           & $0.85 \pm 0.00$       & 22.20               & 1365.78          \\
            8 $\rightarrow$ 9                                           & \texttt{143.4M}    & $0.84 \pm 0.01$            & 22.2                          & 1384.4           & $0.83 \pm 0.01$       & 22.20               & 1383.31          \\
            9 $\rightarrow$ 10                                          & \texttt{143.4M}    & $0.82 \pm 0.08$            & 22.2                          & 1385.3           & $0.71 \pm 0.01$       & 22.20               & 1385.92          \\
            10 $\rightarrow$ 11                                         & \texttt{143.4M}    & $0.81 \pm 0.08$            & 22.2                          & 1383.2           & $0.72 \pm 0.03$       & 22.20               & 1381.78          \\
            11 $\rightarrow$ 12                                         & \texttt{143.4M}    & $0.87 \pm 0.02$            & 22.2                          & 1378.8           & $0.82 \pm 0.01$       & 22.20               & 1394.63          \\
            12 $\rightarrow$ 13                                         & \texttt{143.4M}    & $0.86 \pm 0.02$            & 22.2                          & 1384.5           & $0.83 \pm 0.01$       & 22.20               & 1390.65          \\
            13 $\rightarrow$ 14                                         & \texttt{143.4M}    & $0.80 \pm 0.06$            & 22.2                          & 1385.2           & $0.73 \pm 0.02$       & 22.20               & 1385.23          \\
            14 $\rightarrow$ 15                                         & \texttt{143.4M}    & $0.84 \pm 0.04$            & 22.2                          & 1390.0           & $0.79 \pm 0.01$       & 22.20               & 1387.43          \\
            15 $\rightarrow$ 16                                         & \texttt{143.4M}    & $0.85 \pm 0.02$            & 22.2                          & 1402.7           & $0.82 \pm 0.00$       & 22.20               & 1381.80          \\
            16 $\rightarrow$ 17                                         & \texttt{143.4M}    & $0.87 \pm 0.01$            & 22.2                          & 1402.8           & $0.85 \pm 0.00$       & 22.20               & 1387.02          \\
            17 $\rightarrow$ 18                                         & \texttt{143.4M}    & $0.85 \pm 0.02$            & 22.2                          & 1402.3           & $0.83 \pm 0.01$       & 22.20               & 1389.71          \\
            18 $\rightarrow$ 19                                         & \texttt{143.4M}    & $0.87 \pm 0.01$            & 22.2                          & 1403.5           & $0.85 \pm 0.01$       & 22.20               & 1393.53          \\
            19 $\rightarrow$ 20                                         & \texttt{143.4M}    & $0.85 \pm 0.02$            & 22.2                          & 1403.9           & $0.82 \pm 0.00$       & 22.20               & 1390.19          \\
            20 $\rightarrow$ 21                                         & \texttt{143.4M}    & $0.87 \pm 0.02$            & 22.2                          & 1340.2           & $0.84 \pm 0.00$       & 22.20               & 1332.27          \\
            \midrule
            original                                                    & \texttt{149.01M}    & $0.88 \pm 0.00$            & 23.25                         & 1337.25          & $0.88 \pm 0.00$       & 23.25               & 1347.46          \\
            \bottomrule
        \end{tabular}
    }
\end{table}

\subsubsection{Evaluation with Original Classification Heads}\label{app:original-heads}

As mentioned in the main paper, our primary evaluation involves training a new linear classifier on top of the frozen model backbone to simulate a realistic transfer learning scenario. However, the original papers for \deits{} \citep{pmlr-v139-touvron21a} and \vits{} \citep{beyer2022better} report performance using the classification head that was part of the original pretraining.
\begin{table}[ht]
\centering
\caption{\textbf{Comparison of Original vs. Retrained Classification Heads.} \gls{toast} performance on \imagenet{} using the frozen, pretrained head (Original) versus a linear classifier trained on the frozen backbone (Retrained). All approximations in this table use the \textbf{Linear} translator. The relative ranking of approximations remains consistent across both settings.}
    \label{tab:original_heads}
    \centering
    \small
        \begin{tabular}{cccc}
        \toprule
        Encoder & Approximation & Original Head Acc. $\uparrow$ & Retrained Head Acc. $\uparrow$ \\
        \midrule
        \multirow{5}{*}{\rotatebox[origin=c]{90}{\deits{}}}
        & $3 \rightarrow 4, 9 \rightarrow 11$ & $72.44$ & $68.39 \pm 0.13$ \\
        & $3 \rightarrow 4, 9 \rightarrow 10$ & $77.25$ & $71.35 \pm 0.22$ \\
        & $2 \rightarrow 3$ & $78.69$ & $73.19 \pm 0.19$ \\
        & $10 \rightarrow 11$ & $78.78$ & $73.78 \pm 0.28$ \\
        & original & $79.66$ & $73.85 \pm 0.39$ \\
        \midrule
        \multirow{5}{*}{\rotatebox[origin=c]{90}{\vits{}}}
        & $1 \rightarrow 2$ & $76.62$ & $70.32 \pm 0.38$ \\
        & $2 \rightarrow 3$ & $78.25$ & $71.26 \pm 0.03$ \\
        & $3 \rightarrow 4$ & $78.25$ & $71.40 \pm 0.22$ \\
        & $4 \rightarrow 5$ & $77.66$ & $70.98 \pm 0.16$ \\
        & original & $79.86$ & $73.24 \pm 0.13$ \\
        \bottomrule
        \end{tabular}
        \end{table}
To confirm that our conclusions are robust and not an artifact of our evaluation protocol, we conducted an additional set of experiments using the official, pretrained classification heads from the original model checkpoints. For consistency with our main experiments, we use the same number of samples (500) for the approximation. In this setup, we do not train a new classifier; we simply evaluate the accuracy of the frozen, approximated models using their original heads. The results, presented in \Cref{tab:original_heads}, are fully consistent with the main conclusions of our paper. They confirm that our block approximation method provides a favorable accuracy-efficiency trade-off, even when evaluated with the original model heads. The relative drop in accuracy when approximating different layers follows the same patterns observed in our primary experiments, reinforcing the validity of our approach.

\subsubsection{Computational Efficiency vs. Accuracy}\label{app:acc-eff-tradeoff}
\begin{figure}[h]
    \centering
    \includegraphics[width=.7\textwidth]{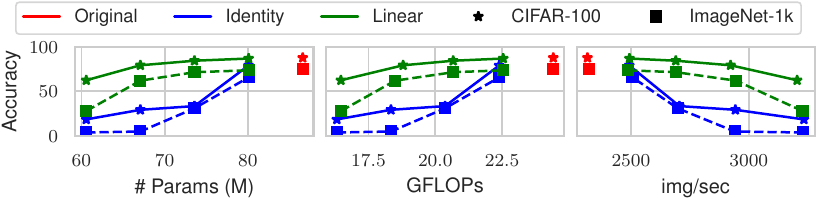}
    \caption{\textbf{Accuracy-efficiency trade-off for different approximation strategies.} Each subplot shows the accuracy against a different efficiency metric: the number of parameters (left), \gls{gflops} (center), and inference throughput (right). The image shows that the linear translator achieves a superior accuracy-efficiency trade-off.}
    \label{fig:dino_acc_eff}
\end{figure}

To quantify the effectiveness of different approximation methods, we analyze the trade-off between downstream accuracy and computational cost. \Cref{fig:dino_acc_eff} presents this analysis on a \dinob{} model using both \cifarhf{} and \imagenet{} against three standard efficiency metrics: parameter count, \gls{gflops}, and inference throughput. Across all metrics, the proposed linear translator (green) establishes a more favorable Pareto frontier compared to the baseline identity-based approach (blue). This indicates that for any given efficiency budget (e.g., a specific \gls{gflops} target), the linear translator consistently yields a model with higher accuracy.

\subsubsection{Analysis of Misclassifications}\label{app:misclassification-analysis}
\begin{figure}[h]
    \centering
    \includegraphics[width=.85\textwidth]{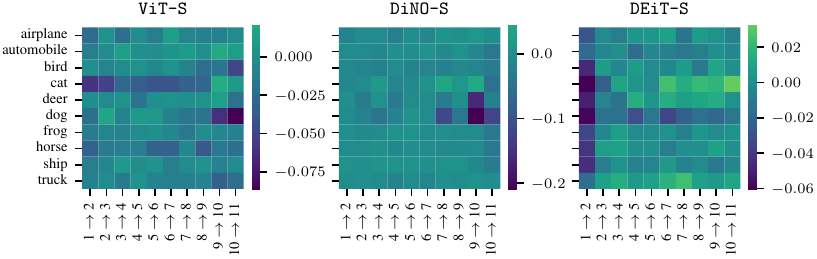}
    \caption{\textbf{Per-class accuracy delta on \cifart{} when a single block is approximated in \vits{}, \dinos{}, and \deits{}}. Cell values indicate the relative change in the accuracy with respect to the original model. Brighter (green) cells indicate an accuracy gain for the class, while darker (blue) cells indicate an accuracy drop.\looseness=-1}
    \label{fig:relative_accuracy}
\end{figure}

\begin{figure}[h]
    \centering
    \includegraphics[width=\textwidth]{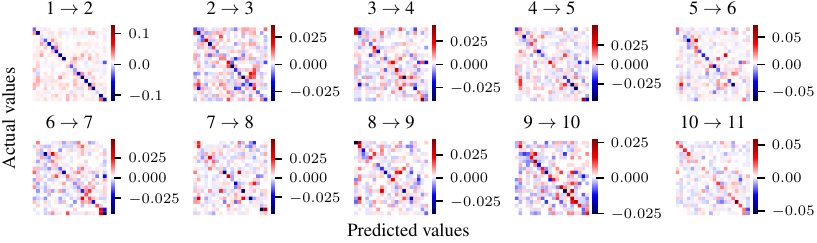}
    \caption{\textbf{Normalized relative confusion matrix when single blocks are approximated for \deits{} on \cifarhc{}}. Diagonal cells capture the per‑class change in accuracy, whereas off‑diagonal cells capture changes in misclassifications between classes. Red (positive) values on the diagonal mean the approximation improves that class's accuracy. Red off‑diagonal values mean more misclassifications. Conversely, blue (negative) off‑diagonal values indicate fewer misclassifications, and blue values on the diagonal indicate a drop in per‑class accuracy.}
    \label{fig:delta_confusion_matrix}
\end{figure}

In this section, we examine changes in per-class accuracy and misclassification patterns. As shown in \Cref{fig:relative_accuracy}, models behave differently at block approximations. \dinos{} remains remarkably stable across blocks and classes, with the only degradation appearing for classes dog (when approximating blocks 10 or 11) and deer (for block 10 approximation). \vits{} shows a similar drop for class dog on its final block. Instead, the most noticeable hit occurs for class cat when the earlier blocks are approximated. For \deits{}, several mid‑to‑late block approximations improve accuracy for various classes, whereas the very first block causes a clear relative decline in nearly every class. These observations suggest strategies like preferring late‑block approximation for \deits{}, or reserving extra samples for the linear transformation in order to recover the accuracy of difficult classes for the model.\looseness=-1
In order to further investigate how the predictions change while approximating blocks, we plot the difference in the normalized confusion matrix before and after the approximation. In \Cref{fig:delta_confusion_matrix}, we show the delta confusion matrix for \deits{} on \cifarhc{}. Also, here we can see how approximating the very first block confuses the model and degrades per-class accuracy (i.e., negative delta along the diagonal). On the other hand, approximating the last block acts as a regularizer, resulting in an overall gain in the per-class accuracy and, as a consequence, fewer misclassifications (negative deltas off-diagonal). This supports results shown in \Cref{fig:relative_accuracy,table:classification-results-cifar100}.\looseness=-1

\begin{figure}[h]
    \centering
    \includegraphics[width=.8\linewidth]{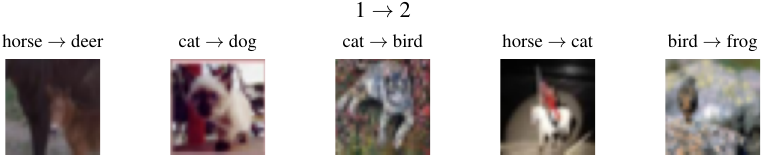}
    \includegraphics[width=.8\linewidth]{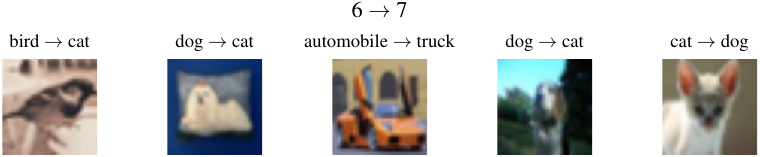}
    \includegraphics[width=.8\linewidth]{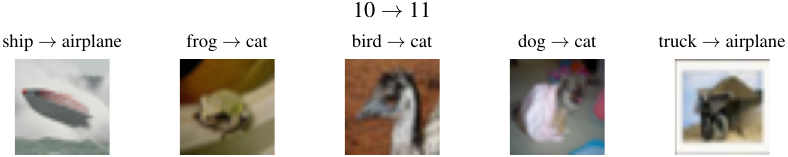}
    \caption{\textbf{Visualization of misclassified samples after approximating a block of \vits{} on \cifart{}}. Images from \cifart{} whose label \emph{flips from correct to incorrect} when specific blocks are approximated. The title reports the true class followed by the wrong prediction.}
    \label{fig:misclassification}
\end{figure}

Additionally, \Cref{fig:misclassification} shows representative \cifart{} images that become misclassified after approximating a block of \vits{}. The patterns we observe mirror the trends in \Cref{fig:relative_accuracy,fig:delta_confusion_matrix}: when approximating earlier blocks, we observe many images belonging to class cat to be misclassified. Instead, when approximating later blocks, we observe images of the class dog to be misclassified. Together, these qualitative examples show that understanding these block-specific vulnerabilities allows us to steer the approximation procedure, informing choices about which blocks to approximate based on the observed impact on the final model's class-wise performance.

\end{document}